\documentclass[11pt]{article}

\usepackage{iftex}
\ifPDFTeX
  \usepackage[T1]{fontenc}
  \usepackage[utf8]{inputenc}
\fi
\usepackage{lmodern}
\usepackage{microtype}

\usepackage[a4paper,margin=2.5cm,footskip=1.1cm]{geometry}
\usepackage{amsmath,amssymb}
\usepackage{graphicx}
\usepackage[table,dvipsnames]{xcolor}
\usepackage{booktabs}
\usepackage{longtable}
\usepackage{tabularx}
\usepackage{multirow}
\usepackage{makecell}
\usepackage{array}
\usepackage{arydshln}  
\usepackage{threeparttable}
\usepackage{rotating}
\usepackage{adjustbox}
\usepackage{fancyhdr}
\fancypagestyle{firstpage}{%
  \fancyhf{}%
  \lhead{\includegraphics[height=20pt]{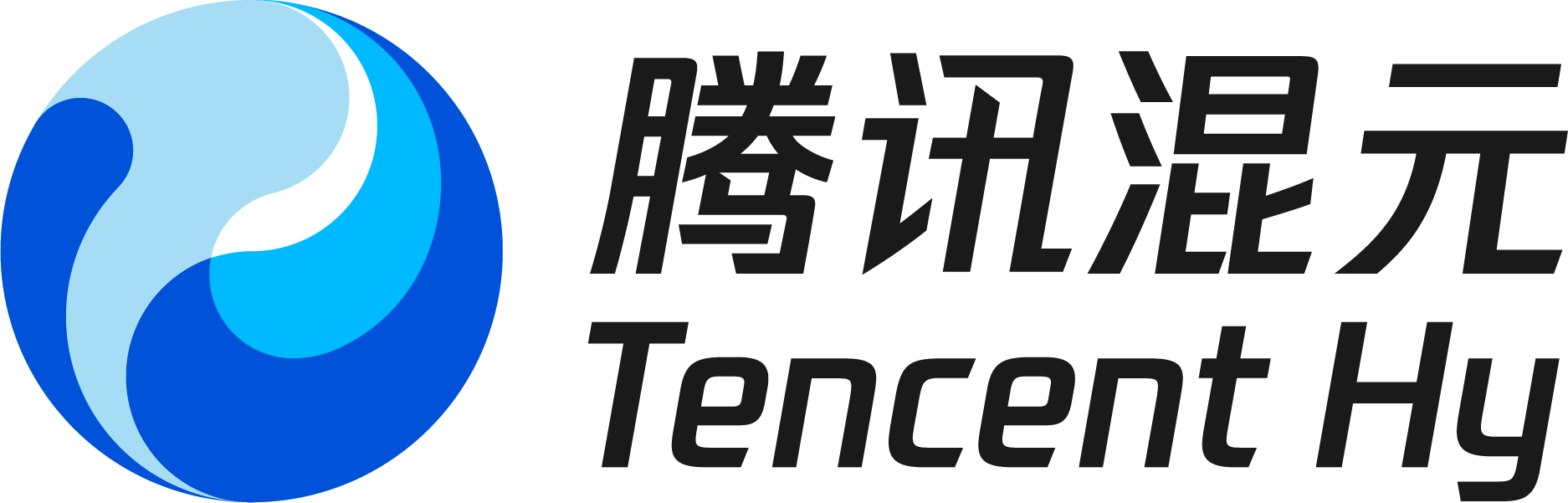}}%
  \cfoot{\thepage}%
}

\usepackage{placeins}
\usepackage{enumitem}
\usepackage{titlesec}
\usepackage[font=small,labelfont=bf,skip=6pt]{caption}
\usepackage{subcaption}
\usepackage{pgfplots}
\usepackage{tikz}
\usepackage[numbers,sort&compress]{natbib}

\pgfplotsset{compat=1.18}
\usetikzlibrary{patterns,positioning,calc,arrows.meta,shapes.geometric,fit,backgrounds,matrix}
\usepgfplotslibrary{fillbetween,groupplots,polar,colormaps}

\definecolor{linkblue}{HTML}{1A4F8B}
\definecolor{inkgray}{HTML}{4A4A4A}
\definecolor{rulegray}{HTML}{BFBFBF}

\definecolor{tierone}{HTML}{1B7F4B}
\definecolor{tiertwo}{HTML}{4E8FCF}
\definecolor{tierthree}{HTML}{D08B1E}
\definecolor{tierfour}{HTML}{C0392B}

\definecolor{axF}{HTML}{1F6FB2}   
\definecolor{axC}{HTML}{7A4FA3}   
\definecolor{axP}{HTML}{2E8B57}   
\definecolor{axQ}{HTML}{B8860B}   
\definecolor{axE}{HTML}{C0567A}   

\definecolor{mdA}{HTML}{2B3A67}
\definecolor{mdB}{HTML}{1F6FB2}
\definecolor{mdC}{HTML}{2E9BA8}
\definecolor{mdD}{HTML}{5FA85B}
\definecolor{mdE}{HTML}{D9A13B}
\definecolor{mdF}{HTML}{D3703A}
\definecolor{mdG}{HTML}{B23A48}

\definecolor{seq0}{HTML}{F7F5FB}
\definecolor{seq1}{HTML}{E4DCEF}
\definecolor{seq2}{HTML}{C9B8DF}
\definecolor{seq3}{HTML}{A48BC9}
\definecolor{seq4}{HTML}{7A5BAA}
\definecolor{seq5}{HTML}{503080}

\definecolor{humancol}{HTML}{2C3E50}
\definecolor{autocol}{HTML}{1F6FB2}
\definecolor{goodcell}{HTML}{E8F3EC}
\definecolor{badcell}{HTML}{FBEAE8}
\definecolor{softrow}{HTML}{F4F6F8}

\usepackage[colorlinks=true,allcolors=linkblue,breaklinks=true]{hyperref}
\usepackage{cleveref}

\providecommand{\zhen}[2]{#1}

\usepackage{xeCJK}
\xeCJKsetup{CJKmath=true}
\setCJKsansfont{FandolHei-Regular.otf}[BoldFont = FandolHei-Bold.otf]
\setCJKmonofont{FandolFang-Regular.otf}

\newcommand{\bench}{DramaChain\,Bench}
\newcommand{\pipe}{DramaChain Agent}
\newcommand{\labeler}{DramaChain Labeling System}
\newcommand{\judger}{DramaChain Agentic Judge}
\newcommand{\dimsys}{DramaChain Dimensions}

\newcommand{\dc}[1]{{\small\ttfamily #1}}
\newcommand{\tl}[1]{{\scriptsize\ttfamily #1}}
\newcommand{\md}[1]{\mbox{\small\ttfamily #1}}
\newcommand{\hbase}{\textsc{hb}}

\newcommand{\axtag}[2]{\textcolor{#1}{\textbf{#2}}}
\newcommand{\axF}{\axtag{axF}{F}}
\newcommand{\axC}{\axtag{axC}{C}}
\newcommand{\axP}{\axtag{axP}{P}}
\newcommand{\axQ}{\axtag{axQ}{Q}}
\newcommand{\axE}{\axtag{axE}{E}}

\definecolor{tierfive}{HTML}{8E2E22}
\newcommand{\tier}[1]{%
  \ifnum#1=1\colorbox{tierone}{\textcolor{white}{\scriptsize\bfseries\,T1\,}}\fi
  \ifnum#1=2\colorbox{tiertwo}{\textcolor{white}{\scriptsize\bfseries\,T2\,}}\fi
  \ifnum#1=3\colorbox{tierthree}{\textcolor{white}{\scriptsize\bfseries\,T3\,}}\fi
  \ifnum#1=4\colorbox{tierfour}{\textcolor{white}{\scriptsize\bfseries\,T4\,}}\fi
  \ifnum#1=5\colorbox{tierfive}{\textcolor{white}{\scriptsize\bfseries\,T5\,}}\fi
}

\newcommand{\up}[1]{\textcolor{tierone}{$+$#1}}
\newcommand{\dn}[1]{\textcolor{tierfour}{$-$#1}}

\newcommand{\best}[1]{\textbf{#1}}
\newcommand{\sbest}[1]{\underline{#1}}

\newcommand{\fail}[1]{\textcolor{tierfour}{#1}}
\newcommand{\pass}[1]{\textcolor{tierone}{#1}}

\newcolumntype{R}[1]{>{\raggedleft\arraybackslash}p{#1}}
\newcolumntype{L}[1]{>{\raggedright\arraybackslash}p{#1}}
\newcolumntype{C}[1]{>{\centering\arraybackslash}p{#1}}
\newcolumntype{d}{>{\centering\arraybackslash}p{1.05cm}}

\renewcommand{\arraystretch}{1.12}
\newcommand{\tabnote}[1]{\vspace{2pt}\par{\footnotesize\textcolor{inkgray}{#1}}}

\titleformat{\section}{\large\bfseries}{\thesection}{0.6em}{}
\titleformat{\subsection}{\normalsize\bfseries}{\thesubsection}{0.6em}{}
\titleformat{\subsubsection}{\normalsize\itshape}{\thesubsubsection}{0.6em}{}
\titlespacing*{\section}{0pt}{12pt}{5pt}
\titlespacing*{\subsection}{0pt}{9pt}{3pt}
\titlespacing*{\subsubsection}{0pt}{7pt}{2pt}

\setlist[itemize]{leftmargin=1.35em,itemsep=1.5pt,topsep=3pt,parsep=0pt}
\setlist[enumerate]{leftmargin=1.6em,itemsep=1.5pt,topsep=3pt,parsep=0pt}

\pgfplotsset{
  dcbase/.style={
    tick align=outside, tick pos=left,
    axis line style={rulegray!85, semithick},
    separate axis lines, every outer x axis line/.append style={rulegray!85},
    every outer y axis line/.append style={rulegray!85},
    grid=major, grid style={rulegray!35, very thin},
    axis on top=false,
    label style={font=\small}, tick label style={font=\footnotesize, color=inkgray},
    title style={font=\small\bfseries},
    legend style={font=\footnotesize, draw=none, fill=white, fill opacity=0.85,
                  text opacity=1, inner sep=2pt},
    legend cell align=left,
    every axis plot/.append style={thick},
  },
  dcradar/.style={
    grid=both, grid style={rulegray!45, very thin},
    axis line style={draw=none}, tick style={draw=none},
    xtick distance=1, ymin=0,
    yticklabel style={font=\tiny, color=inkgray, fill=white, fill opacity=0.75,
                      text opacity=1, inner sep=0.6pt},
    xticklabel style={font=\tiny, color=inkgray},
    legend style={font=\tiny, draw=none, fill=none},
    every axis plot/.append style={very thick, mark=none, fill opacity=0.06},
  },
  dcbar/.style={dcbase, ybar, bar width=9pt, ymajorgrids, xmajorgrids=false,
                enlarge x limits=0.09},
  dchbar/.style={dcbase, xbar, bar width=7pt, xmajorgrids, ymajorgrids=false,
                 enlarge y limits=0.07, nodes near coords align={horizontal},
                 every node near coord/.append style={font=\tiny, color=inkgray}},
}

\hypersetup{
  pdftitle={DramaChain Bench: An End-to-End Benchmark for Short-Drama Generation},
  pdfsubject={Benchmark technical report},
  bookmarksnumbered=true,
}

\crefname{section}{Sec.}{Secs.}
\Crefname{section}{Section}{Sections}
\crefname{table}{Tab.}{Tabs.}
\Crefname{table}{Table}{Tables}
\crefname{figure}{Fig.}{Figs.}
\Crefname{figure}{Figure}{Figures}
\crefname{appendix}{App.}{Apps.}

\title{\vspace{-1.4em}\bfseries
  \bench: An End-to-End Benchmark for Short-Drama Generation\\[0.25em]
  \large\normalfont From Production Pipeline and Annotation System to
  Validated Automated Evaluation}

\author{%
  Haoyuan Shi\textsuperscript{1,$*$}, \quad
  Mingtao Chen\textsuperscript{1,$*$}, \quad
  Shuo Jiang\textsuperscript{1}, \quad
  Ziyan Chen\textsuperscript{1,2} \\[0.2em]
  Xuyi Sheng\textsuperscript{3}, \quad
  Yiming Liu\textsuperscript{1}, \quad
  Ying Zhang\textsuperscript{1}, \quad
  Miao Wang\textsuperscript{1,4} \\[0.2em]
  Jianxiang Lu\textsuperscript{1}, \quad
  Fanyang Lu\textsuperscript{1}, \quad
  Songyuanyi Lu\textsuperscript{1}, \quad
  Xiele Wu\textsuperscript{1} \\[0.2em]
  Zhichao Hu\textsuperscript{1,$\dagger$}, \quad
  Yuhong Liu\textsuperscript{1}, \quad
  Richeng Xuan\textsuperscript{1,$\S$}
  \\[0.5em]
  \normalsize
  \textsuperscript{1}Hunyuan, Tencent \quad
  \textsuperscript{2}Beijing Film Academy \quad
  \textsuperscript{3}Peking University \\[0.15em]
  \normalsize
  \textsuperscript{4}Shenzhen University \\[0.15em]
  \normalsize
  \textsuperscript{$*$}Equal Contribution \qquad
  \textsuperscript{$\dagger$}Project Lead \qquad
  \textsuperscript{$\S$}Corresponding Author \\[0.5em]
  \normalsize
  Project page: \href{https://dramachain-bench.github.io/}{\texttt{https://dramachain-bench.github.io}}
}
\date{}

\begin{document}
\pagenumbering{roman}
\maketitle
\thispagestyle{firstpage}

\begin{abstract}
\noindent
Commercial short-drama production follows a multi-stage chain: script, storyboard, keyframe imagery,
shot-level video, and the finished short drama. Most existing benchmarks evaluate solely the
video-generation stage using manually constructed inputs instead of real upstream pipeline outputs. This
leaves two critical questions unanswerable: whether each stage adheres to the original
\emph{script} intent (rather than only its immediate input prompt), and whether disparate shots remain
coherent after assembly into multi-episode releases. We present \bench, the first short-drama
benchmark that evaluates every stage of the complete production chain. It is built upon three in-house
systems sharing one dimension system, \dimsys{}: five evaluation axes instantiated at every stage,
resolving into $63$ leaf dimensions. \pipe{} is calibrated against commercial short-drama platforms in both workflow and
finished short-drama quality, enabling stage-wise fair comparison across models. \labeler{} provides
three independent professional assessments for each of the $5{,}569$ items in the reference set, with
defects localised in space and time where applicable and selected from a predefined defect list.
The full annotation collection contains $17{,}488$ valid
scores and $255{,}925$ traceable attribution records. Controlled
substitutions indicate that upstream defects carry through the pipeline, so final episode quality
is not governed by video generation alone. \judger{} then scores every leaf dimension automatically,
gathering evidence through iterative tool use before judging against a per-item checklist; it
reaches a mean model-level PLCC of $0.918$ against human scores. Using this framework, we evaluate
$37$ generation models ($20$ text LLMs, $8$ image models and $9$ video models), $22$ of them with human
annotation. Judge models trained on these annotations with supervised fine-tuning and GRPO achieve
higher agreement with human ratings than the prompted agentic judge.
\end{abstract}

\clearpage
{\footnotesize\setlength{\parskip}{0pt}\linespread{0.97}\selectfont\tableofcontents}
\clearpage
\pagenumbering{arabic}

\section{Introduction}
\label{sec:intro}

Short drama has become one of the most important deployment scenarios for generative video. Global
micro-drama revenue hit USD $11$ billion in 2025 and is expected to reach USD $14$ billion in 2026.
China contributed roughly $83\%$ of 2025 revenue, while overseas markets are growing fast, with Omdia
forecasting USD $3$ billion of non-China revenue in 2026~\citep{omdia2025,omdia2026}. Commercial
platforms have settled on a mature and largely shared production paradigm. OiiOii, Flova, XiaoYunQue
and LibTV take a story premise from the user and execute a staged pipeline: write the script, break
it into a storyboard, render keyframe images, animate those keyframes into shot videos, and
finally assemble the result into the finished short drama. Academic work has converged on the same paradigm,
most typically in agent frameworks: an agentic pipeline assigns each stage to a specialised role and
hands one stage's output to the next for further processing~\citep{vimax,mavis,animaker}.
Industry platform and academic paper are therefore building the
same thing: a chain in which every stage consumes the upstream output, so what a preceding stage hands
downstream sets the ceiling on what every later stage can do.

The production chain has matured; evaluation has not adopted the same view of it. The first generation of
benchmarks targeted single video clips, scoring capability dimensions against model-level human
preference~\citep{vbench,vbench2}; a second, of which FilmBench~\citep{filmbench} is
representative, extended the target to film language and multi-shot
settings~\citep{msvbench,entitybench,vistorybench}; a third began organising its criteria along the
production workflow~\citep{evalverse,directorbench}. None of them scores what one stage of a pipeline
hands to the next. Even in the short-drama domain proper, where such an intermediate artefact is
scored, it was mined from finished drama rather than produced by the pipeline under
test~\citep{dramadirector}. Three limitations follow. (1) \textbf{Stages are tested in
isolation}, so no benchmark can say which stage caused a delivery defect or how far that defect
travelled. (2) \textbf{Inputs are
authored for the test rather than produced by a pipeline} (reverse-engineered prompts, shots cut from
existing footage), so they are clean and self-contained, unlike the machine-written artefact a stage
receives in production. (3) \textbf{Supervision does not localise the defect}: human labels validate a metric by
model-level correlation and, where single items are annotated, record a preference or an overall rating
rather than which dimension failed, where, and why, so a scorer can be validated on its ranking but not
on whether it found the defect.

\begin{figure*}[t]
\centering
\includegraphics[width=\linewidth]{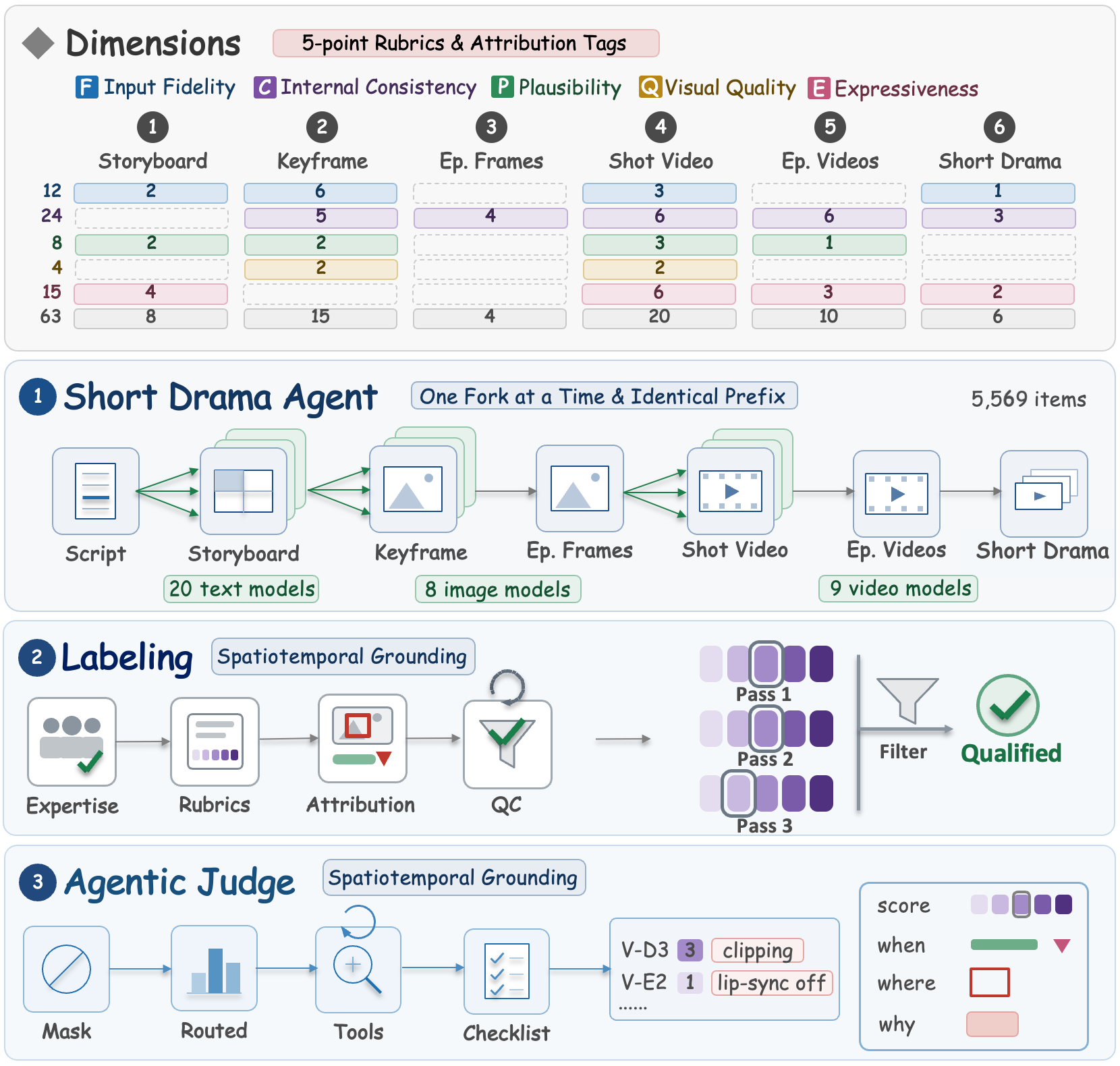}
\caption{\zhen{Overview of \bench{}. \dimsys{} defines five axes across six granularities; \pipe{}
produces items, \labeler{} supplies human references, and \judger{} automates scoring. The item
count denotes the complete three-annotator reference set.}{\bench{} 概览。\dimsys{} 在六个粒度上定义五条评价轴；
\pipe{} 产出条目，\labeler{} 提供人工参考，\judger{} 自动评分。图中条目数指具有完整三人评估的参考集。}}
\label{fig:overview}
\end{figure*}

To address these gaps we present \bench, a short-drama benchmark that scores every stage of a full
production chain, on items that chain itself produced. Five evaluation axes (input fidelity,
internal consistency, generation plausibility, visual quality and cinematic expressiveness) are
defined once and instantiated at six granularities, resolving into $63$ leaf dimensions, with a
reference set of $5{,}569$ triply annotated items. To address the limitations above, \bench{} is built on three systems of our
own (\cref{fig:overview}).

\pipe{} executes an end-to-end full short-drama production pipeline in a fully automated manner. From
supplied scripts, it generates the storyboards, keyframes, shot videos and finished dramas for the
$20$ dramas and $60$ episodes that form the foundation of our benchmark. Both its workflow
and output short-drama quality are benchmarked against real-world production platforms. This design
prevents a flawed pipeline itself from becoming the object of evaluation. The framework forks
exclusively at the stage under test: competing models are fed identical upstream artefacts and prompt
strings, while all other stages remain fixed. \labeler{} employs three professional annotators to
independently score each item across all applicable dimensions using a five-point decidable rubric
for every leaf dimension. Rubric tiers define observable criteria mapped to a closed vocabulary of
attribution tags, with score deductions localised in space and time where applicable. This yields $17{,}488$ scores
and $255{,}925$ reviewable attributions contributed by $543$ annotators. \judger{} automatically
replicates this human reference standard: it aggregates routed measurements derived from
dimension-specific criteria and may invoke external tools iteratively, before producing
final judgements against a per-item checklist. It achieves a mean model-level PLCC of $0.918$ against
the human annotation panel.

This report evaluates $22$ models with human annotation ($9$ text LLMs, $7$ image models and
$6$ video models) on every applicable dimension of the items each model produced (coverage gaps in
\cref{sec:cov}), plus $15$ further models scored by \judger{} alone, without human annotation. A shippable configuration exists in automated scores: among the annotated
models, \md{gpt-5.5-xhigh}, \md{gpt-image-2} and \md{seedance-2.0} lead their stages and deliver a
short drama at $3.30$, above the $3.0$ usable line, and \md{seedance-2.5} reaches
$3.70$. Upstream defects then accumulate along the chain rather than staying local (in automated scores of controlled substitutions, degrading
one upstream stage costs a clip $0.07$--$0.13$ points but the finished short drama $0.25$--$0.83$), so the
quality of what ships is not determined by the video generation stage alone. Automated scoring
tracks that board at a mean model-level PLCC of $0.918$, and judge models trained on the annotations
agree with human ratings more closely than the prompted judge (\cref{sec:judgetrain,sec:ablation}).

\bench{} supports extensible evaluation rather than static benchmarking: its stage-level branching
lets a new model enter with single-stage generation only, and automated scoring reduces the human
annotation needed to place it on the board. We will release all benchmark data, the judge models and
the \pipe{} generation pipeline.
\section{Related Work}
\label{sec:related}

\subsection{Storytelling video generation}
\label{sec:rw:gen}

One line of work generates a multi-shot narrative inside one model rather than a staged pipeline:
HoloCine~\citep{holocine} denoises a whole scene jointly with sparse cross-shot attention,
CausalCine~\citep{causalcine} generates causally across shot boundaries in real time. Most of the
remaining work is memory: updating keyframe stores, entity-indexed banks, slots gated at shot
boundaries, scene recall with adaptive forgetting, and a closed loop that refines each segment against a
multimodal memory~\citep{storymem,emvid,unityshots,echoforcing,videomemory,a2rd}. One work places the
problem elsewhere: ReCA~\citep{reca} locates it in how context is \emph{allocated} rather than how long it is. Control is exposed through camera paths, keyframe conditioning or storyboard
sketches~\citep{shotdirector,smartdirector,drawvideo}. That the memory literature is this large is itself the finding:
long-range consistency degrades with distance. And control stays coarse: a prompt, a camera path or a
sketch is not a per-stage specification, so a requirement that comes out wrong cannot be re-issued to
the stage responsible; only the whole run can be generated again.

The second line decomposes the problem into staged roles that hand artefacts to one another, and the
works differ mainly in where the stages are cut and whether search is added on top. The earliest are
animation systems: Anim-Director~\citep{animdirector} drives a controllable animation pipeline from a
large multimodal model, and AniMaker~\citep{animaker} follows it with multi-agent generation and an
MCTS-inspired search over candidate clips. UniVA~\citep{univa} takes the same idea to arbitrary video
tasks through a plan-and-execute agent over tool servers, while MAViS~\citep{mavis} and
ViMax~\citep{vimax} build full storytelling pipelines that distribute script, shot design, character
styling, keyframes and animation across specialised agents, with MAViS also covering audio; Co-Director~\citep{codirector}
searches creative directions with a bandit, and Soap2Soap~\citep{soap2soap} applies the same division of
labour to whole-series remaking. Two systems target short drama:
DramaDirector~\citep{dramadirector} grounds planning in geometry retrieved from real short-drama shots,
and One Sentence One Drama~\citep{onedrama} couples script generation to production through multi-agent
debate over pacing and staged review; both locate the difficulty in the hand-offs: a text storyboard
underspecifies the geometry the renderer needs, and errors accumulate across script, keyframe and video
stages unless review loops correct them. Commercial platforms run frameworks of the same
shape (\cref{sec:intro}), which is what makes per-stage evaluation both possible and necessary, and is
the position this benchmark takes.

\subsection{Datasets and benchmarks}
\label{sec:rw:bench}

Existing corpora supply material for these systems but not what a chain-level benchmark needs. They are
assembled at scale from footage the pipeline under test did not produce: movie-level material, multi-shot
corpora, shots mined from live-action drama, and subject-consistent triplets built from web video with
part of their references synthesised~\citep{moviebench,muss,cinedance,dramadirector,opens2v}. Two things are missing throughout. None
records a \emph{full generation chain} (the script, storyboard, keyframes, shot videos and the finished
short drama of one episode, produced by one pipeline), so the input to any stage is a human-made artefact
rather than the output of the model that would feed that stage in deployment. And none carries
annotation that \emph{localises} an error and \emph{attributes} it: labels are ratings or captions over a
whole clip, so a corpus can establish that one sample is worse than another without recording where the
defect is or what caused it, which is precisely the signal needed to charge a defect to a stage.

Benchmarks have grown in three steps. VBench~\citep{vbench,vbench2} set the template: decompose quality
into capability dimensions, implement each with a generalist VLM or a specialist detector, and validate
the suite against model-level human preference. A first group refines it, moving to MLLM-based scoring,
adding per-sample attribution, testing how far an omni-LLM judge can be trusted, and isolating physical
commonsense~\citep{videobench,videogeneval,qsave,omnijudge,videophy2}; the recurring finding is that such
judges are competitive on semantic alignment and weak wherever temporal resolution matters, a split this
report reproduces per dimension (\cref{sec:meta}). A second wave keeps the template and extends the
target (film language, cross-shot consistency, storyboard panels, multi-talker and multi-reference
audio-video, narrative richness, long compositional prompts, memory and minute-scale
generation~\citep{filmbench,msvbench,entitybench,vistorybench,mtavg2,multiref,narrlv,locot2v,%
vgifscore,mbench,longav}), alongside a line that asks whether vision-language models read cinematography
at all~\citep{cinetech,shotbench}; what carries a score throughout is a finished artefact rather than the
hand-offs that produced it. VideoWeaver~\citep{videoweaver} marks the boundary: its judge does read the
agent's execution trace and intermediate files, but it asks whether the workflow completed its steps, not
whether each intermediate artefact meets a rubric. The same movement is visible outside video: in text-to-image generation,
Qwen-Image-Bench~\citep{qwenimagebench} replaces prompt-level alignment scores with a taxonomy of
verifiable rubrics designed with professional artists. The closest work aligns
evaluation with the workflow: EvalVerse~\citep{evalverse} is pipeline-aware, though what is aligned there
is the \emph{criteria} while items are still evaluated as finished video, so a defect cannot be charged to
a stage; DirectorBench~\citep{directorbench} records whether each checkpoint passes and locates the
bottleneck between units rather than inside them, the same shape as the finding in \cref{sec:prop},
but it never scores the artefact one stage hands to the next against a rubric. In short drama the evaluation suites arrive attached to generation
systems and still stop short of the chain: script continuation, a storyboard-and-video protocol over
storyboards mined from real dramas, and the pipeline as a black
box~\citep{dramabenchscript,dramadirector,onedrama}. \Cref{tab:related} places \bench{} against this
literature, where two properties hold across every prior row: the items are authored for evaluation or
cut from existing media, so the input to the stage under test never came from the model that would
actually feed it; and human labels validate a metric by model-level correlation, so where single items
are annotated the label is a preference or an overall rating rather than a per-dimension score with the
deduction localised.

\providecommand{\nr}{{\color{inkgray}n/r}}
\begin{table}[htbp]
\centering
\caption{\zhen{Comparison with related benchmarks. \emph{Full-chain} jointly evaluates native intermediate
outputs and assembled multi-episode short dramas. Outputs: S storyboards, I images, V videos,
D finished dramas (episode/series-level). Counts are reported or estimated ($\approx$, $<$);
\nr{}: insufficient information; $\times$: not provided by the published protocol.
$^{\dagger}$: evaluation suite in a generation-system paper. Our benchmark counts triply annotated instances.}{与相关基准的对比。
\emph{Full-chain} 指对原生中间产物和组装后的多集短剧进行联合评估。
输出类型：S分镜、I图像、V视频、D成片（单集／整剧级）。数量采用披露值或估计值（$\approx$、$<$）；
\nr{}：信息不足；$\times$：已发表协议未提供该项能力。$^{\dagger}$：生成系统论文中的评测集。
本基准统计具有三份独立人工标注的样本。}}
\label{tab:related}
\footnotesize
\setlength{\tabcolsep}{4.5pt}
\begin{adjustbox}{max width=\linewidth}
\begin{tabular}{@{}l ccccc@{}}
\toprule
\textbf{Benchmark} &
\makecell{Full-Chain\\Short-Drama\\Evaluation} &
\makecell{Outputs\\Scored} &
\makecell{Leaf\\Dimensions} &
\makecell{Human-\\Annotated\\Items} &
\makecell{Deductions Localised\\in Space \& Time} \\
\midrule
VBench \citep{vbench} & \fail{$\times$} & V & $16$ & \nr & \fail{$\times$} \\
VBench-2.0 \citep{vbench2} & \fail{$\times$} & V & $18$ & \nr & \fail{$\times$} \\
FilmBench \citep{filmbench} & \fail{$\times$} & V & $38$ & $2{,}878$ & \fail{$\times$} \\
MSVBench \citep{msvbench} & \fail{$\times$} & V & $20$ & $\approx\!300$ & \fail{$\times$} \\
EntityBench \citep{entitybench} & \fail{$\times$} & V & $51$ & $200$ & \fail{$\times$} \\
EvalVerse \citep{evalverse} & \fail{$\times$} & V & $45$ & \nr & \fail{$\times$} \\
DirectorBench \citep{directorbench} & \fail{$\times$} & V & $40$ & $<\!100$ & \fail{$\times$} \\
Short-Drama-Bench$^{\dagger}$ \citep{onedrama} & \fail{$\times$} & V & $15$ & $\approx\!1.8$k & \fail{$\times$} \\
DramaBoard$^{\dagger}$ \citep{dramadirector} & \fail{$\times$} & S, I, V & $9$ & $\approx\!120$ & \fail{$\times$} \\
VideoWeaver$^{\dagger}$ \citep{videoweaver} & \fail{$\times$} & V & $13$ & $200$ & \fail{$\times$} \\
\midrule
\textbf{\bench{}} (this report) &
\pass{$\checkmark$} & \textbf{S, I, V, D} & $\mathbf{63}$ & $\mathbf{5{,}569}$ & \pass{$\checkmark$} \\
\bottomrule
\end{tabular}
\end{adjustbox}
\end{table}

\section{\bench}
\label{sec:bench}
\label{sec:benchmark}

\bench{} consists of three built infrastructure components and \dimsys{}, rather than a standalone
dataset release.
\textbf{\dimsys{}} (\cref{sec:dims}): all six stages share the same five evaluation axes,
so one criterion can be followed along the chain and the effect of an upstream defect on everything
downstream becomes observable.
\textbf{\pipe{}} (\cref{sec:sourcing}): items are executed, not authored, on a self-built pipeline
benchmarked against commercial short-drama platforms. Forking occurs exclusively at the stage under
test, enabling stage-wise comparability across models.
\textbf{\labeler{}} (\cref{sec:annot}): three annotators score independently and localise every
deduction in space and time with a tag from a fixed vocabulary, which makes it verifiable whether
automated scoring found the right defect and not merely the right score.
\textbf{\judger{}} (\cref{sec:judge}): rather than let small-model measurements feed a
large-model verdict, we discard the metrics that disagree with humans, gather evidence through iterative
agentic tool calls, and score against a per-item checklist written in the same rubric text and defect vocabulary the annotators used.

\subsection{\dimsys{}: five axes across six stages, $63$ leaf dimensions}
\label{sec:dims}

All six stages share one set of five evaluation axes, and sharing them is what makes a defect traceable
rather than merely visible. A storyboard that crams several actions into one shot is scored on \axP{}
for its executability, and the shots generated from that storyboard are scored on \axP{} again, so a
video that comes out physically incoherent can be read against the plausibility of the instruction it
was given instead of being charged to the video model by default. How far such a defect travels is
measured in \cref{sec:prop}.

\axF{} \textbf{input fidelity} asks whether the stage did what its input asked, checked against the
upstream artefact. \axC{} \textbf{internal consistency} asks whether it is the same person, room and
prop as in the other shot, checked against a sibling artefact. \axP{} \textbf{generation plausibility} asks whether
this is physically and causally possible, with no external referent. \axQ{} \textbf{visual quality} is
technical and aesthetic quality of the pixels, pure perception. \axE{} \textbf{cinematic expressiveness}
covers framing, camera work, performance, audio design and watchability: perception plus taste. Each
leaf dimension under them carries a five-level decidable description and a fixed set of attribution
tags, and \textbf{humans and the automated scorer read the same rubric text and draw defect tags from
the same vocabulary} (\cref{sec:tagvocab}). \Cref{fig:teaser} gives the resulting grid; which axes
activate at which stage, the full per-dimension definitions and the schema's own counting are
\cref{app:dims}.

\begin{figure}[htbp]
\centering
\begin{adjustbox}{max width=\linewidth}
\begin{tikzpicture}[
  x=1cm, y=1cm,
  stagenum/.style={circle, fill=inkgray, text=white, font=\scriptsize\bfseries,
                   inner sep=0pt, minimum size=3.6mm},
  stagename/.style={font=\scriptsize\bfseries, align=center, inner sep=0pt},
  modlab/.style={font=\scriptsize, text=inkgray!25!black, inner sep=0pt},
  leafbox/.style={draw=#1!45, fill=#1!8, rounded corners=1.5pt, line width=0.4pt,
                  text width=2.24cm, align=left, font=\tiny,
                  inner xsep=2.6pt, inner ysep=2.4pt},
  slot/.style={draw=rulegray!85, densely dashed, line width=0.4pt, rounded corners=1.5pt,
               minimum width=2.423cm, minimum height=0.34cm, inner sep=0pt},
  axtile/.style={fill=#1, text=white, font=\scriptsize\bfseries, inner sep=0pt,
                 minimum size=3.4mm, rounded corners=0.7pt},
  flowarr/.style={-{Stealth[length=1.6mm,width=1.2mm]}, draw=inkgray!65, line width=0.5pt},
]

\hyphenpenalty=10000 \exhyphenpenalty=10000

\def\xa{1.333}\def\xb{4.0}\def\xc{6.667}\def\xd{9.333}\def\xe{12.0}\def\xf{14.667}
\def\ya{-3.40}\def\yb{-5.66}\def\yc{-7.62}\def\yd{-8.91}\def\ye{-10.56}

\foreach \la/\lb/\lm/\tone/\lab in {%
  0.00/2.667/1.333/9/{\textbf{Text} $\cdot$ 8 dims},
  2.667/8.000/5.333/16/{\textbf{Image} $\cdot$ 19 dims},
  8.000/16.00/12.00/24/{\textbf{Video} $\cdot$ 36 dims}}
{
  \fill[inkgray!\tone] (\la,0) -- ({\lb-0.19},0) -- (\lb,-0.23) -- ({\lb-0.19},-0.46)
                    -- (\la,-0.46) -- ({\la+0.19},-0.23) -- cycle;
  \node[modlab] at (\lm,-0.23) {\lab};
}

\foreach \cx/\num/\nm/\mo in {%
  \xa/1/{Storyboard\\design}/{20 models},
  \xb/2/{Single\\keyframe}/{8 models},
  \xc/3/{Episode\\keyframes}/{8 models},
  \xd/4/{Shot\\video}/{9 models},
  \xe/5/{Episode\\video}/{9 models},
  \xf/6/{Short\\drama}/{9 models}}
{
  \node[stagenum] (g\num) at (\cx,-0.86) {\num};
  \node[stagename, anchor=north] at (\cx,-1.08) {\nm};
  \node[modlab, anchor=north] at (\cx,-1.80) {\mo};
}
\foreach \i/\j in {1/2, 2/3, 3/4, 4/5, 5/6} { \draw[flowarr] (g\i) -- (g\j); }
\draw[flowarr] (0.28,-0.86) -- (g1);
\draw[flowarr] (g6) -- (15.72,-0.86);
\draw[rulegray, line width=0.5pt] (0,-2.14) -- (16,-2.14);

\foreach \bt/\code/\col/\nm/\tot in {%
  -2.26/F/axF/{Input fidelity}/12,
  -4.40/C/axC/{Internal consistency}/24,
  -6.85/P/axP/{Generation plausibility}/8,
  -8.26/Q/axQ/{Visual quality}/4,
  -9.42/E/axE/{Cinematic expressiveness}/15}
{
  \fill[\col!11, rounded corners=1.5pt] (0,\bt) rectangle (16,{\bt-0.28});
  \node[axtile=\col] at (0.30,{\bt-0.14}) {\code};
  \node[anchor=west, font=\scriptsize\bfseries, text=\col!35!black]
    at (0.54,{\bt-0.145}) {\nm};
  \node[anchor=east, font=\tiny, text=\col!40!black]
    at (15.84,{\bt-0.145}) {\tot\ of $63$ leaf dimensions};
}

\foreach \cx/\yy in {\xc/\ya, \xe/\ya, \xa/\yb, \xc/\yc, \xf/\yc,
                     \xa/\yd, \xc/\yd, \xe/\yd, \xf/\yd, \xb/\ye, \xc/\ye}
  { \node[slot] at (\cx,\yy) {}; }

\node[leafbox=axF] (nF1) at (\xa,\ya)
  {Event coverage\\Dialogue fidelity};
\node[leafbox=axF] (nF2) at (\xb,\ya)
  {Subject attributes\\Action, interaction\\Scene \& layout\\Framing \& style\\%
   Face identity\\Appearance};
\node[leafbox=axF] (nF4) at (\xd,\ya)
  {Subject attributes\\Framing \& style\\Frame adherence};
\node[leafbox=axF] (nF6) at (\xf,\ya)
  {Event completion};

\node[leafbox=axC] (nC2) at (\xb,\yb)
  {Character\\Subject--scene fit\\Scene vs.\ reference\\Screen direction\\Style};
\node[leafbox=axC] (nC3) at (\xc,\yb)
  {Cross-shot character\\Subject--scene fit\\Cross-shot scene\\Cross-shot style};
\node[leafbox=axC] (nC4) at (\xd,\yb)
  {Character\\Scene \& props\\Style\\Action\\Subject--scene fit\\On-screen text};
\node[leafbox=axC] (nC5) at (\xe,\yb)
  {Cross-shot character\\Cross-shot scene \& props\\Cross-shot style\\%
   Cross-shot action\\Subject--scene fit\\Cross-shot text};
\node[leafbox=axC] (nC6) at (\xf,\yb)
  {Style\\Cross-episode consistency\\Cross-episode progression};

\node[leafbox=axP] (nP1) at (\xa,\yc)
  {Executability\\Addition restraint};
\node[leafbox=axP] (nP2) at (\xb,\yc)
  {Human anatomy\\Scene physics};
\node[leafbox=axP] (nP4) at (\xd,\yc)
  {Motion smoothness\\Physics adherence\\Causal \& temporal};
\node[leafbox=axP] (nP5) at (\xe,\yc)
  {Cross-shot causal \& temporal};

\node[leafbox=axQ] (nQ2) at (\xb,\yd)
  {Technical quality\\Aesthetic style};
\node[leafbox=axQ] (nQ4) at (\xd,\yd)
  {Picture \& material\\Style \& motion};

\node[leafbox=axE] (nE1) at (\xa,\ye)
  {Narrative flow\\Shooting rhythm\\Emotion delivery\\Audiovisual style};
\node[leafbox=axE] (nE4) at (\xd,\ye)
  {Speech quality\\Audio-visual sync\\Sound \& music\\Composition\\Camera movement\\%
   Performance};
\node[leafbox=axE] (nE5) at (\xe,\ye)
  {Cross-shot speech\\Cross-shot sound \& music\\Cross-shot performance};
\node[leafbox=axE] (nE6) at (\xf,\ye)
  {Watchability\\Pacing \& cuts};

\begin{scope}[on background layer]
\foreach \col/\na/\nb in {%
  axF/nF1/nF2,
  axC/nC2/nC3, axC/nC3/nC4, axC/nC4/nC5, axC/nC5/nC6,
  axP/nP1/nP2, axP/nP4/nP5,
  axE/nE4/nE5, axE/nE5/nE6}
{
  \fill[\col!16] (\na.north east) -- (\nb.north west) -- (\nb.south west)
              -- (\na.south east) -- cycle;
  \draw[\col!45, line width=0.4pt] (\na.north east) -- (\nb.north west);
  \draw[\col!45, line width=0.4pt] (\na.south east) -- (\nb.south west);
}
\end{scope}

\end{tikzpicture}
\end{adjustbox}
\caption{\zhen{The overview of \dimsys{}. Five evaluation axes are defined once and instantiated at the
six production granularities, resolving into $63$ leaf dimensions.}{\dimsys{} 概览。
五条评测轴只定义一次，再在六个生产粒度上各自实例化，细化为 $63$ 个叶子维度。}}
\label{fig:teaser}
\end{figure}

\subsection{\pipe{}: building a short-drama production system to commercial-platform parity}
\label{sec:sourcing}

The items are produced by a pipeline rather than written for the evaluation. \pipe{} is the short-drama
generation agent system the benchmark is generated on, built on the ViMax framework~\citep{vimax} with
its chain reverse-engineered from commercial platforms and then optimised. Alongside the six stages
that are scored it produces the intermediate artefacts a production line needs: character portrait sheets, scene and element
reference sheets, per-shot descriptions and a camera tree. It has the following three properties.

\begin{figure*}[htbp]
\centering
\includegraphics[width=\linewidth]{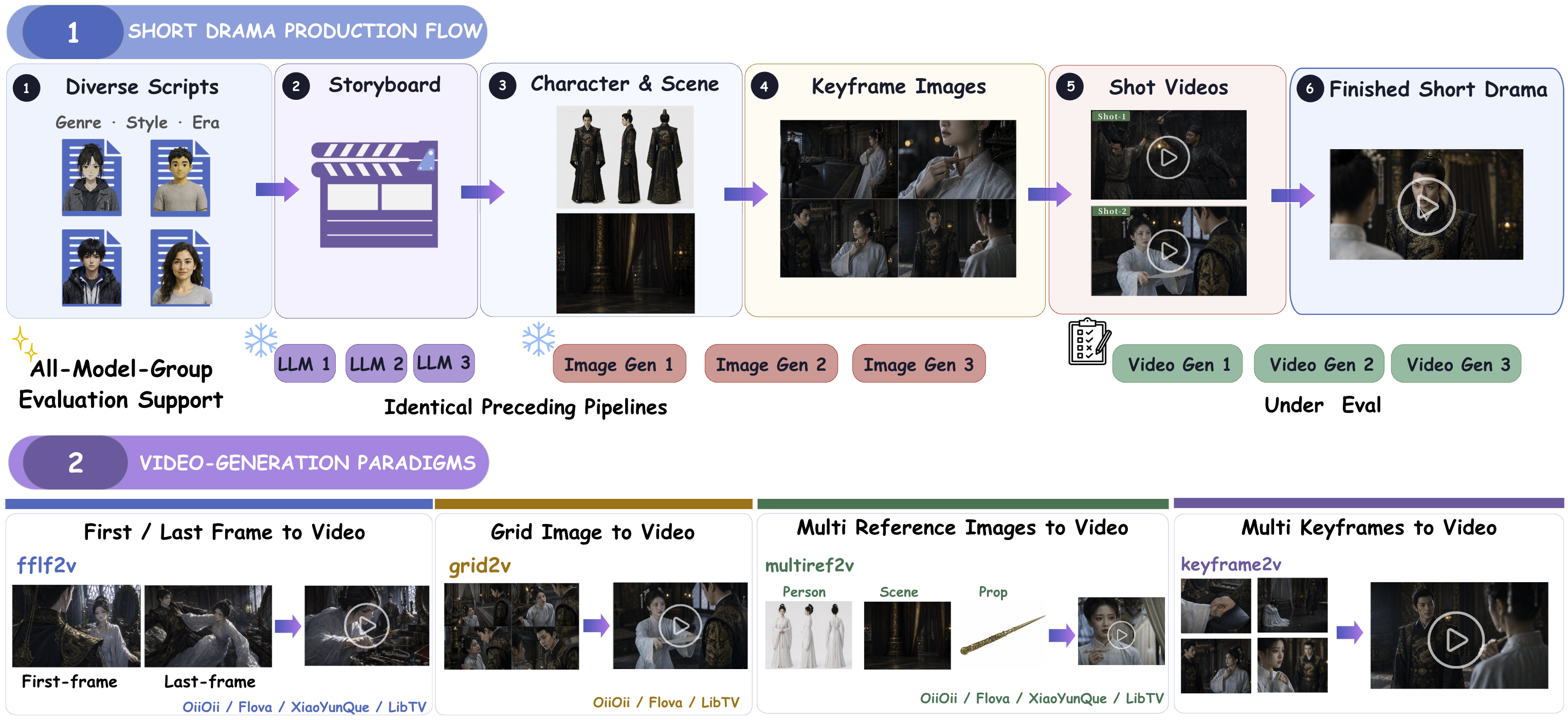}
\caption{\zhen{The overview of \pipe{}. It runs the six-stage production chain and forks only at the
stage under evaluation, so competing models receive the identical upstream artefact, and the quality of
the short drama it delivers is benchmarked against commercial short-drama platforms.}{\pipe{} 概览。
它跑完整的六环节生产链路，且只在被评环节分叉，因此参评模型拿到的是完全相同的上游产物；
它交付的短剧质量对标商用短剧平台。}}
\label{fig:sourcing}
\end{figure*}

\begin{figure}[htbp]
\centering
\setlength{\tabcolsep}{1.5pt}
\renewcommand{\arraystretch}{0.62}
\begin{tabular}{@{}cccc@{}}
\includegraphics[width=0.243\linewidth]{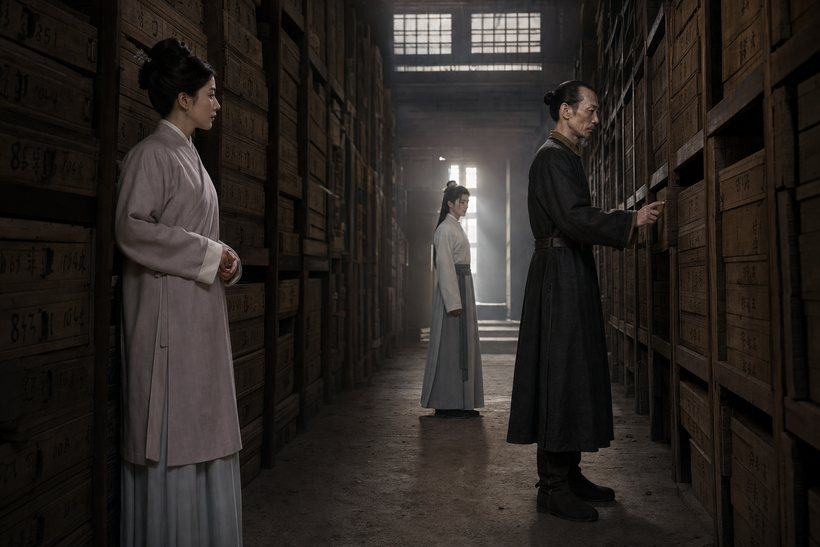} &
\includegraphics[width=0.243\linewidth]{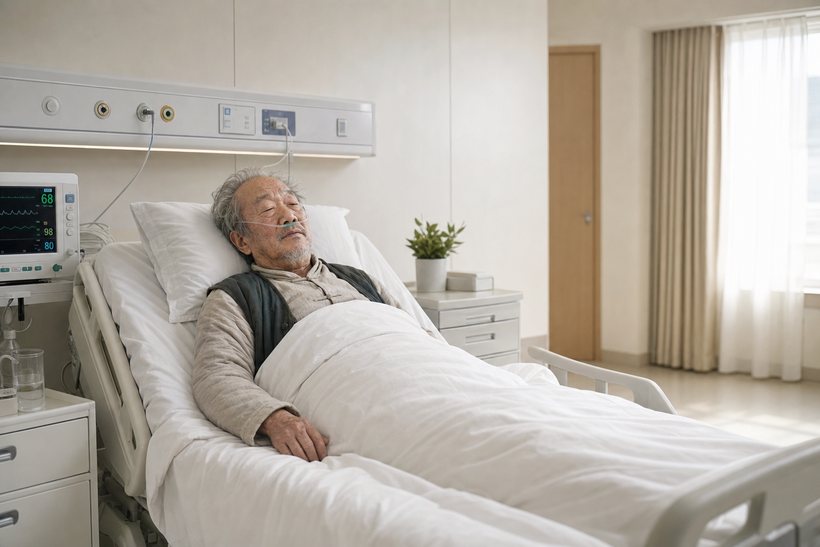} &
\includegraphics[width=0.243\linewidth]{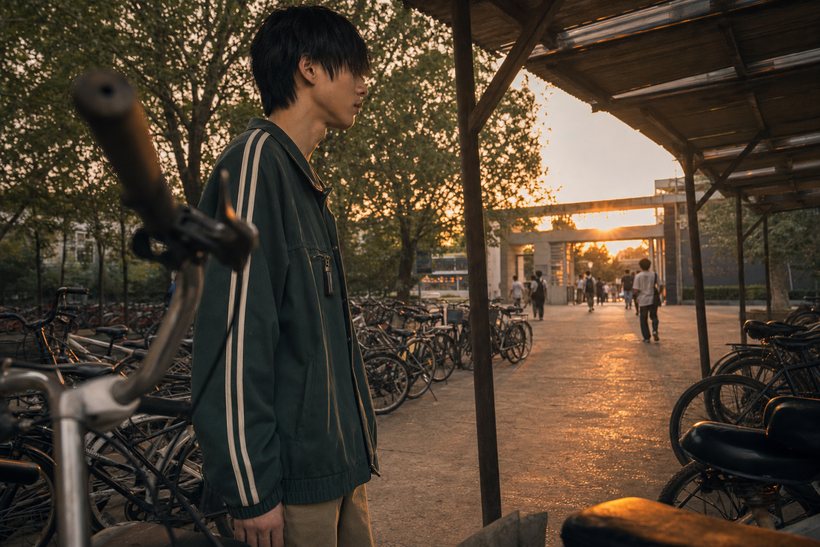} &
\includegraphics[width=0.243\linewidth]{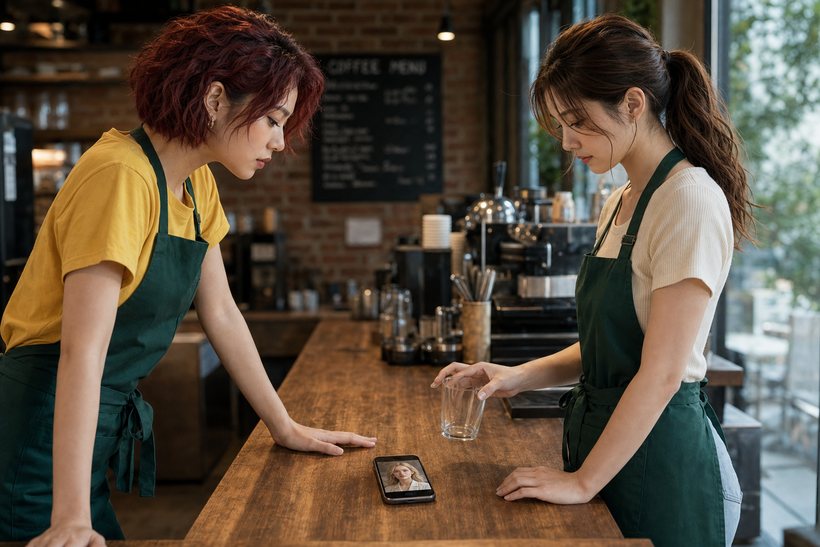} \\
{\scriptsize live action $\cdot$ zh} & {\scriptsize live action $\cdot$ zh} &
{\scriptsize live action $\cdot$ zh} & {\scriptsize live action $\cdot$ \textbf{en}} \\
{\tiny court intrigue} & {\tiny urban romance} & {\tiny campus youth} & {\tiny contract marriage} \\
\addlinespace[3pt]
\includegraphics[width=0.243\linewidth]{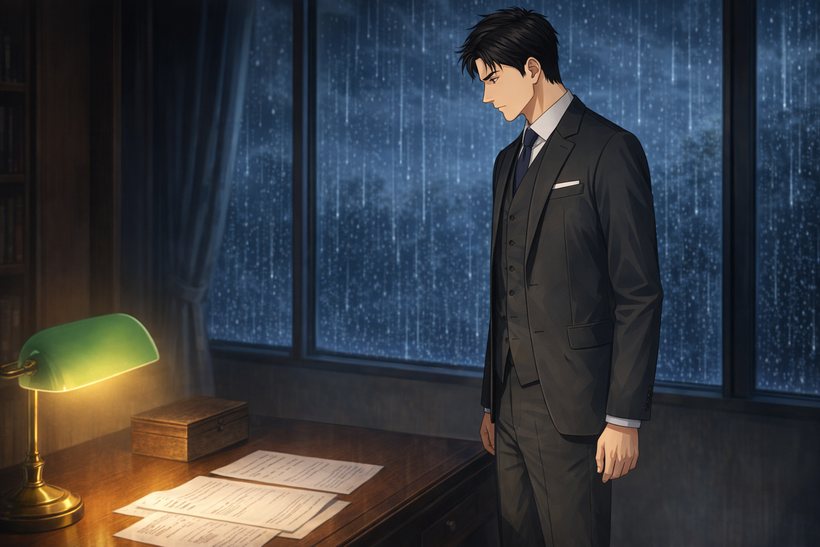} &
\includegraphics[width=0.243\linewidth]{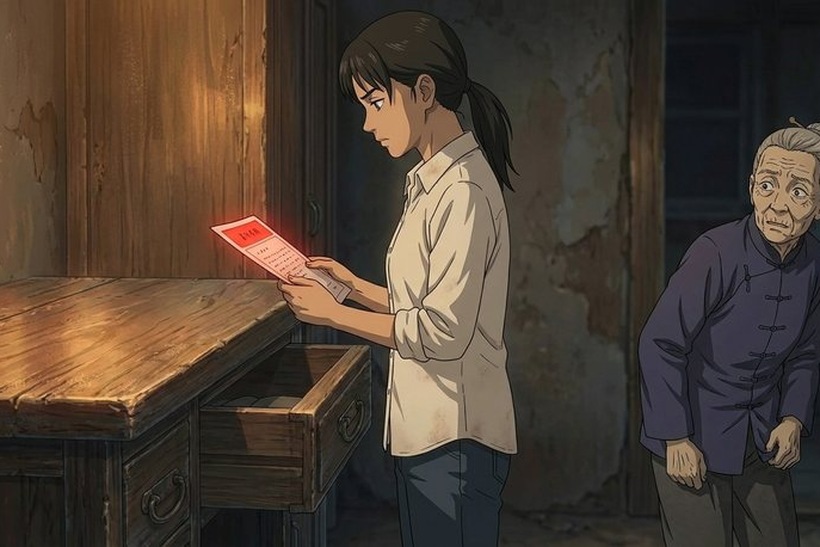} &
\includegraphics[width=0.243\linewidth]{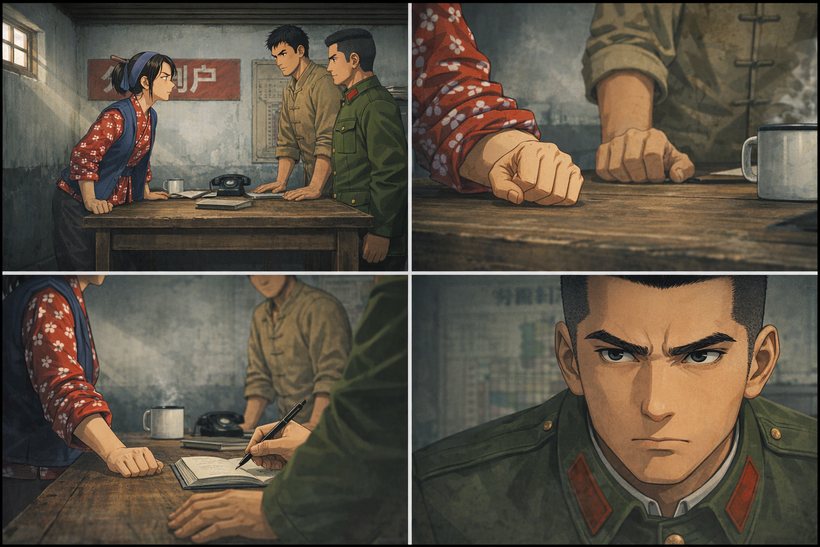} &
\includegraphics[width=0.243\linewidth]{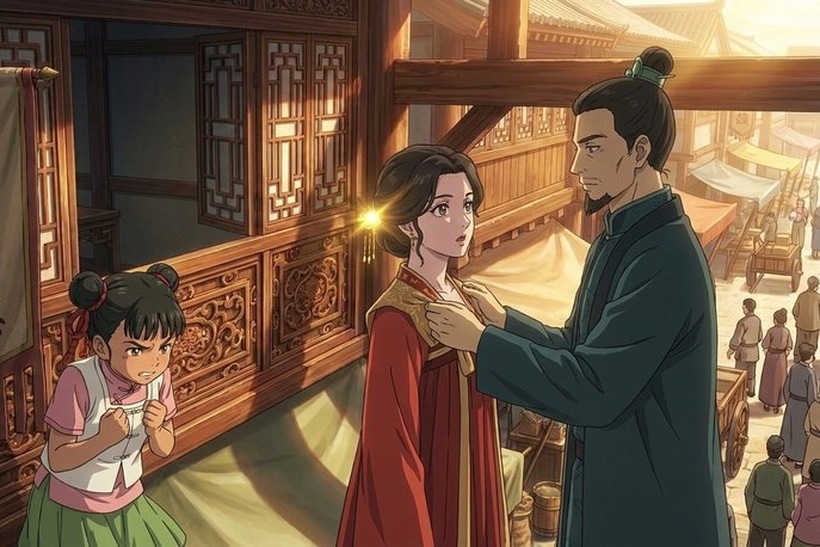} \\
{\scriptsize animated $\cdot$ zh} & {\scriptsize animated $\cdot$ zh} &
{\scriptsize animated $\cdot$ zh} & {\scriptsize animated $\cdot$ zh} \\
{\tiny dynastic romance} & {\tiny swapped heiresses} & {\tiny rural farming} & {\tiny forced possession} \\
\end{tabular}

\vspace{4pt}
\begin{adjustbox}{max width=\linewidth}
\small
\begin{tabular}{@{}l@{\hspace{9pt}}l@{\hskip 22pt}l@{\hspace{9pt}}l@{}}
\toprule
\textbf{Coverage} & \textbf{Split} & \textbf{Coverage} & \textbf{Split} \\
\midrule
Channel & male / all-audience / female &
  Visual style & live action 10 : animated 10 \\
Period & contemporary, historical, modern, fantasy &
  Dialogue language & Chinese 16 : English 4, scene descriptions all Chinese \\
Genre & 20 distinct sub-genres, one per drama &
  Input paradigm & \texttt{fflf2v}, \texttt{grid2v}, \texttt{multiref2v} \\
\bottomrule
\end{tabular}
\end{adjustbox}

\caption{\zhen{What the benchmark's items look like, with the corpus splits below. Full manifest:
\cref{app:dramas}.}{基准的题面长什么样，下方为语料的各切面构成。完整清单见 \cref{app:dramas}。}}
\label{fig:corpus}
\end{figure}

\paragraph{Data diversity.}
$20$ short dramas $\times$ $3$ consecutive episodes $=60$ episodes, spread across channel, period and
genre, crossed with visual style (live-action / animated, $10$:$10$) and dialogue language (Chinese /
English, $16$:$4$) so every slice supports a stratified check. The scripts are supplied inputs rather
than pipeline outputs; each drama has a shared cast list and follows a commercial shooting-script
format (\cref{app:dramas}). \Cref{fig:corpus} shows eight items drawn from eight of the
twenty dramas; the full manifest is \cref{app:dramas}.

\begin{figure}[htbp]
\centering
\setlength{\tabcolsep}{1.1pt}
\renewcommand{\arraystretch}{0.6}
\newcommand{\parfrm}[1]{\raisebox{-0.5\height}{%
  \includegraphics[width=0.229\linewidth]{figures/img/#1}}}
\begin{tabular}{@{}r@{\hspace{3pt}}cccc@{}}
&
\shortstack{{\scriptsize\textbf{Ours}}\\[0.5pt]{\tiny 65\,s $\cdot$ 720p $\cdot$ 24\,fps}} &
\shortstack{{\scriptsize OiiOii}\\[0.5pt]{\tiny 98\,s $\cdot$ 720p $\cdot$ 30\,fps}} &
\shortstack{{\scriptsize Xiaoyunque}\\[0.5pt]{\tiny 43\,s $\cdot$ 720p $\cdot$ 30\,fps}} &
\shortstack{{\scriptsize Flova}\\[0.5pt]{\tiny 72\,s $\cdot$ 1080p $\cdot$ 30\,fps}} \\
\addlinespace[2pt]
\multicolumn{5}{@{}l}{\tiny\textcolor{inkgray}{scene 1 $\cdot$ peril on the old road}} \\
{\tiny 6\%} & \parfrm{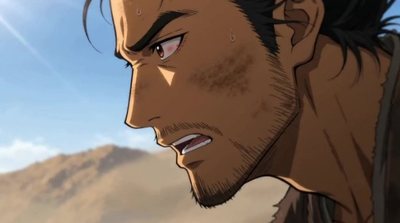} & \parfrm{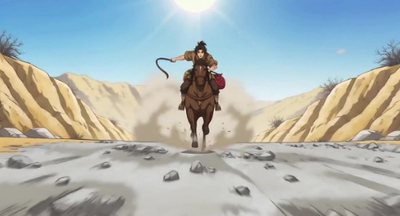} & \parfrm{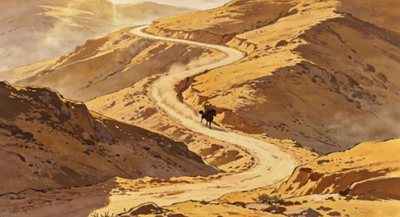} & \parfrm{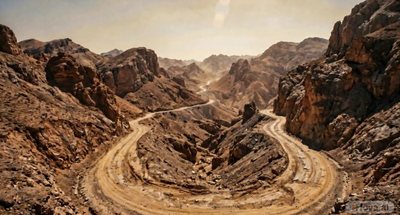} \\
{\tiny 19\%} & \parfrm{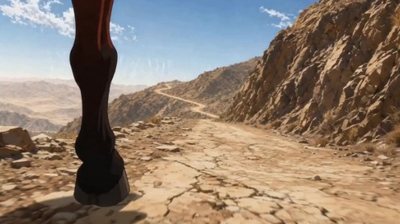} & \parfrm{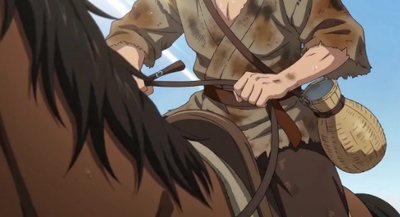} & \parfrm{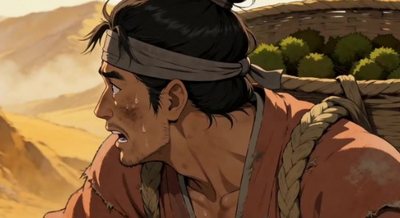} & \parfrm{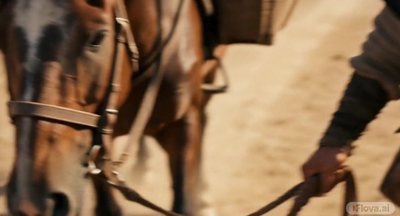} \\
\addlinespace[2pt]
\multicolumn{5}{@{}l}{\tiny\textcolor{inkgray}{scene 2 $\cdot$ relief arrives by air}} \\
{\tiny 31\%} & \parfrm{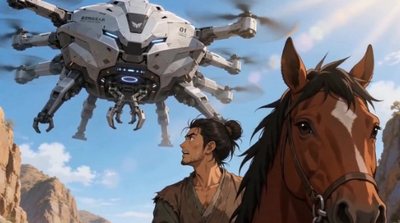} & \parfrm{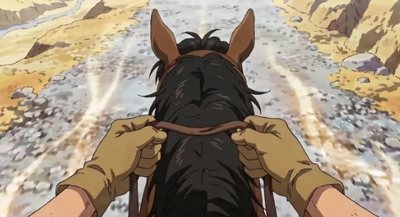} & \parfrm{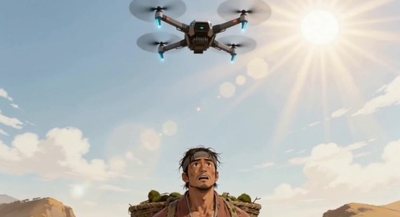} & \parfrm{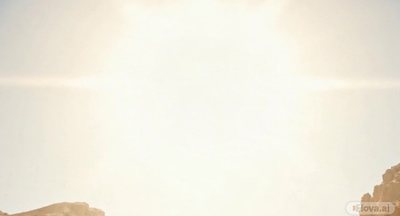} \\
{\tiny 44\%} & \parfrm{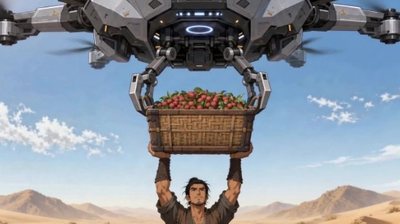} & \parfrm{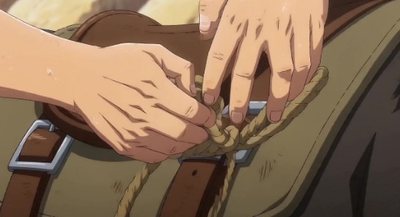} & \parfrm{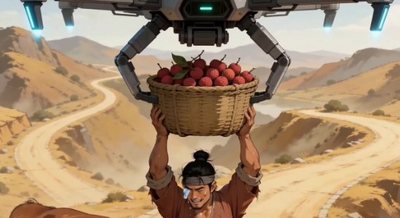} & \parfrm{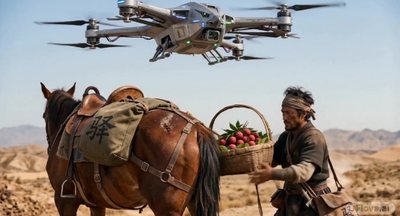} \\
{\tiny 56\%} & \parfrm{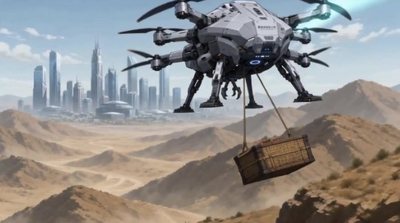} & \parfrm{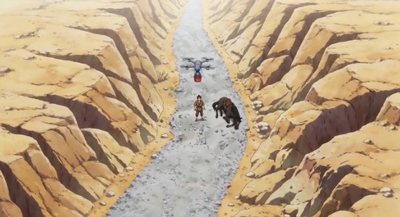} & \parfrm{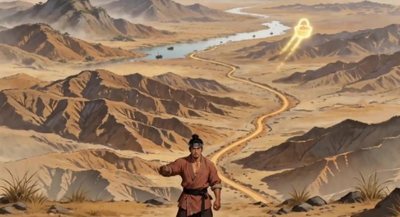} & \parfrm{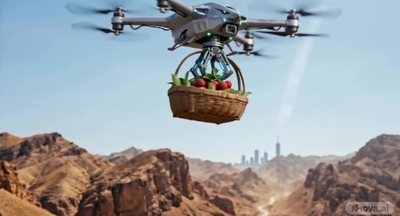} \\
\addlinespace[2pt]
\multicolumn{5}{@{}l}{\tiny\textcolor{inkgray}{scene 3 $\cdot$ fresh fruit reaches home}} \\
{\tiny 69\%} & \parfrm{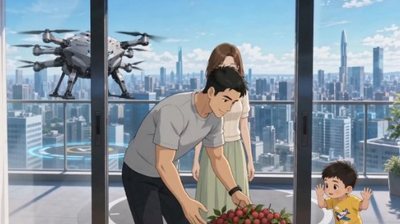} & \parfrm{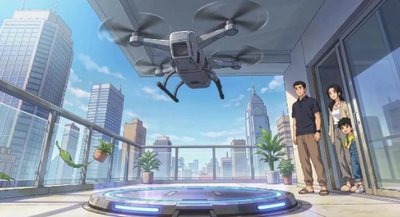} & \parfrm{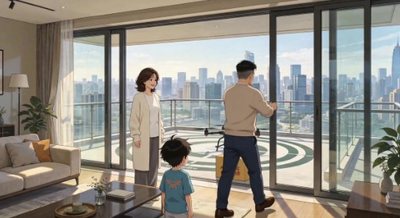} & \parfrm{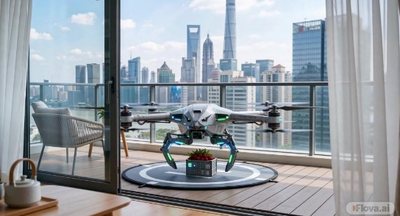} \\
{\tiny 81\%} & \parfrm{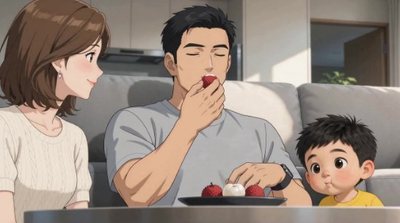} & \parfrm{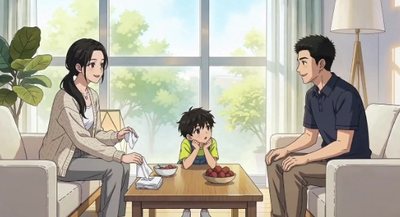} & \parfrm{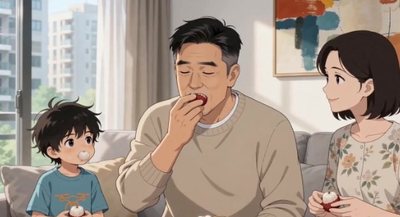} & \parfrm{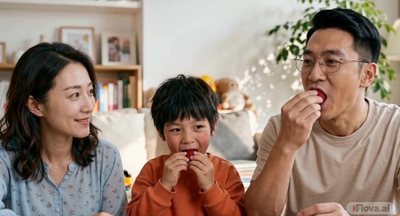} \\
{\tiny 94\%} & \parfrm{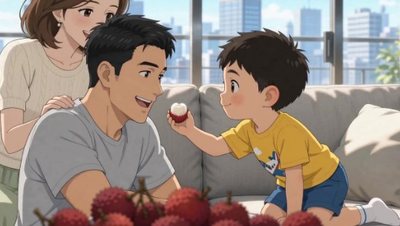} & \parfrm{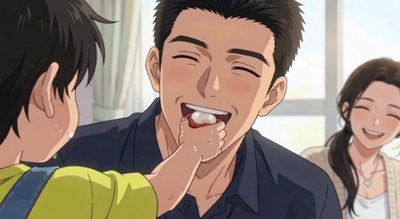} & \parfrm{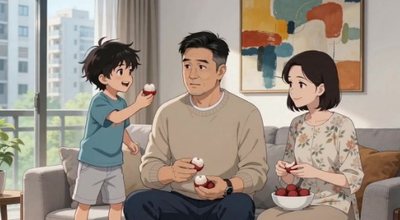} & \parfrm{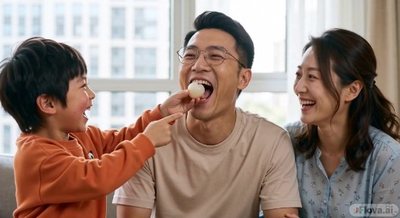} \\
\end{tabular}

\caption{\zhen{%
\pipe{} (\textbf{Ours}) and three commercial platforms process the same script of three scenes and
eight shots, fully automatically in a single pass without regeneration or manual editing. Each
system determines its own cut ($43$--$98$\,s), so rows align relative progress rather than timestamps.
Expert review rated all four in the same band for narrative pacing, overall consistency and cinematic
expressiveness, with differences in stylistic orientation. This single-script example under default
settings illustrates a practical production setting, not a controlled comparison, general platform
equivalence or ranking.}{%
\pipe{}（\textbf{Ours}）与三家商用平台处理同一份含三场、八个镜头的剧本，均全自动单遍运行，无重生成或人工剪辑。
各系统自行确定剪辑与成片时长（$43$--$98$ 秒），因此各行按相对进度而非时间戳对齐。
专家评审将四者的叙事节奏、整体一致性和影视表现力评为同一档，差异在于风格取向。
这一默认设置下的单剧本案例用于展示实际制作场景，不构成受控比较、平台普遍等效的证据或排名。}}
\label{fig:parity}
\end{figure}

\paragraph{Parity with commercial platforms.}
A benchmark generated by a weak pipeline measures the pipeline, not the models, so parity is required
of the chain, of the ways it can drive a stage, and of the output. The chain is the one those platforms
run from the script onward: storyboard, keyframe images, shot videos, then the short drama, fully
automatic throughout.
Platforms also differ in \emph{how} they drive the video stage, and \pipe{} implements four input
paradigms, differing in the artefact handed to the video model (\cref{fig:sourcing}): a first/last frame pair
(\texttt{fflf2v}), a sliced grid panel (\texttt{grid2v}), character, element and scene references with no
frames (\texttt{multiref2v}), or independently drawn keyframes (\texttt{keyframe2v}). The first three are
evaluated here. For the output we ran one identical script through \pipe{} and three commercial
short-drama platforms end to end, with no human editing, frame selection or best-of-$n$ reruns; expert
review found shot division, character and scene consistency, camera language and audio all landing in
the same band, with the differences stylistic. \Cref{fig:parity} shows the four outputs on the same script
of three scenes and eight shots; it illustrates a practical production setting rather than a
controlled platform comparison. \Cref{fig:exhibit} shows one drama's three episodes produced by
\pipe{}, sampled at equal intervals to illustrate the continuity required beyond individual shots.

\paragraph{Stage-by-stage comparability.}
A score difference between two models must come from the models. Every shot slot is generated once by
each participating model at every granularity, and the pipeline forks only at the
stage under test (\cref{fig:sourcing}): everything before the fork is generated once, so competing
models receive the identical upstream artefact and the same prompt string, and evaluating a new model costs
one stage of generation rather than a rebuilt corpus. Coverage gaps and duration-budget exceptions are listed
in \cref{sec:cov}.

\subsection{\labeler{}: annotation for information content and discriminative power}
\label{sec:annot}

\begin{figure}[htbp]
\centering
\includegraphics[width=\linewidth]{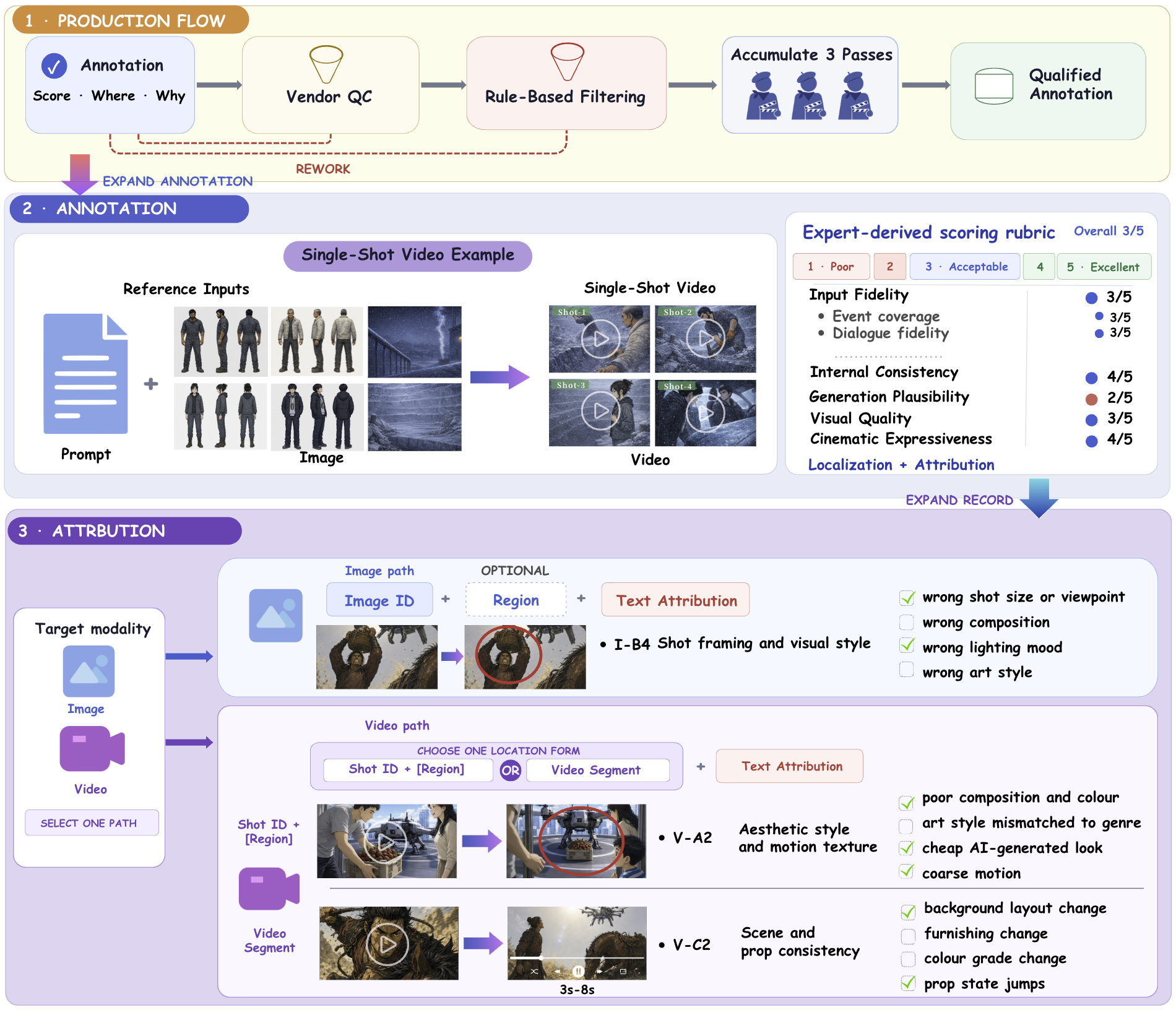}
\caption{\zhen{\labeler{} workflow. Each item is assigned three independent assessments using the
expert-derived rubric. Scores below full marks require a dimension-specific tag and explanation,
with spatial or temporal localisation where applicable. Annotations undergo vendor quality control
and rule-based filtering before entering the qualified corpus.}{\labeler{} 工作流程。每个条目分配三份基于专家评分细则的独立评估。
未获满分的维度须附专属标签和解释，并在适用时提供空间或时间定位。标注经过供应商质检和规则过滤后进入合格语料。}}
\label{fig:annotation}
\end{figure}

Comparable prior work constructs its human evaluation layer via small-sample scoring or method-level
comparisons, rather than item-wise exhaustive coverage; such approaches can only validate model-level
ranking outcomes. In contrast, \labeler{} leverages professional human annotators, item-wise
multi-dimensional scoring, and mandatory spatio-temporal attribution (\cref{fig:annotation}). The
full annotation collection yields $17{,}488$ scored entries and $255{,}925$ verifiable attributions,
with a median annotation time of $23$ minutes per item. The protocol targets independent scores from
three annotators per item; the reference set comprises the $5{,}569$ items with all three
assessments, and the collection-wide totals above are not restricted to it. Since point deductions are localised where applicable
and assigned an attribution, we can verify whether automated scoring
identifies the actual underlying defect, rather than merely producing a plausible final score.

\paragraph{Annotator profile.}
Our annotation pool comprises $543$ annotators. Recruitment adopts stage-specific skill screening:
annotators for storyboard design are required to have directing or screenwriting backgrounds, while
annotators for other stages need substantial experience in image or video annotation. Annotation tasks
are distributed across multiple supplier teams, each operating its own annotator cohort and dedicated
quality-control reviewers.

\paragraph{Spatio-temporal attribution.}
For any dimension that does not receive a perfect score, annotators must localise the issue where
applicable and select a corresponding reason from a pre-defined tag set for that dimension. The annotation interface
dynamically adapts to the task stage, supporting bounding-box frame selection, time intervals, single
or ranged shot indices, and free-form text input. This core constraint enables the analyses presented in the rest of this paper: it distinguishes the scenario where ``all three annotators dislike an
output'' from ``all three annotators pinpoint the same defect''; it defines a standardised output
format that the automated scorer must follow; and it automatically exposes unsupported, evidence-free
score deductions.

\paragraph{Quality control and filtering.}
\label{sec:filter}
Group-level reviewers sample a configurable fraction of their team's outputs (one-fifth by default).
Failed annotations are archived and reassigned for re-annotation. A quality dashboard further flags
low-quality submissions for revision based on nine heuristic rules, including annotation speed far
below the granularity-matched median, deductions with no associated attribution, and near-zero
correlation with leave-one-out consensus.

\paragraph{Consensus and human baseline.}
Human consensus is the mean of available annotator scores at both leaf and composite levels. To
contextualise automated agreement, the leave-one-out human baseline \hbase{} computes PLCC between
each annotator's score and the mean of the remaining annotators, pooling all pairs. \judger{} is
compared with the annotator mean on items with both human and automated scores (\cref{tab:agree}).

\subsection{\judger{}: multi-tool agentic scoring}
\label{sec:judge}

\paragraph{Baseline and hybrid pipelines.}
The baseline approach feeds input material and evaluation rubrics directly into a VLM to produce a
single JSON-formatted verdict. It suffers from a fundamental structural limitation: short-drama defects
often lie in local frame regions or inter-segment discontinuities, both invisible to holistic
impression-based assessment. Existing hybrid pipelines mitigate this by equipping large models with
measurements from specialist models~\citep{msvbench,entitybench,multiref}. Following this paradigm, we
integrate a full suite of modules for detection-segmentation~\citep{yolo26,sam2},
identity and general visual-similarity evidence~\citep{arcface,dinov3}, scene-consistency~\citep{dreamsim},
style-consistency~\citep{csd}, aesthetic-quality~\citep{musiq,nima,dover},
motion~\citep{raft}, audio
analysis~\citep{whisper,dnsmos,syncnet,imagebind} and cross-model corroboration with VideoPhy-AutoEval~\citep{videophy2}. We then
raise a largely overlooked question: do these metrics align with human judgements?

\paragraph{The metric audit.}
\label{sec:step2}
Most metrics correlate weakly with the human annotations, at $|\mathrm{SRCC}|$ no greater than $0.2$.
Degradation is severe for aesthetic and image-quality metrics: fourteen photography-oriented metrics
yield negative correlation, while AIGC-specific perceptual models remain below $0.2$. Their failure
stems from divergent priors: photography-domain metrics assume real-world photographs, whereas AIGC
models treat short-drama intrinsic stylisation as anomalous deviation (\cref{sec:audit}). This
benchmark further invalidates three leaf dimensions including \dc{V-D2}, whose human baseline is
near-random. Aligning the automated scorer against random-like references yields uninterpretable outputs.

\begin{figure}[p]
\centering
\includegraphics[height=0.90\textheight]{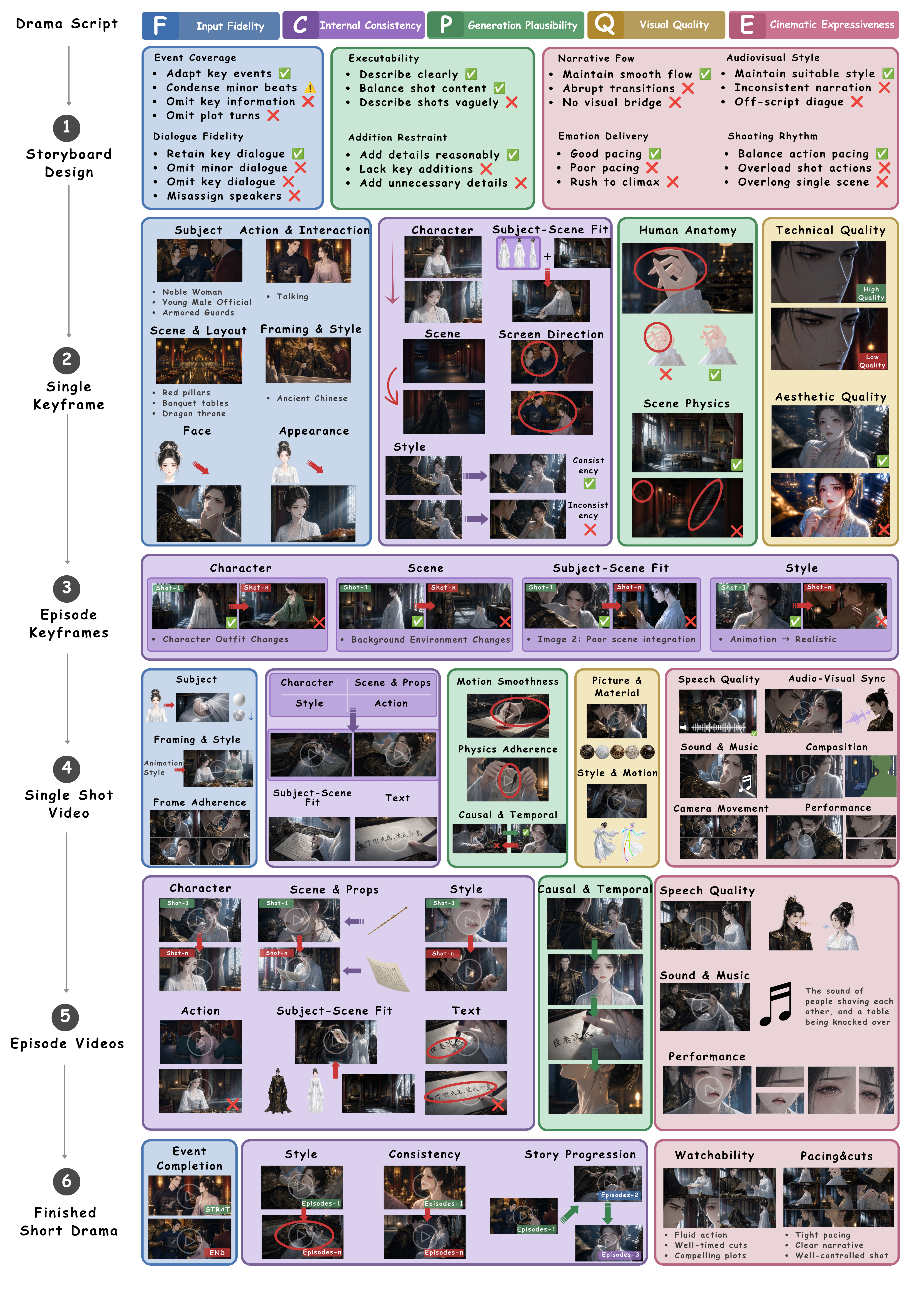}
\caption{\zhen{The overview of \judger{}. Rows are the six granularities in production order and
columns the five evaluation axes; every cell shows the checks run there and the evidence they are decided
on, taken from the annotators' own criteria and tag vocabulary.}{\judger{} 概览。
行是按生产顺序排列的六个粒度，列是五个评测轴；每格给出该处运行的核查项与它们所依据的证据，
判据与归因词表都取自标注员自己使用的那一套。}}
\label{fig:judgemap}
\end{figure}

\paragraph{Multi-tool agentic scoring.}
Against this backdrop, \judger{} retains only metrics with non-negligible correlation that supply
low-level perceptual signals unavailable to LLMs, using them as auxiliary cues rather than direct
scoring inputs. Each leaf dimension is scored independently and concurrently over multiple
agentic rounds. The system automatically routes relevant measurements for each target dimension; the
agent may additionally invoke perception tools (regional zoom, timestamp-based frame extraction)
for fine-grained inspection before scoring against item-specific checklists. Output verdicts include
annotator-derived defect tags and spatio-temporal localisation, enabling direct cross-checking whenever
model-human disagreements arise. \Cref{fig:judgemap} lays out the checks run at every granularity,
and \cref{fig:judge} presents a field-by-field verdict example.

\input{figures/fig_judge}

\paragraph{Deployed models and settings.}
\label{sec:judgeconfig}
{\sloppy We adopt stage-specific scoring configurations targeted at respective bottlenecks. Storyboard design
averages outputs from \md{gpt-5.6-sol-xhigh} and a \md{claude-fable-5} variant to reduce
single-family dependence. Keyframe evaluation uses \md{doubao-seed-2.1-pro} as the primary judging
model, and shot-level video evaluation uses \md{gemini-3.1-pro}.\par}

\section{Experiment}
\label{sec:results}

\subsection{Setup}
\label{sec:setup}

Nine text LLMs, seven image models and six video models were scored by both human annotation and
automated evaluation, and fifteen further models (eleven text, one image and three video) by
automated evaluation alone on the same items: the $\dagger$ rows of \cref{tab:main}. \md{wan-3.0} and
\md{seedance-2.5} ran under a longer duration budget with replanned storyboards, so their units
match by drama, episode, paradigm and shot index (\cref{sec:cov}). Scores are out of $5$, and \textbf{\boldmath $3.0$ is the line
between usable and needing rework}. Model names are the vendors' own; a trailing \texttt{-xhigh},
\texttt{-max} or \texttt{-high} is the reasoning-effort setting the run used rather than part of the
name. Model-side content review and generation failures left several
models short of the full item set; those models are scored on the slice they ran and are marked
$\ddagger$ in \cref{tab:main}; their gaps are listed in \cref{sec:cov}. The board, model rankings and
stage-level descriptions use automated scores; human scores are marked as such.

\subsection{Main results}
\label{sec:mainresults}
\label{sec:leaderboard}
\label{sec:quant}

\begin{table*}[htbp]
\centering
\caption{\zhen{\textbf{Main results (automated scores).} $5$-point scores over six stages and five axes;
\emph{All} averages all applicable leaf dimensions, weighting axes by leaf count. \fail{Red}: below the
$3.0$ usable line; \best{bold}: top annotated model per column; dash: axis undefined for the stage.
Stage headers give model-level PLCC/SRCC against the three-annotator human board. $\dagger$: scored
automatically only, excluded from agreement metrics and bold marking; $\ddagger$: partial corpus,
scored on evaluated subsets (\cref{sec:cov}).}{\textbf{主要结果（自动化评分）。}六个环节、五条评测轴上的 $5$ 分制得分；
\emph{All} 为全部适用叶子维度的均值，各轴按叶子维度数加权。\fail{红色}：低于 $3.0$ 可用线；\best{加粗}：该列中有人工标注模型的最优值；
短横：该环节未定义此轴。环节标题处给出与三人共识人工榜的模型级 PLCC/SRCC。$\dagger$：仅由自动化评测打分，
不计入一致性统计与最优标记；$\ddagger$：只跑了部分语料，按实际评测的子集计分（\cref{sec:cov}）。}}
\label{tab:main}
\footnotesize
\renewcommand{\arraystretch}{0.80}
\setlength{\aboverulesep}{1pt}\setlength{\belowrulesep}{1pt}
\begin{adjustbox}{max totalheight=0.87\textheight,center}
\begin{tabular*}{\linewidth}{@{\extracolsep{\fill}}l >{\columncolor{axF!8}}c >{\columncolor{axC!8}}c >{\columncolor{axP!8}}c >{\columncolor{axQ!8}}c >{\columncolor{axE!8}}c c@{}}
\toprule
\textbf{Model} & \makecell{\axF{}\\\textcolor{axF!80!black}{Input}\\\textcolor{axF!80!black}{fidelity}} & \makecell{\axC{}\\\textcolor{axC!80!black}{Internal}\\\textcolor{axC!80!black}{consistency}} & \makecell{\axP{}\\\textcolor{axP!80!black}{Generation}\\\textcolor{axP!80!black}{plausibility}} & \makecell{\axQ{}\\\textcolor{axQ!80!black}{Visual}\\\textcolor{axQ!80!black}{quality}} & \makecell{\axE{}\\\textcolor{axE!80!black}{Cinematic}\\\textcolor{axE!80!black}{expressiveness}} & \textbf{All} \\
\midrule
\multicolumn{7}{@{}l}{\textbf{\textcircled{1} Storyboard design}\ \ \scriptsize text $\to$ text $\cdot$ PLCC $0.936$ $\cdot$ SRCC $0.800$} \\
\md{gpt-5.5-xhigh} & \best{4.84} & \textcolor{rulegray}{--} & \best{3.89} & \textcolor{rulegray}{--} & 4.10 & \best{4.23} \\
\md{claude-opus-4.8-max} & 4.51 & \textcolor{rulegray}{--} & 3.80 & \textcolor{rulegray}{--} & \best{4.19} & 4.17 \\
\md{hy3} & 4.36 & \textcolor{rulegray}{--} & 3.65 & \textcolor{rulegray}{--} & 3.78 & 3.89 \\
\md{doubao-seed-2.1-pro} & 4.26 & \textcolor{rulegray}{--} & 3.65 & \textcolor{rulegray}{--} & 3.82 & 3.89 \\
\md{glm-5.2} & 4.01 & \textcolor{rulegray}{--} & 3.66 & \textcolor{rulegray}{--} & 3.74 & 3.79 \\
\md{gemini-3.1-pro} & 4.08 & \textcolor{rulegray}{--} & 3.43 & \textcolor{rulegray}{--} & 3.67 & 3.71 \\
\md{kimi-k2.6} & 3.66 & \textcolor{rulegray}{--} & 3.42 & \textcolor{rulegray}{--} & 3.74 & 3.64 \\
\md{qwen3.7-max} & 3.24 & \textcolor{rulegray}{--} & 3.48 & \textcolor{rulegray}{--} & 3.59 & 3.48 \\
\md{mimo-v2.5-pro} & \fail{2.68} & \textcolor{rulegray}{--} & 3.53 & \textcolor{rulegray}{--} & 3.41 & 3.26 \\
\hdashline[0.4pt/1.6pt]
\md{gpt-6-astra-max}$^{\dagger}$ & 4.90 & \textcolor{rulegray}{--} & 4.25 & \textcolor{rulegray}{--} & 4.28 & 4.43 \\
\md{gpt-5.6-sol-max}$^{\dagger}$ & 4.95 & \textcolor{rulegray}{--} & 4.02 & \textcolor{rulegray}{--} & 4.31 & 4.40 \\
\md{glm-5.3-max}$^{\dagger}$ & 4.78 & \textcolor{rulegray}{--} & 3.88 & \textcolor{rulegray}{--} & 4.20 & 4.26 \\
\md{claude-fable-5-max}$^{\dagger}$ & 4.55 & \textcolor{rulegray}{--} & 3.99 & \textcolor{rulegray}{--} & 4.20 & 4.24 \\
\md{claude-opus-5-max}$^{\dagger}$ & 4.45 & \textcolor{rulegray}{--} & 3.85 & \textcolor{rulegray}{--} & 4.30 & 4.22 \\
\md{kimi-k3}$^{\dagger}$ & 4.67 & \textcolor{rulegray}{--} & 3.87 & \textcolor{rulegray}{--} & 4.17 & 4.22 \\
\md{qwen3.8-max}$^{\dagger}$ & 4.55 & \textcolor{rulegray}{--} & 3.88 & \textcolor{rulegray}{--} & 3.94 & 4.08 \\
\md{hy4-preview}$^{\dagger}$ & 4.69 & \textcolor{rulegray}{--} & 3.56 & \textcolor{rulegray}{--} & 3.91 & 4.02 \\
\md{grok-4.5}$^{\dagger}$ & 3.96 & \textcolor{rulegray}{--} & 3.85 & \textcolor{rulegray}{--} & 3.92 & 3.92 \\
\md{gemini-3.8-flash-high}$^{\dagger}$ & 4.18 & \textcolor{rulegray}{--} & 3.42 & \textcolor{rulegray}{--} & 3.79 & 3.80 \\
\md{gemini-3.6-flash-high}$^{\dagger}$ & 4.04 & \textcolor{rulegray}{--} & 3.65 & \textcolor{rulegray}{--} & 3.71 & 3.78 \\
\midrule
\multicolumn{7}{@{}l}{\textbf{\textcircled{2} Single keyframe}\ \ \scriptsize text $+$ refs $\to$ pixels $\cdot$ PLCC $0.959$ $\cdot$ SRCC $0.750$} \\
\md{gpt-image-2} & \best{3.65} & \best{4.32} & \best{4.47} & \best{4.13} & \textcolor{rulegray}{--} & \best{3.99} \\
\md{gpt-image-1.5} & 3.25 & 3.96 & 4.46 & 3.90 & \textcolor{rulegray}{--} & 3.72 \\
\md{nano-banana-2} & 3.33 & 3.92 & 4.25 & 3.71 & \textcolor{rulegray}{--} & 3.68 \\
\md{seedream-5.0-pro} & 3.56 & 3.92 & 3.74 & 3.52 & \textcolor{rulegray}{--} & 3.64 \\
\md{nano-banana-pro} & 3.17 & 3.89 & 4.34 & 3.79 & \textcolor{rulegray}{--} & 3.62 \\
\md{wan2.7-image-pro} & \fail{2.95} & 3.63 & 4.03 & 3.46 & \textcolor{rulegray}{--} & 3.38 \\
\md{seedream-5.0-lite}$^{\ddagger}$ & \fail{2.96} & 3.40 & 3.89 & 3.40 & \textcolor{rulegray}{--} & 3.32 \\
\hdashline[0.4pt/1.6pt]
\md{gpt-image-2.5}$^{\dagger}$ & 3.83 & 4.46 & 4.64 & 4.23 & \textcolor{rulegray}{--} & 4.12 \\
\midrule
\multicolumn{7}{@{}l}{\textbf{\textcircled{3} Episode keyframes}\ \ \scriptsize a whole episode of frames $\cdot$ PLCC $0.973$ $\cdot$ SRCC $0.964$} \\
\md{gpt-image-2} & \textcolor{rulegray}{--} & \best{3.82} & \textcolor{rulegray}{--} & \textcolor{rulegray}{--} & \textcolor{rulegray}{--} & \best{3.82} \\
\md{gpt-image-1.5} & \textcolor{rulegray}{--} & 3.60 & \textcolor{rulegray}{--} & \textcolor{rulegray}{--} & \textcolor{rulegray}{--} & 3.60 \\
\md{nano-banana-2} & \textcolor{rulegray}{--} & 3.47 & \textcolor{rulegray}{--} & \textcolor{rulegray}{--} & \textcolor{rulegray}{--} & 3.47 \\
\md{seedream-5.0-pro} & \textcolor{rulegray}{--} & 3.28 & \textcolor{rulegray}{--} & \textcolor{rulegray}{--} & \textcolor{rulegray}{--} & 3.28 \\
\md{nano-banana-pro} & \textcolor{rulegray}{--} & 3.04 & \textcolor{rulegray}{--} & \textcolor{rulegray}{--} & \textcolor{rulegray}{--} & 3.04 \\
\md{wan2.7-image-pro} & \textcolor{rulegray}{--} & 3.00 & \textcolor{rulegray}{--} & \textcolor{rulegray}{--} & \textcolor{rulegray}{--} & 3.00 \\
\md{seedream-5.0-lite}$^{\ddagger}$ & \textcolor{rulegray}{--} & \fail{2.91} & \textcolor{rulegray}{--} & \textcolor{rulegray}{--} & \textcolor{rulegray}{--} & \fail{2.91} \\
\hdashline[0.4pt/1.6pt]
\md{gpt-image-2.5}$^{\dagger}$ & \textcolor{rulegray}{--} & 4.00 & \textcolor{rulegray}{--} & \textcolor{rulegray}{--} & \textcolor{rulegray}{--} & 4.00 \\
\midrule
\multicolumn{7}{@{}l}{\textbf{\textcircled{4} Shot video}\ \ \scriptsize pixels $\to$ time $+$ audio $\cdot$ PLCC $0.755$ $\cdot$ SRCC $0.829$} \\
\md{seedance-2.0}$^{\ddagger}$ & 3.44 & \best{3.27} & 3.06 & \best{3.77} & 3.05 & \best{3.24} \\
\md{happyhorse-1.1} & 3.34 & 3.21 & 3.04 & 3.43 & 3.08 & 3.18 \\
\md{kling-3.0-omni} & 3.22 & 3.23 & \best{3.08} & 3.28 & \fail{2.88} & 3.11 \\
\md{pixverse-c1} & 3.07 & 3.09 & \fail{2.92} & 3.42 & \best{3.10} & 3.10 \\
\md{wan2.7}$^{\ddagger}$ & \best{3.45} & \fail{2.96} & \fail{2.87} & 3.22 & \fail{2.92} & 3.04 \\
\md{veo-3.1}$^{\ddagger}$ & 3.17 & \fail{2.77} & \fail{2.73} & 3.32 & \fail{2.76} & \fail{2.88} \\
\hdashline[0.4pt/1.6pt]
\md{seedance-2.5}$^{\dagger\ddagger}$ & 3.24 & 3.40 & \fail{2.97} & 3.71 & \fail{2.85} & 3.17 \\
\md{minimax-h3}$^{\dagger}$ & 3.43 & 3.20 & \fail{2.83} & 3.46 & 3.11 & 3.17 \\
\md{wan-3.0}$^{\dagger}$ & 3.51 & 3.26 & \fail{2.74} & 3.42 & \fail{2.97} & 3.15 \\
\midrule
\multicolumn{7}{@{}l}{\textbf{\textcircled{5} Episode video}\ \ \scriptsize shots in sequence $\cdot$ PLCC $0.935$ $\cdot$ SRCC $0.943$} \\
\md{seedance-2.0}$^{\ddagger}$ & \textcolor{rulegray}{--} & \best{\fail{2.96}} & 3.41 & \textcolor{rulegray}{--} & 3.32 & \best{3.11} \\
\md{happyhorse-1.1} & \textcolor{rulegray}{--} & \fail{2.84} & \best{3.64} & \textcolor{rulegray}{--} & 3.37 & 3.08 \\
\md{kling-3.0-omni} & \textcolor{rulegray}{--} & \fail{2.79} & 3.31 & \textcolor{rulegray}{--} & \best{3.48} & 3.06 \\
\md{wan2.7}$^{\ddagger}$ & \textcolor{rulegray}{--} & \fail{2.57} & 3.24 & \textcolor{rulegray}{--} & 3.35 & \fail{2.87} \\
\md{veo-3.1}$^{\ddagger}$ & \textcolor{rulegray}{--} & \fail{2.47} & 3.27 & \textcolor{rulegray}{--} & 3.18 & \fail{2.76} \\
\md{pixverse-c1} & \textcolor{rulegray}{--} & \fail{2.42} & 3.08 & \textcolor{rulegray}{--} & 3.16 & \fail{2.71} \\
\hdashline[0.4pt/1.6pt]
\md{seedance-2.5}$^{\dagger\ddagger}$ & \textcolor{rulegray}{--} & 3.16 & 3.11 & \textcolor{rulegray}{--} & 3.50 & 3.25 \\
\md{wan-3.0}$^{\dagger}$ & \textcolor{rulegray}{--} & 3.12 & 3.70 & \textcolor{rulegray}{--} & 3.42 & 3.25 \\
\md{minimax-h3}$^{\dagger}$ & \textcolor{rulegray}{--} & \fail{2.87} & 3.17 & \textcolor{rulegray}{--} & 3.27 & \fail{2.99} \\
\midrule
\multicolumn{7}{@{}l}{\textbf{\textcircled{6} Short drama}\ \ \scriptsize the finished short drama $\cdot$ PLCC $0.947$ $\cdot$ SRCC $1.000$} \\
\md{seedance-2.0}$^{\ddagger}$ & \best{4.29} & 3.38 & \textcolor{rulegray}{--} & \textcolor{rulegray}{--} & \fail{2.68} & \best{3.30} \\
\md{happyhorse-1.1} & 3.82 & 3.38 & \textcolor{rulegray}{--} & \textcolor{rulegray}{--} & \best{\fail{2.69}} & 3.22 \\
\md{kling-3.0-omni} & 3.37 & \best{3.41} & \textcolor{rulegray}{--} & \textcolor{rulegray}{--} & \fail{2.54} & 3.11 \\
\md{wan2.7}$^{\ddagger}$ & 4.12 & \fail{2.80} & \textcolor{rulegray}{--} & \textcolor{rulegray}{--} & \fail{2.59} & \fail{2.95} \\
\md{pixverse-c1} & 3.05 & \fail{2.88} & \textcolor{rulegray}{--} & \textcolor{rulegray}{--} & \fail{2.58} & \fail{2.81} \\
\md{veo-3.1}$^{\ddagger}$ & \fail{2.67} & 3.07 & \textcolor{rulegray}{--} & \textcolor{rulegray}{--} & \fail{2.44} & \fail{2.79} \\
\hdashline[0.4pt/1.6pt]
\md{seedance-2.5}$^{\dagger\ddagger}$ & 4.40 & 3.93 & \textcolor{rulegray}{--} & \textcolor{rulegray}{--} & 3.00 & 3.70 \\
\md{wan-3.0}$^{\dagger}$ & 4.23 & 3.56 & \textcolor{rulegray}{--} & \textcolor{rulegray}{--} & \fail{2.74} & 3.47 \\
\md{minimax-h3}$^{\dagger}$ & 3.77 & 3.33 & \textcolor{rulegray}{--} & \textcolor{rulegray}{--} & \fail{2.51} & 3.20 \\
\bottomrule
\end{tabular*}
\end{adjustbox}
\end{table*}

\Cref{tab:main} is the automated board for all six stages, broken out over the five evaluation axes.

\paragraph{The strongest chain at annotation time.}
\label{sec:sota}
Among models covered by human annotation, the strongest chain by automated score uses \md{gpt-5.5-xhigh}
for storyboarding, \md{gpt-image-2} for keyframes, and \md{seedance-2.0} for video. Within each
modality, the leading annotated model is invariant to evaluation granularity: \md{gpt-image-2} tops
both keyframe stages ($3.99$, $3.82$), and \md{seedance-2.0} leads all three video stages ($3.24$,
$3.11$, $3.30$). This chain operates near the usable line. Its finished short drama receives an automated score of $3.30$
(only $0.3$ above the line), and its best shot-video result is $3.24$, so the strongest annotated
chain meets the line with little margin. In the expanded set, \md{gpt-6-astra-max} leads
storyboards and \md{gpt-image-2.5} both keyframe stages; \md{seedance-2.0} still leads shot
video, while \md{seedance-2.5} has the highest automated finished-drama score ($3.70$) and shares the
highest episode-video score ($3.25$) with \md{wan-3.0}.

\paragraph{Performance degrades along the text-to-video pipeline.}
\label{sec:collapse}
Mean automated scores over the human-annotated models decay progressively through the pipeline: $3.78$ for
storyboard design, $3.62$ for single
keyframes, $3.30$ for episode keyframes, $3.09$ for shot video, $2.93$ for episode
video, with a modest rebound to $3.03$ for the finished short drama. No pixel- or video-generation
stage attains the mean of $3.78$ achieved at the text stage; the three video granularities all sit
within $0.1$ of the usable line, with episode video below it. Two factors are consistent with this
degradation. First, generation difficulty rises as
outputs acquire pixels, temporal dynamics and audio, applying the same evaluation lens to
progressively harder downstream outputs. Second, defects propagate rather than remain localised: each
stage consumes outputs from its predecessor, with no component empowered to fix inherited flaws
(\cref{sec:prop}).

\paragraph{Benchmark discrimination is strongest for criteria with explicit reference anchors.}
\label{sec:gaps}
Averaged across applicable stages, top-to-bottom gaps in automated scores are $1.22$ for \axF{} input
fidelity, $0.70$ for \axC{} internal consistency, $0.64$ for \axQ{} visual quality, $0.53$ for \axP{}
generation plausibility, and $0.42$ for \axE{} cinematic expressiveness. This ordering is consistent
with reference explicitness: \axF{} is validated against upstream outputs and \axC{} against peer
artefacts, whereas \axP{}, \axQ{} and \axE{} lack external references. Automated benchmarks appear
most discriminative when anchors are well-defined, so small gaps on weakly anchored metrics may
reflect benchmark limitations rather than comparable model performance. Human annotation supports
the diagnosis but not the ordering: it too separates models
most on \axF{}, while ranking \axE{} second widest where this board ranks it last.

\paragraph{Models exhibit heterogeneous strengths across dimensions.}
\label{sec:profiles}
Aggregate composite scores mask dimensional trade-offs. For single keyframes, \md{seedream-5.0-pro} and
\md{nano-banana-pro} differ by only $0.02$ in automated composite score. Yet \md{seedream-5.0-pro} leads by $0.39$
on \axF{} input fidelity but lags by $0.60$ on \axP{} generation plausibility. For shot video,
\md{kling-3.0-omni} and \md{pixverse-c1} differ by merely $0.01$ overall; \md{kling-3.0-omni} is $0.16$
higher on \axP{} but $0.22$ lower on \axE{} cinematic expressiveness. In both pairs, per-dimension
gaps are an order of magnitude larger than composite gaps. Composite scores are thus suitable for
tier grouping yet misleading for model selection: one should ask ``better on which
dimension'' instead of ``which model is better''. Concrete cases demonstrating this effect are
provided in \cref{app:cases}.

\subsection{Agreement validation}
\label{sec:align}

\label{sec:meta}
\label{sec:meta-agree}
\label{sec:meta-axis}
\label{sec:meta-rank}
\label{sec:meta-fail}
\label{sec:hbdef}

\begin{table}[htbp]
\centering
\caption{\zhen{Agreement with human ratings by stage. \textbf{Model Level}: model-score correlations;
\textbf{Item Level}: per-item correlation and the leave-one-out human baseline (\hbase);
\textbf{Score Acc}: bias, MAE and accuracy within tolerance. \emph{Paired items} have human and
automated scores: all collected text and image items, $1{,}149$ of $2{,}186$ shot videos, $238$ of
$306$ episode videos and $201$ of $206$ short dramas. Counts describe collection and pairing,
not the complete reference set.}{分阶段与人工评分的一致性。\textbf{Model Level}：模型级分数相关性；
\textbf{Item Level}：逐条目相关性及留一法人工基线（\hbase）；\textbf{Score Acc}：偏差、MAE及容差内准确率。
\emph{Paired items} 同时具有人评与自动评分：全部已采集文本／图像条目、$2{,}186$ 条镜头视频中的 $1{,}149$ 条、
$306$ 条整集视频中的 $238$ 条，以及 $206$ 部短剧中的 $201$ 部。这些数量描述采集与配对分析，而非完整参考集。}}
\label{tab:agree}
\small
\setlength{\tabcolsep}{4.2pt}
\begin{tabular}{@{}l c cc c >{\columncolor{softrow}}c >{\columncolor{softrow}}c >{\columncolor{softrow}}c c ccc@{}}
\toprule
& & \multicolumn{2}{c}{\textbf{Model Level}} & &
\multicolumn{3}{c}{\textbf{Item Level}} & &
\multicolumn{3}{c}{\textbf{Score Acc}} \\
\cmidrule(lr){3-4}\cmidrule(lr){6-8}\cmidrule(lr){10-12}
Stage & \makecell{Paired\\items} & PLCC & SRCC & &
\hbase{} & Judge & Ratio & & Bias & MAE & Within $\pm0.5$ \\
\midrule
\textcircled{1} Storyboard design & 533   & 0.936 & 0.800 & & 0.217 & \pass{0.509} & \pass{$2.35\times$} & & $+0.28$ & 0.416 & 67.5\% \\
\textcircled{2} Single keyframe   & 2{,}125 & 0.959 & 0.750 & & 0.449 & \pass{0.536} & \pass{$1.19\times$} & & $+0.41$ & 0.529 & 54.2\% \\
\textcircled{3} Episode keyframes & 429   & \best{0.973} & 0.964 & & 0.233 & \pass{0.344} & \pass{$1.48\times$} & & $+0.57$ & 0.733 & 38.2\% \\
\textcircled{4} Shot video & 1{,}149 & 0.755 & 0.829 & & 0.380 & \fail{0.301} & \fail{$0.79\times$} & & $-0.35$ & 0.436 & 62.5\% \\
\textcircled{5} Episode video  & 238   & 0.935 & 0.943 & & 0.240 & \fail{0.203} & \fail{$0.85\times$} & & $+0.14$ & 0.433 & 66.0\% \\
\textcircled{6} Short drama  & 201   & 0.947 & \best{1.000} & & 0.450 & \fail{0.384} & \fail{$0.85\times$} & & $+0.42$ & 0.425 & 66.7\% \\
\midrule
\addlinespace[1pt]
\textbf{Mean} & & \textbf{0.918} & \textbf{0.881} & & 0.328 & 0.379 & & & & 0.495 & \\
\bottomrule
\end{tabular}
\tabnote{\zhen{Correlations use composite scores. \emph{Judge} is per-item PLCC with the
three-annotator consensus; \hbase{} is the leave-one-out human baseline, and \emph{Ratio} is
Judge\,/\,\hbase. Green meets or exceeds the baseline; red falls below it.}{相关性基于综合分。
\emph{Judge} 为自动评分与三人共识的逐条目PLCC，\hbase{} 为留一法人工基线，\emph{Ratio} 为 Judge\,/\,\hbase。
绿色表示达到或超过基线，红色表示低于基线。}}
\end{table}

\paragraph{Agreement against the human reference.}
All $5{,}569$ items in the reference set carry three independent annotations, establishing dual-level
reference benchmarks for subsequent correlation analysis. At the model level, the reference standard
is the ranking derived from three-annotator consensus, with correlation calculated directly against
this human ranking. At the item level, evaluation relies on consensus scores and adopts the
leave-one-out human baseline (\hbase, a rotated and pooled paradigm where one annotator predicts
the average score of the other two) to eliminate invalid absolute correlation results without
criterion reproducibility. Detailed derivation and illustrative cases are provided in
\cref{app:agreeslices}. Cross-level agreement analysis demonstrates robust model-level consistency:
\judger{} achieves a mean PLCC of $0.918$ across six evaluation stages, with perfect alignment
between automated and human rankings (SRCC $= 1.000$) for short dramas. The fifteen additional models
in \cref{tab:main} rely on this model-level agreement: they are scored automatically and are not
themselves checked against human ratings. At the individual item level, agreement varies distinctly
across pipeline stages. Text and image stages reach or exceed the human baseline; storyboard design
reaches $2.35\times$ the baseline, although several storyboard baselines are themselves low
(\cref{sec:storyboard}). In contrast, the three video stages fall to $0.79$--$0.85\times$ the
baseline: correlation remains positive, but per-item agreement stays below the agreement between
annotators.

\paragraph{Determinants of agreement.}
The stage-wise divergence of automated-human agreement tracks two attributes of the evaluation
tasks rather than criterion subjectivity. First, evaluation modality shapes
consistency performance. The critical performance divide lies between static image and dynamic video
modalities, rather than between single and multi-asset tasks (\cref{fig:ratio}). Despite requiring
more cognitively demanding cross-frame comparison, episode keyframe scoring achieves
$1.48\times$ baseline consistency; in contrast, isolated single video clip evaluation only reaches
$0.79\times$ the baseline. Modality-aggregated results show a steady consistency decay for automated
scoring: $0.509$ for text, $0.440$ for pixel-based images, and $0.296$ for time-and-audio video
content. Conversely, human annotator consistency rises across these modalities ($0.217$, $0.341$,
$0.357$), revealing an inverse trend where the automated scorer performs worst in scenarios with the
highest human consensus. Second, the availability of external validation references governs the
hierarchical gap across evaluation axes (\cref{tab:axes}). Agreement scores rank in the order of
\axF{} ($0.423$), \axC{} ($0.235$), \axP{} ($0.146$), \axE{} ($0.132$), and \axQ{} ($0.121$). Only
\axF{} delivers a net improvement over the human baseline ($+0.107$, superior in $9$ out of $12$
tests), while \axQ{} shows no baseline advantage in any of its four test cases.

\begin{figure}[htbp]
\centering
\begin{adjustbox}{max width=\linewidth}
\begin{tikzpicture}
\begin{axis}[
  dcbar, width=14cm, height=5.6cm,
  bar width=22pt,
  xtick={1,2,3,4,5,6},
  xticklabels={%
    \shortstack{\textcircled{1}\\storyboard},
    \shortstack{\textcircled{2}\\single kf.},
    \shortstack{\textcircled{3}\\episode kf.},
    \shortstack{\textcircled{4}\\shot video},
    \shortstack{\textcircled{5}\\episode video},
    \shortstack{\textcircled{6}\\short drama}},
  xticklabel style={font=\scriptsize, align=center},
  ymin=0, ymax=3.05, ytick={0,0.5,1.0,1.5,2.0,2.5},
  ylabel={per-item agreement $\div$ human baseline},
  nodes near coords, point meta=explicit symbolic,
  every node near coord/.append style={font=\scriptsize\bfseries, yshift=1pt},
  xmin=0.45, xmax=6.55,
]

\addplot[draw=tierone!70, fill=tierone!35, bar shift=0pt]
  coordinates {(1,2.35) [2.35$\times$] (2,1.19) [1.19$\times$] (3,1.48) [1.48$\times$]};
\addplot[draw=tierfour!70, fill=tierfour!30, bar shift=0pt,
         every node near coord/.append style={anchor=north, yshift=-3.5pt,
                                              font=\scriptsize\bfseries,
                                              text=tierfour!80!black}]
  coordinates {(4,0.79) [0.79$\times$] (5,0.85) [0.85$\times$] (6,0.85) [0.85$\times$]};

\draw[black, dashed, thick] (axis cs:0.45,1) -- (axis cs:6.55,1);
\node[anchor=north east, font=\scriptsize\itshape, align=right]
  at (axis cs:6.5,2.95)
  {$1.0$ = human baseline};

\node[anchor=south, font=\scriptsize, text=tierone!75!black]
  at (axis cs:2,2.62) {text and pixels: above the baseline};
\node[anchor=south, font=\scriptsize, text=tierfour!80!black]
  at (axis cs:5,1.30) {time and audio: below the baseline};

\end{axis}
\end{tikzpicture}
\end{adjustbox}
\caption{\zhen{Automated per-item agreement as a multiple of the leave-one-out human baseline,
using PLCC. Values above $1$ indicate agreement exceeding the baseline.}{自动评分的逐条目一致性相对于留一法人工基线的倍数，使用PLCC衡量。
高于 $1$ 表示超过该基线。}}
\label{fig:ratio}
\end{figure}

\paragraph{Predictors of automated agreement.}
\label{sec:quadrant}
One might expect human--machine agreement to degrade as criteria become more subjective, yet this
study observes the opposite pattern: subjectivity shows no positive correlation with automated
scoring error. The most subjective dimension (audiovisual style treatment) reaches $6.5\times$ the
baseline, a ratio inflated by a near-zero human baseline, while the weakest-performing dimensions
(image quality, motion smoothness, environmental physics, audio-visual sync) are all objective
metrics. Two task-level attributes track agreement more closely than subjectivity: whether criteria
can be decomposed into enumerable check items, and whether judgement requires fine-grained
perceptual discrimination (\cref{tab:quadrant}).

\begin{table}[htbp]
\centering
\caption{\zhen{Dimensions grouped by perceptual requirements (rows) and whether criteria can be
expressed as checklists (columns). Cells report automated\,/\,human-baseline per-item PLCC, their
ratio, and implications for automation.}{按感知要求（行）及能否转化为检查清单（列）对维度分组。
单元格报告自动评分\,/\,人工基线的逐条目PLCC、二者比值及自动化建议。}}
\label{tab:quadrant}
\footnotesize
\setlength{\tabcolsep}{6pt}
\renewcommand{\arraystretch}{1.15}
\begin{tabularx}{\linewidth}{@{}l
  >{\raggedright\arraybackslash}X
  >{\raggedright\arraybackslash}X@{}}
\toprule
& \textbf{Criterion enumerable as check items} & \textbf{Criterion not enumerable} \\
\midrule
\makecell[tl]{\textbf{Perception}\\\textbf{not needed}} &
\cellcolor{goodcell}\textbf{Above the baseline.}\newline
{\scriptsize\setlength{\tabcolsep}{3pt}
\adjustbox{max width=\linewidth}{%
\begin{tabular}{@{}l r l@{}}
\dc{S-A2} dialogue fidelity & $0.728$/$0.442$ & $1.6\times$\\
\dc{S-A1} event coverage & $0.497$/$0.150$ & $3.3\times$\\
\dc{S-C4} audiovisual style & $0.265$/$0.041$ & $6.5\times^{\ast}$\\
\end{tabular}}}\newline
\emph{An external answer exists to compare against: the event list, the verbatim dialogue, the
style instruction.} &
\cellcolor{softrow}\textbf{Both sides near zero: repair the criterion.}\newline
{\scriptsize\setlength{\tabcolsep}{3pt}
\adjustbox{max width=\linewidth}{%
\begin{tabular}{@{}l r l@{}}
\dc{S-B2} restraint in additions & $0.110$/$-0.023$ & ---\\
\dc{I-D5} style consistency & $0.130$/$0.131$ & $0.99\times$\\
\end{tabular}}}\newline
\emph{The obstacle is the operability of the criterion, not annotation cost; tuning prompts here
yields little.} \\
\addlinespace[4pt]
\makecell[tl]{\textbf{Perception}\\\textbf{needed}} &
\cellcolor{badcell}\textbf{Below the baseline: add perception.}\newline
{\scriptsize\setlength{\tabcolsep}{3pt}
\adjustbox{max width=\linewidth}{%
\begin{tabular}{@{}l r l@{}}
\dc{V-E2} audio-visual sync & $0.075$/$0.448$ & $0.17\times$\\
\dc{V-D1} motion smoothness & $0.069$/$0.208$ & $0.33\times$\\
\dc{V-A1} image quality & $0.076$/$0.191$ & $0.40\times$\\
\dc{I-E2} environmental physics & $0.152$/$0.234$ & $0.65\times$\\
\end{tabular}}}\newline
\emph{All perceptual, and the human baseline is itself normal or high: the problem is real and
people can see it, the scorer cannot.} &
\emph{\textcolor{rulegray}{No dimension falls here.}} \\
\bottomrule
\end{tabularx}
\tabnote{\zhen{$^{\ast}$\,Inflated by a near-zero human baseline ($0.041$); see \cref{sec:storyboard}.}%
{$^{\ast}$\,人工基线接近于零（$0.041$），比值被放大；见 \cref{sec:storyboard}。}}
\end{table}

\subsection{Training a judge model}
\label{sec:judgetrain}

\paragraph{Training data.}
Training examples pair benchmark artefacts with mean human scores and associated attribution text.
High-disagreement labels are down-weighted or excluded.
Low-scoring items and instruction-mismatched negatives broaden score coverage, while general video
samples supplement video-judge training. The test subsets in \cref{sec:ablation} are held out from
both supervised fine-tuning (SFT) and GRPO.

\paragraph{Training algorithm.}
We fine-tune two open-weight models with modality encoders frozen: \md{Qwen3-Omni-30B-A3B} for
video, and \md{Qwen3.8-27B} for keyframes and storyboards. SFT learns structured rationales and rubric scores through token,
score-regression and instruction-following objectives. Scores are continuous expectations over five band tokens,
$\hat{s}_{d}=\sum_b v_b p^{(d)}_b$, where $v_b\in\{1,\ldots,5\}$ and $p^{(d)}_b$ is the renormalised
probability of band $b$ at dimension $d$. GRPO then optimises the reward
\begin{equation}
R \;=\; R_{\text{score}} + 0.2\,R_{\text{fmt}} + 0.3\,R_{\text{attr}},
\label{eq:reward}
\end{equation}
\begin{equation}
R_{\text{score}} \;=\; \sum_{\ell \in \{1,2\}} \omega_{\ell}
  \operatorname{mean}_{d \in \mathcal{D}_{\ell}}
  \big[\,1 - \tfrac{1}{4}|\hat{s}_{d} - s^{\ast}_{d}|\,\big]_{+} ,
\label{eq:rscore}
\end{equation}
where $s^{\ast}_{d}$ is the human target and $(\omega_1,\omega_2)=(0.6,0.4)$ weight axis-level
dimensions $\mathcal{D}_1$ and their leaves $\mathcal{D}_2$. $R_{\text{fmt}}$ checks the output
schema; $R_{\text{attr}}$ compares predicted localisations and tags with available human markers.
The reward thus combines score agreement, format compliance and defect attribution.
\Cref{app:train} summarises modality-specific supervision and inference timing.

\subsection{Ablation study}
\label{sec:ablation}

\paragraph{Setup.}
We compare four prompted architectures: the full system, without routed small-model evidence,
without callable tools, and without either (a bare LMM). Measurement-only scoring and a trained judge
serve as separate baselines. The subsets contain $210$ keyframes from seven generation models and $180$ shot
videos from six.

\paragraph{Perception and tool use provide complementary gains.}
\Cref{tab:ablation} compares automated and human means over $14$ model--paradigm cells using
PLCC/SRCC, with shared video dimensions and arm-specific keyframe coverage. With each stage's prompted judge fixed
(\cref{sec:judgeconfig}), removing routed evidence or tools lowers keyframe PLCC from $0.937$ to $0.903$
or $0.891$, respectively; video PLCC falls from $0.800$ to $0.701$ or $0.656$. The full system
also exceeds bare LMM and measurement-only baselines on keyframes ($0.880$, $0.677$) and video
($0.445$, $0.632$). This pattern is consistent with complementary roles: small models supply
perceptual evidence, tools enable targeted inspection, and the LMM turns their evidence into rubric-based scores.

\paragraph{Training yields the strongest reported agreement.}
The trained judge achieves the highest PLCC/SRCC in both blocks: $0.955/0.912$ on keyframes and
$0.851/0.837$ on video. Built on open-weight models, it receives only the generation inputs, the
artefact and the rubric, without routed measurements or tools, yet exceeds the full stack by $0.018$
and $0.051$ PLCC on keyframes and video. The gains in both correlation measures are consistent with learning
rubric-specific score calibration and defect cues from annotated examples, rather than relying on
instructions alone.

\begin{table}[htbp]
\centering
\setlength{\abovecaptionskip}{4pt}
\caption{\zhen{\textbf{Scoring-architecture ablations.} \best{Bold}/\sbest{underline}: best/second-best.}{\textbf{评分架构消融。}\best{加粗}/\sbest{下划线}：最优/次优。}}
\label{tab:ablation}
\scriptsize
\setlength{\tabcolsep}{6pt}
\begin{tabular*}{\linewidth}{@{\extracolsep{\fill}}l cc@{}}
\toprule
\textbf{Scoring architecture} & \textbf{PLCC} & \textbf{SRCC} \\
\midrule
\multicolumn{3}{@{}l}{\textbf{\textcircled{2} Single keyframe}\ \ \scriptsize $210$ items $\cdot$ $7$ generation models} \\
The trained scorer (\cref{sec:judgetrain}) & \best{0.955} & \best{0.912} \\
\judger{}, the full stack & \sbest{0.937} & \sbest{0.886} \\
\quad without the routed measurements & 0.903 & 0.798 \\
\quad without the model-callable tools & 0.891 & 0.833 \\
\quad without either, a bare LMM judge & 0.880 & 0.802 \\
\hdashline[0.4pt/1.6pt]
The routed measurements alone, with no LMM judge & 0.677 & 0.771 \\
\midrule
\multicolumn{3}{@{}l}{\textbf{\textcircled{4} Shot video}\ \ \scriptsize $180$ items $\cdot$ $6$ generation models} \\
The trained scorer (\cref{sec:judgetrain}) & \best{0.851} & \best{0.837} \\
\judger{}, the full stack: routed measurements $+$ model-callable tools & \sbest{0.800} & \sbest{0.811} \\
\quad without the routed measurements & 0.701 & 0.675 \\
\quad without the model-callable tools & 0.656 & 0.688 \\
\quad without either, a bare LMM judge & 0.445 & 0.248 \\
\hdashline[0.4pt/1.6pt]
The routed measurements alone, with no LMM judge & 0.632 & 0.688 \\
\bottomrule
\end{tabular*}
\end{table}

\subsection{Stage-wise analysis}
\label{sec:perstage}

\noindent
The three parts below follow the three \emph{modalities} of the chain rather than individual
granularities, and together they cover all six: storyboard design is \textcircled{1}, the text
stage; keyframe images covers \textcircled{2} and \textcircled{3}; and shot video covers
\textcircled{4} to \textcircled{6}.

\subsubsection{Storyboard design}
\label{sec:storyboard}

Nine text-based large language models convert episode scripts into storyboard scripts, yielding
$533$ evaluation items. The automated scores of the nine range from $3.26$ to $4.23$
(\cref{tab:main}), the widest performance gap among the three single-asset granularities. Following
the annotation round, eleven additional models were evaluated on the same item set.

\begin{table}[htbp]
\centering
\caption{\zhen{Storyboard design, all eight leaf dimensions: human mean (averaging the three
annotators), automated mean, and per-item agreement on both sides. \hbase{} is the leave-one-out
human baseline. Ordered by code.}{分镜设计，八个叶子维度全表：人工三人均分、自动化均分，
以及两侧的逐题一致性。\hbase{} 是留一法人工基线。按编号排序。}}
\label{tab:sbdims}
\small
\begin{tabular}{@{}l l rr c rr r@{}}
\toprule
& & \multicolumn{2}{c}{\textbf{Mean score}} & &
\multicolumn{2}{c}{\textbf{Per-item PLCC}} & \\
\cmidrule(lr){3-4}\cmidrule(lr){6-7}
\textbf{Dimension} & \textbf{Axis} & human & judge & & \hbase{} & judge & ratio \\
\midrule
\dc{S-A1} event coverage         & \axF & 4.11 & 4.01 & & 0.150 & 0.497 & $3.3\times$ \\
\dc{S-A2} dialogue fidelity      & \axF & 4.17 & 3.91 & & 0.442 & \best{0.728} & $1.6\times$ \\
\dc{S-B1} executability          & \axP & 3.41 & 3.89 & & 0.046 & 0.076 & $1.7\times$ \\
\dc{S-B2} restraint in additions & \axP & 3.96 & 3.34 & & $-$0.023 & 0.110 & --- \\
\dc{S-C1} narrative flow         & \axE & 4.03 & 4.12 & & 0.199 & 0.256 & $1.3\times$ \\
\dc{S-C2} shooting rhythm        & \axE & 3.47 & \fail{2.74} & & 0.092 & 0.307 & $3.3\times$ \\
\dc{S-C3} emotional expression   & \axE & 3.93 & 4.18 & & 0.128 & 0.332 & $2.6\times$ \\
\dc{S-C4} audiovisual style      & \axE & 4.11 & 4.09 & & 0.041 & 0.265 & \best{$6.5\times$} \\
\midrule
\textbf{Composite}               &      & 3.51 & 3.79 & & 0.217 & 0.509 & $2.35\times$ \\
\bottomrule
\end{tabular}
\end{table}

\paragraph{Performance bottlenecks lie in schedulable production defects.}
In automated means, seven of the eight leaf dimensions meet the usable line, six of them scoring between
$3.89$ and $4.18$, which indicates that the models faithfully reproduce the core script elements:
plot events, character dialogue and scene presentation. The two weakest dimensions both govern
production logic: shooting rhythm at $2.74$ and restraint in additions at $3.34$
(\cref{tab:sbdims}). The bottleneck of the text stage lies in production scheduling and duration
budgeting rather than in linguistic generation capability. Duration budgeting follows quantitative
arithmetic rules and can be verified before any visual rendering, so the dominant defect of this
stage can be identified and corrected in pre-production.

Nine of the additional models have an earlier generation from the same family in the comparison.
All nine receive higher automated scores than their predecessors, by $+0.02$ to $+0.60$; the newest,
\md{gpt-6-astra-max} at $4.43$, has the highest score in this stage.

\paragraph{Agreement ratios depend on human reproducibility.}
Per-item agreement is $0.509$ against a human baseline of $0.217$ ($2.35\times$), and every dimension
exceeds its own baseline (\cref{tab:sbdims}). Baseline multiples reflect variation in human
reproducibility, not simply automated scoring performance. Dialogue fidelity can be checked word by
word against the script and has high agreement for both the automated scorer and humans
($0.728$ against $0.442$). In contrast, audiovisual style treatment has the largest ratio, at
$6.5\times$, despite automated agreement of $0.265$, because the human baseline is $0.041$:
annotators barely agree with one another on it. For restraint in additions, the baseline of
$-0.023$ makes the ratio uninterpretable. A large multiple supports automatability only if the
human criterion is reproducible; otherwise, higher automated agreement does not establish validity.
These dimensions still need clearer operational criteria (\cref{tab:quadrant}).

\paragraph{Additional models are scored automatically.}
The eleven additional models enter \cref{tab:main} with per-axis scores and no agreement statistics
of their own, at zero incremental human annotation cost. Their scores extend the judge validated on
the annotated models and are not themselves checked against human ratings.

\subsubsection{Keyframe images}
\label{sec:image}

Seven image models render storyboard scripts as keyframes, yielding $2{,}125$ single-panel items and
$429$ episode-level items. Models receive similar single-panel scores, but the human composite
first falls below the usable line at episode-keyframe level ($2.74$; \cref{tab:imgdims}).

\paragraph{The deficit is concentrated in placing specified content.}
In automated means, three of the fifteen single-panel dimensions fall below the usable line, and all three ask
whether content the pipeline had already specified was placed correctly: subject attributes against
the character sheet ($2.75$), the scene against its reference image ($2.72$), and camera framing and
visual style against the shot description ($2.91$). Nothing outside that family fails: the fifteen
dimensions span $2.72$--$4.59$, and neither picture-quality dimension, neither generation-plausibility dimension
and no style-consistency dimension falls below the line (\cref{tab:imgdims}). What limits the
single-panel stage is adherence to the specification rather than the ability to render a picture,
which locates the remedy in conditioning and reference handling rather than in rendering capacity.

\paragraph{Cross-shot consistency breaks down first.}
In automated means, two of the four episode-level dimensions fall below the usable line, and the lower of the two
is counter-intuitive: scene consistency ($2.50$) is worse than character consistency
($2.87$). The lower scene score reflects shifts in layout, furnishings and lighting rather than
changes to cast members (\cref{tab:imgdims}). When the same set of panels is assessed as a full
episode, it is the persistence of entities across frames that degrades, not the appearance of any
individual frame. That persistence is a property invisible to single-panel evaluation.

\paragraph{Close composite scores conceal divergent capability profiles.}
\label{sec:closepairs}
The one model pair on this board with a composite-score difference no larger than $0.03$
illustrates the axis-based capability structure introduced in \cref{sec:profiles}.
\md{seedream-5.0-pro} and \md{nano-banana-pro} differ by merely $0.02$ in automated composite score.
Yet \md{seedream-5.0-pro} leads by $0.39$ on \axF{} input fidelity and falls behind by $0.60$ on
\axP{} generation plausibility: an axis-level gap an order of magnitude larger than the composite difference.
Strengths manifest along entire capability axes rather than across scattered individual
dimensions. Composite scores are therefore sufficient for model tiering yet misleading when
selecting models for concrete production requirements.

\paragraph{Explicit references support automated agreement.}
The conventional expectation is that objective criteria automate easily and subjective ones do not,
and this stage does not sort that way. Nine of the fifteen single-panel dimensions reach or beat the
human baseline and all four episode-level ones do, while the four substantial shortfalls are exactly
the \axQ{} and \axP{} criteria (\cref{tab:imgdims}). Three of those four (technical quality, human
anatomy and environmental physics) are the most physically objective criteria in the taxonomy.
What distinguishes the outcome appears to be whether the item states the answer, not how objective
the criterion sounds. Technical quality reads as the most measurable of the fifteen, yet a soft,
low-contrast frame is as easily a deliberate texture as a defect, and nothing in the item settles
which the drama intended: the automated scorer reaches only $0.167$ against a baseline of $0.275$. Action and
interaction, a judgement about what is happening in the picture, is the stage's best-served
dimension at $0.527$ against $0.401$, because the shot description states what should be happening.

\begin{table}[htbp]
\centering
\caption{\zhen{The keyframe stages, every leaf dimension, ordered by code: human (three-annotator average) and automated mean scores, and per-item agreement on both sides. \hbase{} is the leave-one-out human baseline.}{两档分镜图的全部叶子维度，按编号排序：人工三人均分与自动化均分，以及两侧的逐题一致性。\hbase{} 为留一法人工基线。}}
\label{tab:imgdims}
\small
\setlength{\tabcolsep}{4.5pt}
\begin{adjustbox}{max width=\linewidth}
\begin{tabular}{@{}l l rr c rr r@{}}
\toprule
& & \multicolumn{2}{c}{\textbf{Mean score}} & &
\multicolumn{2}{c}{\textbf{Per-item PLCC}} & \\
\cmidrule(lr){3-4}\cmidrule(lr){6-7}
\textbf{Dimension} & \textbf{Axis} & human & judge & & \hbase{} & judge & ratio \\
\midrule
\multicolumn{8}{@{}l}{\textbf{\textcircled{2} Single keyframe}} \\
\dc{I-A1} technical quality & \axQ & 3.10 & 3.92 & & 0.275 & \fail{0.167} & $0.6\times$ \\
\dc{I-A2} aesthetic style & \axQ & 3.14 & 3.53 & & 0.194 & \fail{0.146} & $0.8\times$ \\
\dc{I-B1} subject attributes & \axF & 3.05 & \fail{2.75} & & 0.362 & \pass{0.525} & $1.5\times$ \\
\dc{I-B2} action and interaction & \axF & \fail{2.94} & 3.20 & & 0.401 & \pass{0.527} & $1.3\times$ \\
\dc{I-B3} scene and spatial layout & \axF & 3.28 & 3.28 & & 0.306 & \pass{0.443} & $1.4\times$ \\
\dc{I-B4} shot framing and visual style & \axF & \fail{2.82} & \fail{2.91} & & 0.224 & \pass{0.296} & $1.3\times$ \\
\dc{I-C1} face identity & \axF & 3.25 & 3.96 & & 0.440 & \pass{0.485} & $1.1\times$ \\
\dc{I-C2} appearance and costume & \axF & 3.69 & 3.72 & & 0.409 & \pass{0.471} & $1.2\times$ \\
\dc{I-D1} cross-panel character & \axC & 3.50 & 3.97 & & 0.391 & \pass{0.459} & $1.2\times$ \\
\dc{I-D2} subject--scene integration & \axC & 3.15 & 4.59 & & 0.218 & \pass{0.274} & $1.3\times$ \\
\dc{I-D3} scene vs.\ reference & \axC & \fail{2.75} & \fail{2.72} & & 0.189 & \fail{0.182} & $1.0\times$ \\
\dc{I-D4} screen-direction axis & \axC & 3.35 & 3.75 & & 0.166 & \pass{0.317} & $1.9\times$ \\
\dc{I-D5} style consistency & \axC & 4.12 & 4.52 & & 0.131 & \fail{0.130} & $1.0\times$ \\
\dc{I-E1} human anatomy & \axP & 3.21 & 3.87 & & 0.439 & \fail{0.311} & $0.7\times$ \\
\dc{I-E2} environmental physics & \axP & 3.26 & 4.50 & & 0.234 & \fail{0.152} & $0.7\times$ \\
\quad\emph{composite} & & 3.21 & 3.62 & & 0.449 & 0.536 & $1.19\times$ \\
\addlinespace[3pt]\midrule
\multicolumn{8}{@{}l}{\textbf{\textcircled{3} Episode keyframes}} \\
\dc{MI-D1} cross-shot character & \axC & \fail{2.59} & \fail{2.87} & & 0.339 & \pass{0.456} & $1.3\times$ \\
\dc{MI-D2} subject--scene integration & \axC & \fail{2.80} & 4.37 & & 0.110 & \pass{0.229} & $2.1\times$ \\
\dc{MI-D3} cross-shot scene & \axC & \fail{2.42} & \fail{2.50} & & 0.098 & \pass{0.160} & $1.6\times$ \\
\dc{MI-D5} cross-shot style & \axC & 3.22 & 3.60 & & 0.240 & \pass{0.304} & $1.3\times$ \\
\quad\emph{composite} & & 2.74 & 3.30 & & 0.233 & 0.344 & $1.48\times$ \\
\bottomrule
\end{tabular}
\end{adjustbox}
\end{table}

\subsubsection{Shot video and short drama}
\label{sec:video}

Six video models animate keyframes and the results are assembled into episodes, yielding $2{,}186$
shot-level items, $306$ episode-video items and $206$ short dramas. The automated top-to-bottom gap among
annotated models widens from $0.36$ points on single shots to $0.51$ on short dramas.

\paragraph{The weakest dimensions are the ones that unfold over time.}
In automated means, the twenty leaf dimensions of this stage span $1.49$ to $4.15$ (\cref{tab:viddims}). Audio-visual
sync scores the lowest at $1.49$ and character performance also ranks among the bottom at $2.34$,
while the three highest are style consistency ($4.15$), keyframe adherence ($4.14$) and image quality
($4.11$). The criterion separating the two ends is whether a dimension addresses the static
appearance of the frame or the temporal behaviour of the clip, and the models perform worst on the
latter. This is a class of defect that no still-image benchmark can detect at all.

\begin{table}[htbp]
\centering
\caption{\zhen{The three video granularities, every leaf dimension, ordered by code: human (three-annotator average) and automated mean scores, and per-item agreement on both sides. \hbase{} is the leave-one-out human baseline.}{三档视频粒度的全部叶子维度，按编号排序：人工三人均分与自动化均分，以及两侧的逐题一致性。\hbase{} 为留一法人工基线。}}
\label{tab:viddims}
\footnotesize
\setlength{\tabcolsep}{4.5pt}
\begin{adjustbox}{max width=\linewidth}
\begin{tabular}{@{}l l rr c rr r@{}}
\toprule
& & \multicolumn{2}{c}{\textbf{Mean score}} & &
\multicolumn{2}{c}{\textbf{Per-item PLCC}} & \\
\cmidrule(lr){3-4}\cmidrule(lr){6-7}
\textbf{Dimension} & \textbf{Axis} & human & judge & & \hbase{} & judge & ratio \\
\midrule
\multicolumn{8}{@{}l}{\textbf{\textcircled{4} Shot video}} \\
\dc{V-A1} image quality & \axQ & 3.27 & 4.11 & & 0.191 & \fail{0.076} & $0.4\times$ \\
\dc{V-A2} aesthetic style and motion & \axQ & 3.25 & \fail{2.72} & & 0.169 & \fail{0.096} & $0.6\times$ \\
\dc{V-B1} subject attributes & \axF & 3.54 & 3.43 & & 0.314 & \fail{0.264} & $0.8\times$ \\
\dc{V-B2} shot framing and visual style & \axF & 3.33 & \fail{2.55} & & 0.163 & \fail{0.155} & $0.9\times$ \\
\dc{V-B3} keyframe adherence & \axF & 3.48 & 4.14 & & 0.380 & \fail{0.324} & $0.8\times$ \\
\dc{V-C1} character appearance & \axC & 3.80 & 3.44 & & 0.282 & \fail{0.231} & $0.8\times$ \\
\dc{V-C2} scene and prop consistency & \axC & 3.39 & \fail{2.33} & & 0.304 & \fail{0.194} & $0.6\times$ \\
\dc{V-C3} style consistency & \axC & 4.09 & 4.15 & & 0.245 & \fail{0.149} & $0.6\times$ \\
\dc{V-C4} action consistency & \axC & 3.38 & \fail{2.56} & & 0.211 & \pass{0.247} & $1.2\times$ \\
\dc{V-C5} subject--scene integration & \axC & 3.40 & 3.61 & & 0.144 & \fail{0.127} & $0.9\times$ \\
\dc{V-C6} on-screen text & \axC & 3.44 & \fail{2.41} & & 0.345 & \pass{0.393} & $1.1\times$ \\
\dc{V-D1} motion smoothness & \axP & 3.60 & 3.31 & & 0.208 & \fail{0.069} & $0.3\times$ \\
\dc{V-D3} physical plausibility & \axP & 3.10 & 3.16 & & 0.331 & \fail{0.231} & $0.7\times$ \\
\dc{V-D4} causal and temporal & \axP & 3.35 & \fail{2.49} & & 0.158 & \fail{0.136} & $0.9\times$ \\
\dc{V-E1} speech quality & \axE & 3.54 & \fail{2.85} & & 0.474 & \fail{0.354} & $0.8\times$ \\
\dc{V-E2} audio-visual sync & \axE & 3.15 & \fail{1.49} & & 0.448 & \fail{0.075} & $0.2\times$ \\
\dc{V-E3} sound effects and music & \axE & 3.38 & 3.58 & & 0.252 & \fail{0.061} & $0.2\times$ \\
\dc{V-F1} composition & \axE & 3.76 & 4.01 & & 0.216 & \fail{0.118} & $0.6\times$ \\
\dc{V-F2} camera movement & \axE & 3.49 & 3.66 & & 0.147 & \fail{0.070} & $0.5\times$ \\
\dc{V-F3} character performance & \axE & 3.07 & \fail{2.34} & & 0.183 & \fail{0.122} & $0.7\times$ \\
\quad\emph{composite} & & 3.44 & 3.09 & & 0.380 & 0.301 & $0.79\times$ \\
\addlinespace[3pt]\midrule
\multicolumn{8}{@{}l}{\textbf{\textcircled{5} Episode video}} \\
\dc{MV-C1} cross-shot character & \axC & \fail{2.55} & 3.21 & & 0.288 & \fail{0.283} & $1.0\times$ \\
\dc{MV-C2} cross-shot scene and props & \axC & \fail{2.53} & \fail{1.82} & & 0.194 & \fail{0.099} & $0.5\times$ \\
\dc{MV-C3} cross-shot style & \axC & 3.11 & 3.32 & & 0.299 & \pass{0.306} & $1.0\times$ \\
\dc{MV-C4} cross-shot action & \axC & \fail{2.65} & \fail{1.93} & & 0.166 & \fail{$-0.022$} & $-0.1\times$ \\
\dc{MV-C5} subject--scene integration & \axC & 3.19 & 4.11 & & 0.123 & \fail{0.027} & $0.2\times$ \\
\dc{MV-C6} cross-shot text & \axC & 3.10 & \fail{1.83} & & 0.227 & \pass{0.333} & $1.5\times$ \\
\dc{MV-D4} cross-shot causal/temporal & \axP & 3.09 & 3.35 & & 0.083 & \pass{0.087} & $1.0\times$ \\
\dc{MV-E1} cross-shot speech & \axE & \fail{2.75} & 3.06 & & 0.407 & \fail{0.019} & $0.1\times$ \\
\dc{MV-E3} cross-shot sound & \axE & 3.76 & 3.00 & & 0.172 & \fail{0.025} & $0.1\times$ \\
\dc{MV-F3} cross-shot performance & \axE & 3.17 & 3.94 & & 0.122 & \fail{0.017} & $0.1\times$ \\
\quad\emph{composite} & & 2.79 & 2.93 & & 0.240 & 0.203 & $0.85\times$ \\
\addlinespace[3pt]\midrule
\multicolumn{8}{@{}l}{\textbf{\textcircled{6} Short drama}} \\
\dc{O-A1} event completion & \axF & 3.00 & 3.58 & & 0.200 & \pass{0.363} & $1.8\times$ \\
\dc{O-B1} style & \axC & 3.19 & 3.67 & & 0.461 & \fail{0.302} & $0.7\times$ \\
\dc{O-B2} cross-episode consistency & \axC & \fail{2.79} & 3.13 & & 0.433 & \fail{0.306} & $0.7\times$ \\
\dc{O-B3} cross-episode progression & \axC & \fail{2.96} & \fail{2.87} & & 0.130 & \pass{0.149} & $1.1\times$ \\
\dc{O-C1} watchability & \axE & \fail{2.48} & \fail{2.36} & & 0.343 & \fail{0.074} & $0.2\times$ \\
\dc{O-C2} pacing and transitions & \axE & \fail{2.90} & \fail{2.85} & & 0.152 & \fail{$-0.112$} & $-0.7\times$ \\
\quad\emph{composite} & & 2.61 & 3.03 & & 0.450 & 0.384 & $0.85\times$ \\
\bottomrule
\end{tabular}
\end{adjustbox}
\end{table}

\paragraph{A coarser unit of judgement counts a different kind of defect.}
Four of the ten dimensions defined at both shot and episode granularity track consistency
across shots, and the automated means of all four drop as the unit of judgement coarsens: style
consistency from $4.15$ to $3.32$, action
consistency from $2.56$ to $1.93$, scene and prop consistency from $2.33$ to $1.82$, and character
appearance consistency from $3.44$ to $3.21$ (\cref{fig:granularity}). The kind of defect behind a
deduction changes as well. A shot-level deduction indicates an execution error in the individual
clip, an episode-video deduction reflects a mismatch between the clips within one episode, and a
multi-episode deduction corresponds to narrative content the episode invented. That defect class is
entirely absent from shot-level evaluation. Coarsening the unit of judgement reshapes the
spectrum of detectable defects rather than merely altering their count, so results confined to
individual clips cannot be extrapolated to the delivery standard of a short drama.

\begin{figure}[htbp]
\centering
\begin{subfigure}[t]{0.465\linewidth}\centering
\begin{tikzpicture}
\begin{axis}[dcbase, width=5.05cm, height=5.2cm,
  xmin=1.6, xmax=4.5, xtick={2,3,4},
  ymin=0.5, ymax=4.85,
  ytick={1,2,3,4},
  yticklabels={action, scene/props, character, style},
  yticklabel style={font=\scriptsize},
  xticklabel style={font=\scriptsize},
  xlabel={automated mean (of $5$)}, xlabel style={font=\scriptsize},
  xmajorgrids, ymajorgrids=false]
\draw[-{Latex[length=4.5pt,width=4.5pt]}, rulegray, line width=1.1pt] (axis cs:2.56,1) -- (axis cs:1.93,1);
\fill[tierone] (axis cs:2.56,1) circle (2.4pt); \fill[tierfour] (axis cs:1.93,1) circle (2.4pt);
\node[anchor=south, font=\scriptsize, text=rulegray] at (axis cs:2.245,1.10) {$-0.63$};
\draw[-{Latex[length=4.5pt,width=4.5pt]}, rulegray, line width=1.1pt] (axis cs:2.33,2) -- (axis cs:1.82,2);
\fill[tierone] (axis cs:2.33,2) circle (2.4pt); \fill[tierfour] (axis cs:1.82,2) circle (2.4pt);
\node[anchor=south, font=\scriptsize, text=rulegray] at (axis cs:2.075,2.10) {$-0.51$};
\draw[-{Latex[length=4.5pt,width=4.5pt]}, rulegray, line width=1.1pt] (axis cs:3.44,3) -- (axis cs:3.21,3);
\fill[tierone] (axis cs:3.44,3) circle (2.4pt); \fill[tierfour] (axis cs:3.21,3) circle (2.4pt);
\node[anchor=south, font=\scriptsize, text=rulegray] at (axis cs:3.325,3.10) {$-0.23$};
\draw[-{Latex[length=4.5pt,width=4.5pt]}, rulegray, line width=1.1pt] (axis cs:4.15,4) -- (axis cs:3.32,4);
\fill[tierone] (axis cs:4.15,4) circle (2.4pt); \fill[tierfour] (axis cs:3.32,4) circle (2.4pt);
\node[anchor=south, font=\scriptsize, text=rulegray] at (axis cs:3.735,4.10) {$-0.83$};
\end{axis}
\end{tikzpicture}
\subcaption{\zhen{Shot level \textcolor{tierone}{$\bullet$} to episode
  \textcolor{tierfour}{$\bullet$}: the four cross-shot consistency dimensions all fall.}%
  {分镜级 \textcolor{tierone}{$\bullet$} 到集级 \textcolor{tierfour}{$\bullet$}：四个跨镜头一致性维度全部下降。}}
\label{fig:gran:same}
\end{subfigure}\hfill
\begin{subfigure}[t]{0.515\linewidth}\centering
\begin{tikzpicture}
\begin{axis}[dcbase, width=5.25cm, height=5.2cm,
  xmin=1.3, xmax=3.1, xtick={1.5,2.0,2.5,3.0},
  ymin=0.35, ymax=9.65,
  ytick={1,2,3, 4,5,6, 7,8,9},
  yticklabels={\textcircled{6} progression, \textcircled{6} pacing,
               \textcircled{6} watchability,
               \textcircled{5} action, \textcircled{5} text,
               \textcircled{5} scene/props,
               \textcircled{4} performance, \textcircled{4} scene/props,
               \textcircled{4} AV sync},
  yticklabel style={font=\scriptsize},
  xticklabel style={font=\scriptsize},
  xlabel={automated mean (of $5$)}, xlabel style={font=\scriptsize},
  xmajorgrids, ymajorgrids=false]
\draw[rulegray!45, line width=0.3pt] (axis cs:1.3,6.5) -- (axis cs:3.1,6.5);
\draw[rulegray!45, line width=0.3pt] (axis cs:1.3,3.5) -- (axis cs:3.1,3.5);
\addplot[only marks, mark=*, mark size=2.7pt, color=tierthree!85]
  coordinates {(1.49,9) (2.33,8) (2.34,7)};
\addplot[only marks, mark=square*, mark size=2.5pt, color=tierfour!85]
  coordinates {(1.82,6) (1.83,5) (1.93,4)};
\addplot[only marks, mark=triangle*, mark size=3.1pt, color=tierfive!90]
  coordinates {(2.36,3) (2.85,2) (2.87,1)};
\end{axis}
\end{tikzpicture}
\subcaption{\zhen{\textcircled{4} shot video, \textcircled{5} episode video,
  \textcircled{6} short drama.}%
  {\textcircled{4} 分镜视频、\textcircled{5} 集分镜视频、\textcircled{6} 短剧成片。}}
\label{fig:gran:kinds}
\end{subfigure}

\caption{\zhen{Different judgement units reveal different defects.
(\subref{fig:gran:same}) The same four consistency criteria score lower at episode level.
(\subref{fig:gran:kinds}) The three lowest automated means are selected separately at each
granularity, not from a shared set of dimensions: sync, scene/props and performance for shots;
cross-shot scene/props, text and action for episodes; watchability, pacing and progression for
complete dramas.}{不同评判单位揭示不同缺陷。（\subref{fig:gran:same}）同四个一致性标准在单集层面得分更低。
（\subref{fig:gran:kinds}）按各粒度的自动均分分别选最低三项，而非比较同一组维度：镜头层面为音画同步、场景／道具和表演；
单集层面为跨镜头场景／道具、文字和动作；整剧层面为可看性、节奏和推进。}}
\label{fig:granularity}
\end{figure}

\paragraph{Input constraints shape paradigm strengths.}
No paradigm achieves absolute superiority on the shot-video board, and each one's performance
profile is determined by its input constraint mechanism (\cref{tab:paradigm}). The grid panel
paradigm integrates multiple shots into a single image, so every shot shares the pixel information
for character subjects and scene backgrounds, and it achieves the best results in subject--scene
integration ($4.28$), character appearance consistency ($3.67$) and style consistency ($4.32$); each
shot, however, corresponds to a local cell of that image rather than an independent frame of its own,
which puts adherence to a cell ($3.97$) below adherence to a frame pair ($4.28$). The first/last
frame paradigm supplies both endpoints of each shot, which gives it the advantage in keyframe
adherence ($4.28$), the one property its inputs fix exactly; it imposes no constraint on the
intermediate content, and it scores lowest in subject--scene integration ($3.01$) and camera movement
($3.30$). The multi-reference paradigm requires no fixed frame to reproduce, which frees the camera
and yields both the highest camera movement plausibility ($4.09$) and the best sound and music
($3.62$), while the absence of a fixed visual anchor leaves it last in style consistency ($4.04$),
by a margin of only $0.28$. Each paradigm gains its particular strength by virtue of the input constraint it
chooses to waive, so the input format is a deployment decision affecting practical performance
rather than an implementation detail (\cref{app:paradigmcase}).

\begin{table}[htbp]
\centering
\caption{\zhen{The three input paradigms on the shot video stage, as automated means per
paradigm. \best{Bold} leads the row and \fail{red} trails it; every paradigm appears in both
colours. \emph{n/a}: multi-reference supplies no frame to adhere to.}%
{三种输入范式在分镜视频环节上的表现，表内为各范式的自动化均分。\best{加粗}为该行领先、\fail{红色}为该行垫底；
三种范式都同时出现在两种颜色里。\emph{n/a}：多参考范式不提供可供遵循的参考帧。}}
\label{tab:paradigm}
\small
\setlength{\tabcolsep}{9pt}
\begin{tabular}{@{}l ccc@{}}
\toprule
\textbf{Dimension} & \textbf{first/last frame} & \textbf{grid panel} & \textbf{multi-reference} \\
\midrule
\dc{V-C5} subject--scene integration & \fail{3.01} & \best{4.28} & 3.96 \\
\dc{V-F2} camera movement            & \fail{3.30} & 3.82 & \best{4.09} \\
\dc{V-C3} style consistency          & 4.12 & \best{4.32} & \fail{4.04} \\
\dc{V-B3} keyframe adherence         & \best{4.28} & \fail{3.97} & \emph{n/a} \\
\dc{V-C1} character appearance       & \fail{3.29} & \best{3.67} & 3.45 \\
\dc{V-E3} sound and music            & 3.59 & \fail{3.51} & \best{3.62} \\
\bottomrule
\end{tabular}
\end{table}

\paragraph{Reference availability also matters for video.}
The rule holds here as well, in per-item agreement rather than in mean score. Event completion, which
can be verified against the event list in the script, reaches $0.363$ against a baseline of $0.200$,
while watchability of the cut, for which no textual reference exists, falls to $0.074$ against a
baseline of $0.343$ (\cref{tab:viddims}). This stage also supplies a quasi-controlled check of
the pattern: speech quality and audio-visual sync share a modality and a tool chain and carry comparable
human baselines ($0.474$ and $0.448$), yet the automated scorer reaches $0.354$ on the first and
$0.075$ on the second. The main difference is that the script's dialogue provides a clear
textual anchor for speech quality, whereas no such reference exists for sync. At shot level, audio-visual
sync is both the lowest-scoring dimension and the one furthest below its own human
baseline ($0.17\times$), so its position records the ceiling of the automated scorer rather than the
relative ranking of the generation models.

\subsection{Chain-level analysis}
\label{sec:discussion}
\label{sec:chain}

\paragraph{Upstream defects propagate down the generation chain.}
\label{sec:prop}

\begin{figure}[htbp]
\centering
\begin{adjustbox}{max width=\linewidth}
\begin{tikzpicture}
\begin{axis}[
  dcbase, width=14.6cm, height=5.2cm,
  ybar, bar width=16pt,
  xmin=0.25, xmax=14.75,
  ymin=-0.95, ymax=0.02,
  xtick={1,2,3,4, 6,7,8,9, 11,12,13,14},
  xticklabels={%
    \textcolor{tierfour}{\md{hy3}}, \textcolor{tierfour}{\md{mimo-v2.5-pro}},
    \textcolor{axF}{\md{nano-banana-pro}}, \textcolor{axF}{\md{seedream-5.0-lite}},
    \textcolor{tierfour}{\md{hy3}}, \textcolor{tierfour}{\md{mimo-v2.5-pro}},
    \textcolor{axF}{\md{nano-banana-pro}}, \textcolor{axF}{\md{seedream-5.0-lite}},
    \textcolor{tierfour}{\md{hy3}}, \textcolor{tierfour}{\md{mimo-v2.5-pro}},
    \textcolor{axF}{\md{nano-banana-pro}}, \textcolor{axF}{\md{seedream-5.0-lite}}},
  xticklabel style={font=\tiny, rotate=32, anchor=east, xshift=2pt, yshift=-1pt},
  x tick style={draw=none},
  ytick={0,-0.2,-0.4,-0.6,-0.8},
  yticklabel style={font=\scriptsize},
  ylabel={change vs.\ the baseline chain}, ylabel style={font=\scriptsize},
  ymajorgrids, xmajorgrids=false, clip=false,
  nodes near coords, point meta=explicit symbolic,
  every node near coord/.append style={font=\tiny, anchor=north, yshift=-1pt},
]
\addplot[bar shift=0pt, fill=tierfour!85, draw=none] coordinates {
  (1,-0.07) [$-0.07$] (2,-0.13) [$-0.13$]
  (6,-0.06) [$-0.06$] (7,-0.37) [$-0.37$]
  (11,-0.50) [$-0.50$] (12,-0.83) [$-0.83$]};
\addplot[bar shift=0pt, fill=axF!85, draw=none] coordinates {
  (3,-0.10) [$-0.10$] (4,-0.13) [$-0.13$]
  (8,-0.16) [$-0.16$] (9,-0.17) [$-0.17$]
  (13,-0.29) [$-0.29$] (14,-0.25) [$-0.25$]};
\draw[inkgray, line width=0.7pt] (axis cs:0.25,0) -- (axis cs:14.75,0);
\node[anchor=south, font=\scriptsize\bfseries] at (axis cs:2.5,0.005)
  {\textcircled{4} shot video};
\node[anchor=south, font=\scriptsize\bfseries] at (axis cs:7.5,0.005)
  {\textcircled{5} episode video};
\node[anchor=south, font=\scriptsize\bfseries] at (axis cs:12.5,0.005)
  {\textcircled{6} short drama};
\draw[rulegray!60, line width=0.3pt] (axis cs:5,-0.93) -- (axis cs:5,0);
\draw[rulegray!60, line width=0.3pt] (axis cs:10,-0.93) -- (axis cs:10,0);
\end{axis}
\end{tikzpicture}
\end{adjustbox}
\caption{\zhen{Upstream defects accumulate rather than staying local. Four substitutions, each read at three granularities: one
upstream stage is replaced by a weaker model and everything else is held unchanged, either the
storyboard model (\textcolor{tierfour}{red}) or the keyframe model (\textcolor{axF}{blue}). The
baseline chain is \md{gpt-5.5-xhigh} for storyboarding, \md{gpt-image-2} for keyframes and
\md{seedance-2.0} (animation) or \md{happyhorse-1.1} (live action) for video. The same substitution
that costs an individual clip $0.07$--$0.13$ points costs the short drama up to $0.83$.
Full design in \cref{sec:degrade}. Scores are automated; the substituted chains carry no human annotation.}{上游缺陷会累积，不会留在原地。图为四种替换、各自在三个粒度上读取：只把一个上游环节换成更弱的模型、
其余全部保持不变，换掉的或是分镜模型（\textcolor{tierfour}{红}）、或是图像模型
（\textcolor{axF}{蓝}）。基线链路为分镜 \md{gpt-5.5-xhigh}、图像 \md{gpt-image-2}、
视频 \md{seedance-2.0}（动画）或 \md{happyhorse-1.1}（真人）。
同一次替换让单条 clip 只差 $0.07$--$0.13$ 分，却让短剧成片差到 $0.83$ 分。
完整设计见 \cref{sec:degrade}。分数为自动化评分，替换链路没有人工标注。}}
\label{fig:prop}
\end{figure}

If the video generation stage constituted the primary bottleneck, swapping upstream models would
yield negligible changes to multi-episode outputs. We observe substantial shifts instead. With the
video model fixed, we replace one upstream stage with a weaker alternative: the storyboard stage
using \md{hy3} or \md{mimo-v2.5-pro}, and the image stage using \md{nano-banana-pro} or
\md{seedream-5.0-lite}. We run this across four dramas under the first/last-frame paradigm, yielding
four substitutions read at three granularities, and compare them against a baseline chain of the leading annotated models,
with \md{happyhorse-1.1} for live action, which \md{seedance-2.0} does not cover (\cref{sec:degrade}). In automated scores, a single-stage substitution induces quality degradations
of $0.07$--$0.13$ points for individual clips, $0.06$--$0.37$ points for continuous segments, and
$0.25$--$0.83$ points for multi-episode outputs (\cref{fig:prop}); the substituted chains were scored automatically and
carry no human annotation. The measured degradation
scales with the temporal observation window, since decisions made at upstream stages are
irreversible and cannot be revised by any downstream module.

\paragraph{Judgement granularity changes evaluation scores.}
\label{sec:unit}
Individual generated assets often achieve favourable scores in isolation, yet assembling those
standalone frames and shots into continuous segments or full episodes accumulates latent cross-frame
inconsistencies that degrade the overall score sharply. In three-annotator means, one episode's keyframes score $3.33$--$4.00$
when judged one at a time and $2.33$ when judged as a set; one episode's shots, whose individually
judged members range from $3.23$ to $4.00$, obtain only $1.67$ as a continuous segment,
with every penalty stemming from cross-shot mismatches in character presentation, scene and prop
alignment, and action coherence (\cref{fig:unit}). The same granularity-dependent gap holds across the
corpus: across character appearance, scene and props, action coherence and speech
quality alike, human scores drop by close to a full band from shot-level to episode-level evaluation. A single-asset benchmark therefore overlooks cumulative cross-frame
inconsistency and measures an isolated quantity, which makes it an invalid proxy for episode-level
generation quality.

\begin{figure}[htbp]
\centering
\begin{adjustbox}{max width=\linewidth}
\begin{tikzpicture}
\begin{axis}[
  dcbase, width=10.6cm, height=3.5cm,
  xmin=0.9, xmax=4.35,
  xtick={1,1.5,2,2.5,3,3.5,4},
  ymin=0.4, ymax=2.6,
  ytick={1,2},
  yticklabels={\zhen{one episode's shots}{一集的分镜视频},
               \zhen{one episode's keyframes}{一集的分镜图}},
  yticklabel style={font=\scriptsize, align=right},
  xticklabel style={font=\scriptsize},
  xlabel={\zhen{score of the very same artefacts (of $5$)}{同一批产物的得分（$5$ 分制）}},
  xlabel style={font=\scriptsize},
  xmajorgrids, ymajorgrids=false, clip=false]
\draw[tierone!30, line width=9pt] (axis cs:3.33,2) -- (axis cs:4.00,2);
\draw[-{Latex[length=4.5pt,width=4.5pt]}, rulegray, line width=0.9pt]
  (axis cs:3.28,2) -- (axis cs:2.42,2);
\fill[tierfour] (axis cs:2.33,2) circle (2.6pt);
\node[anchor=south, font=\tiny, text=tierone!55!black] at (axis cs:3.665,2.16) {\zhen{one at a time}{逐个判} $3.33$--$4.00$};
\node[anchor=north, font=\tiny, text=tierfour] at (axis cs:2.33,1.86) {\zhen{together}{合起来} $2.33$};
\draw[tierone!30, line width=9pt] (axis cs:3.23,1) -- (axis cs:4.00,1);
\draw[-{Latex[length=4.5pt,width=4.5pt]}, rulegray, line width=0.9pt]
  (axis cs:3.18,1) -- (axis cs:1.76,1);
\fill[tierfour] (axis cs:1.67,1) circle (2.6pt);
\node[anchor=south, font=\tiny, text=tierone!55!black] at (axis cs:3.615,1.16) {\zhen{one at a time}{逐个判} $3.23$--$4.00$};
\node[anchor=north, font=\tiny, text=tierfour] at (axis cs:1.67,0.86) {\zhen{together}{合起来} $1.67$};
\addplot[black, dashed, forget plot] coordinates {(3.0,0.4) (3.0,2.6)};
\node[anchor=south, font=\tiny] at (axis cs:3.0,2.6) {\zhen{usable line}{可用线}};
\end{axis}
\end{tikzpicture}
\end{adjustbox}
\caption{\zhen{The same artefacts scored individually and as an episode (three-annotator means). Changing the judgement
unit exposes cross-shot defects that individual scores do not capture.}{对同一组产物分别进行单项评分与整集评分（三名标注员的均分）。
改变评判单位会揭示单项评分无法反映的跨镜头缺陷。}}
\label{fig:unit}
\end{figure}

\paragraph{Automation limits track modality.}
\label{sec:modality}
Averaged by modality, per-item agreement falls along the generation pipeline ($0.509$ on
text, $0.440$ on pixels, $0.296$ on time and audio; \cref{sec:align}), which may superficially
suggest that later stages are inherently harder to evaluate. Decomposed by evaluation axis and data
modality, however, the decline tracks modality more than stage (\cref{tab:axismod}).
Reading column-wise, \axF{} input fidelity achieves the strongest performance within every modality;
reading row-wise, every axis except \axP{} degrades as the modality incorporates pixels, then temporal
dynamics and audio. The two factors appear additive. A representative example is \axE{} cinematic
expressiveness: all four dimensions reach the human baseline when the requirement is encoded in a
storyboard script, yet zero out of eleven do so when the identical requirement is rendered into
video. Downstream stages do not introduce more subjective criteria; rather, the evidence
required to assess them becomes perceptually harder to extract, which suggests that the remedy lies
in perception rather than prompt engineering.

\begin{table}[htbp]
\centering
\caption{\zhen{Per-item agreement by axis $\times$ modality. Each cell gives the human baseline and
the automated scorer (PLCC), their difference, and how many leaf dimensions reach the baseline.}%
{按轴 $\times$ 模态看逐题一致性。每格给出人工基线与自动化判分器（PLCC）、两者之差，
以及达到基线的叶子维度数。}}
\label{tab:axismod}
\small
\setlength{\tabcolsep}{5pt}
\begin{adjustbox}{max width=\linewidth}
\begin{tabular}{@{}l ccc ccc ccc@{}}
\toprule
& \multicolumn{3}{c}{\textbf{\zhen{text}{文本}}} &
\multicolumn{3}{c}{\textbf{\zhen{pixels}{像素}}} &
\multicolumn{3}{c}{\textbf{\zhen{time $+$ audio}{时序 $+$ 音轨}}} \\
\cmidrule(lr){2-4}\cmidrule(lr){5-7}\cmidrule(lr){8-10}
\textbf{\zhen{Axis}{轴}} &
\hbase{}\,/\,\zhen{judge}{判分} & $\Delta$ & \zhen{win}{胜} &
\hbase{}\,/\,\zhen{judge}{判分} & $\Delta$ & \zhen{win}{胜} &
\hbase{}\,/\,\zhen{judge}{判分} & $\Delta$ & \zhen{win}{胜} \\
\midrule
\axF{} \zhen{input fidelity}{输入保真} & .296\,/\,\best{.613} & \up{.317} & \pass{2/2} &
                     .357\,/\,.458 & \up{.101} & \pass{6/6} &
                     .264\,/\,.277 & \up{.012} & 1/4 \\
\axC{} \zhen{internal consistency}{内部一致} & --- & --- & --- &
                     .209\,/\,.279 & \up{.070} & 7/9 &
                     .257\,/\,.208 & \dn{.049} & 5/15 \\
\axP{} \zhen{generation plausibility}{生成合理} & .011\,/\,.093 & \up{.082} & \pass{2/2} &
                     .337\,/\,.231 & \dn{.105} & \fail{0/2} &
                     .195\,/\,.131 & \dn{.064} & 1/4 \\
\axQ{} \zhen{visual quality}{视觉质量} & --- & --- & --- &
                     .235\,/\,.157 & \dn{.078} & \fail{0/2} &
                     .180\,/\,.086 & \dn{.094} & \fail{0/2} \\
\axE{} \zhen{cinematic expressiveness}{影视表现力} & .115\,/\,.290 & \up{.175} & \pass{4/4} &
                     --- & --- & --- &
                     .265\,/\,\fail{.075} & \dn{.190} & \fail{0/11} \\
\bottomrule
\end{tabular}
\end{adjustbox}
\end{table}

\paragraph{Attribution requires real upstream artefacts.}
Every item here is produced
by running the pipeline and forking only at the stage under test, so a defect is charged to the
stage that produced it rather than to the last stage that touched it: the storyboard that compressed
three speakers into a single shot, leaving the video model to infer who says which line, and the
keyframes that let the scene drift before the people do, are recorded against storyboard design and
the keyframe stage. This is what running the chain buys, and it is the property on which every
result in this section rests.

\section{Conclusion}
\label{sec:conclusion}

Short-drama generation constitutes a multi-stage chain. \bench{} evaluates each stage of this chain
using outputs produced within the chain itself. This design enables defects to be attributed to
their originating stage rather than to the stage where they are merely observed. Three systems make
that possible: \pipe{}, a production pipeline that is calibrated against commercial short-drama
platforms, generates every item, and forks only at the stage under test; \labeler{}, under which
professional annotators score each item on every applicable dimension, with deductions localised
in space and time where applicable and attributed from a fixed vocabulary; and \judger{}, an agentic automated scorer validated
against that human reference before being used to rank models.

We would rather \bench{} be used as a diagnosis than as a leaderboard. The pipeline, the dimension
system and the scoring configurations are reusable as they stand. A new model can be plugged into
any individual stage and scored automatically, although those scores are not themselves checked
against human ratings, and human-dependent components are explicitly exposed instead of being
obfuscated. We encourage future work to conduct evaluation
at the actual failure points of short-drama production: across inter-stage boundaries and throughout
shots within an episode, instead of evaluating individual clips in isolation.
\section{Limitations}
\label{sec:limitations}

Each model-level correlation rests on six to nine annotated models, and several leaf dimensions
have low human reproducibility (\cref{sec:slice-consensus}), which weakens them as alignment
targets. \label{sec:cov}Coverage is also incomplete and non-random: provider-side moderation and
generation failures limited \md{seedream-5.0-lite} to $10$ of the $20$ dramas, \md{veo-3.1} and
\md{wan2.7} to first/last-frame generation, \md{seedance-2.0} to animation and \md{seedance-2.5} to
animated multi-reference items, and these models ($\ddagger$ in \cref{tab:main}) are scored on the
items they produced. Duration budgets differ as well: \md{wan-3.0} and \md{seedance-2.5} ($30$\,s, replanned
storyboards) and \md{veo-3.1} ($10$\,s) are matched by drama, episode, paradigm and shot index. A $63$-dimension
taxonomy invites the assumption that it is exhaustive, so \cref{tab:scope} states the negative
coverage explicitly.

\begin{table}[htbp]
\centering
\caption{\zhen{What the benchmark does not cover, and why.}{本基准未覆盖什么，以及为什么。}}
\label{tab:scope}
\small
\begin{adjustbox}{max width=\linewidth}
\begin{tabular}{@{}l p{6.0cm} p{6.4cm}@{}}
\toprule
& \textbf{Not covered} & \textbf{Why, and what follows} \\
\midrule
\textbf{Stages} &
Post-production: music and sound design, editing and colour grading, subtitling and thumbnails. &
No unified automated approach to these stages exists yet. \\
\addlinespace[2pt]
\textbf{Dimensions} &
All commercial outcome measures: completion, retention, conversion. &
These outcomes are hard to quantify offline. \\
\addlinespace[2pt]
\textbf{Content} &
Hybrid, documentary and other non-narrative formats. &
The axes assume a shot script to be faithful to. \\
\addlinespace[2pt]
\textbf{Systems} &
Text-to-video without an intermediate keyframe. &
The chain always produces keyframes, so a model that skips them cannot enter as it stands. \\
\addlinespace[2pt]
\textbf{Judgment} &
Broad comparisons across judging-model families. &
The reported ablations cover two stages, not a comprehensive comparison of judging models. \\
\bottomrule
\end{tabular}
\end{adjustbox}
\end{table}

\section*{Ethics and Responsible Use}
\phantomsection
\label{sec:ethics}
\addcontentsline{toc}{section}{Ethics and Responsible Use}

All artefacts are synthetic: the scripts are project-original, every character sheet, keyframe and
clip is a model output, and no real-person likeness, voice or commercial drama asset was used; the
three commercial platforms (\cref{sec:sourcing}) ran our scripts through public interfaces. The
corpus inherits the genre's tropes of conflict, coercion and revenge; items were screened against
commercial release policies, but no broader harm audit was performed. Annotators were professional
contracted staff from third-party vendors (\cref{sec:annot}), paid on vendor terms. The benchmark is
meant for per-stage model comparison and defect diagnosis; it neither predicts commercial
performance (\cref{tab:scope}) nor evaluates safety or content policy.
\section*{Acknowledgements}
\phantomsection
\label{sec:ack}
\addcontentsline{toc}{section}{Acknowledgements}

We would like to express our sincere thanks to Yunxin Li, Baotian Hu and Min Zhang (Shenzhen Loop Area
Institute), Yong Xiang (Peking University), Fan Hong (Beijing Film Academy) and Fei Gao (Shenzhen
University) for their support on this paper. We also greatly appreciate the professional advice on film
and television production from Yekai Xu and Tianlun Huang (directors), Yanyi Li (screenwriter), Jiahou
Huang (producer), Jiaxin Yuan (art director), Xiang Chen, Jiayu Li and Zichen Tang (cinematographers),
Ningxuan Zhang (editor), and Shangheng Jiang (colourist). We further thank Huxin Peng and Liang Dong
(Tencent) for their help with data procurement.

\clearpage
\bibliographystyle{plainnat}
\bibliography{refs}

\clearpage
\appendix
\section{The Dimension System and Its Implementation}
\label{app:dims}

\dimsys{} defines $66$ leaf dimensions, and the $63$ scored in this report are those $66$ less the
three retired for a near-zero human baseline (\cref{sec:step2}). Each leaf dimension carries a
five-level decidable rubric and its own attribution-tag vocabulary, which humans and the automated
scorer share; the vocabularies are listed in \cref{sec:tagvocab} and the rubric texts are too long
to reproduce here. Per-dimension scores and agreement are in the per-stage tables of
\cref{sec:perstage}.

\label{app:impl}
\label{app:tools}
\label{app:steps}

\subsection{Implementation of every leaf dimension}
\label{sec:impl}

Every leaf dimension is implemented the same way, and the whole of it is declarative. The schema
entry for a dimension carries three things: a five-level rubric, a closed tag vocabulary, and a
one-line \emph{method} stating what is to be checked; the router scans that line for keywords and
mounts the evidence producers it names, so what a dimension is defined to check determines what gets
measured for it, with no per-dimension code. Evidence then reaches the automated scorer on two
paths. \textbf{Routed measurements} run unconditionally: zero to three of the $29$ producers per
dimension. \textbf{Model-callable tools} can be invoked over several rounds, and a dimension only gets
them if it declares them, which leaves $15$ of the $17$ in the registry reachable. No
\texttt{pixel\_sensitive} dimension declares a set containing the pixel-altering tools, and the router
withholds them there in any case, so image quality is never assessed on a retouched frame. Scoring then emits a band, a tag drawn from that dimension's
list, and a reason.

Reading the tables: \textbf{implementation} paraphrases the method line the router reads;
\textbf{routed evidence} is what \texttt{route\_tools} selects for it, traced by executing the router
rather than by reading the config, with ``---'' marking a dimension for which it selects no producer;
\textbf{model-callable} is the resolved tool set, with ``---'' marking the dimensions that cannot be
examined more closely on request. Dimension names follow \cref{app:dims},
and tool names are shortened, with \cref{tab:tools} giving them in full. One routed producer is
\fail{inert}: the router mounts the producer, but it returns unavailable on every item of that
stage, so the dimension is scored by the judging model alone. \dc{S-C4}'s method line names CSD, an
image model, and the storyboard stage has no pixels; \tl{identity\_sim} is suspended across shots and
on the assembled episode, where identity matching splits one character or merges two often enough to
be unusable. One producer is not routed at all: \tl{cut\_detect} is computed once per video or episode
item and enters every prompt with the base context.

The model-callable column names the tool set the schema declares for that dimension, with the
number of tools in it in parentheses. \tl{detail(2)} is \tl{zoom\_in} with \tl{draw\_bbox};
\tl{count(3)} and \tl{scene(3)} add \tl{count\_marker} to it; \tl{fine(4)} instead adds
\tl{sharpen} and \tl{super\_resolution}, and \tl{lowlight(4)} adds \tl{enhance\_contrast}. On the
video side every set begins with \tl{video\_probe} and \tl{extract\_frame}, since a video has no
still to zoom into until one is pulled: \tl{v\_frame(4)} adds \tl{zoom\_in} and \tl{draw\_bbox},
\tl{v\_count(5)}, \tl{v\_text(5)} and \tl{v\_cut(5)} add \tl{count\_marker}, \tl{ocr\_frame} and
\tl{detect\_scenes} respectively, and \tl{v\_sync(5)}, \tl{v\_audio(4)} and \tl{v\_speech(7)} are
built around the audio tools, dropping the annotation and zoom tools they do not need. \Cref{tab:tools} gives every tool in full.

\subsection*{\textcircled{1} Storyboard design}

{\footnotesize\setlength{\tabcolsep}{4pt}\renewcommand{\arraystretch}{1.28}
\setlength{\LTpre}{2pt}\setlength{\LTpost}{10pt}
\begin{longtable}{@{}p{0.95cm} >{\raggedright\arraybackslash}p{2.75cm}
  >{\centering\arraybackslash}p{0.6cm}
  >{\raggedright\arraybackslash}p{\dimexpr\textwidth-8.9cm-10\tabcolsep\relax}
  >{\raggedright\arraybackslash}p{2.9cm} >{\raggedright\arraybackslash}p{1.7cm}@{}}
\toprule
\textbf{Code} & \textbf{Dimension} & \textbf{Axis} & \textbf{Implementation} & \textbf{Routed evidence} & \textbf{Model-callable} \\
\midrule
\endfirsthead
\toprule
\textbf{Code} & \textbf{Dimension} & \textbf{Axis} & \textbf{Implementation} & \textbf{Routed evidence} & \textbf{Model-callable} \\
\midrule
\endhead
\bottomrule
\endlastfoot
\dc{S-A1} & Event coverage & \axF{} & master agent extracts beats → sub-agent aligns per shot for coverage & \tl{event\_align} & \textcolor{rulegray}{---} \\
\dc{S-A2} & Dialogue fidelity & \axF{} & dialogue text alignment (edit distance + embedding similarity) & \tl{dialogue\_align} & \textcolor{rulegray}{---} \\
\dc{S-B1} & Description executability & \axP{} & VLM: phrasing clarity / parsable content / spatial-physical check & \textcolor{rulegray}{---} & \textcolor{rulegray}{---} \\
\dc{S-B2} & Restraint in additions & \axP{} & LLM per shot: surplus detail + transition elements + key gaps; overstep or consistency harm & \tl{supplement\_audit} & \textcolor{rulegray}{---} \\
\dc{S-C1} & Narrative flow & \axE{} & LLM: cross-scene hard cuts + sequence breaks + shot order + OS misplacement; seven-form banding & \tl{narrative\_flow\_audit} & \textcolor{rulegray}{---} \\
\dc{S-C2} & Shooting rhythm & \axE{} & action-verb count and duration spread per shot + VLM feasibility & \tl{shot\_stats} & \textcolor{rulegray}{---} \\
\dc{S-C3} & Emotional expression & \axE{} & LLM per beat: eight audiovisual channels + director calibration & \tl{emotion\_beats} & \textcolor{rulegray}{---} \\
\dc{S-C4} & Audiovisual style treatment & \axE{} & shot-language classes + genre lexicon + director calibration & \tl{style\_csd}\,\fail{(inert)} & \textcolor{rulegray}{---} \\
\end{longtable}}\addtocounter{table}{-1}

\subsection*{\textcircled{2} Single keyframe}

{\footnotesize\setlength{\tabcolsep}{4pt}\renewcommand{\arraystretch}{1.28}
\setlength{\LTpre}{2pt}\setlength{\LTpost}{10pt}
\begin{longtable}{@{}p{0.95cm} >{\raggedright\arraybackslash}p{2.75cm}
  >{\centering\arraybackslash}p{0.6cm}
  >{\raggedright\arraybackslash}p{\dimexpr\textwidth-8.9cm-10\tabcolsep\relax}
  >{\raggedright\arraybackslash}p{2.9cm} >{\raggedright\arraybackslash}p{1.7cm}@{}}
\toprule
\textbf{Code} & \textbf{Dimension} & \textbf{Axis} & \textbf{Implementation} & \textbf{Routed evidence} & \textbf{Model-callable} \\
\midrule
\endfirsthead
\toprule
\textbf{Code} & \textbf{Dimension} & \textbf{Axis} & \textbf{Implementation} & \textbf{Routed evidence} & \textbf{Model-callable} \\
\midrule
\endhead
\bottomrule
\endlastfoot
\dc{I-A1} & Technical quality & \axQ{} & zoom to check three defect classes; sharpness dominant & \textcolor{rulegray}{---} & \tl{zoom(1)} \\
\dc{I-A2} & Aesthetic style & \axQ{} & CSD style purity + genre lexicon; colour harmony, cheap AI look & \tl{style\_csd} & \textcolor{rulegray}{---} \\
\dc{I-B1} & Subject attributes & \axF{} & SAM segmentation + zoom, 0/1/2 per facet & \tl{sam\_count} & \tl{count(3)} \\
\dc{I-B2} & Action and interaction & \axF{} & per-facet scoring & \textcolor{rulegray}{---} & \tl{detail(2)} \\
\dc{I-B3} & Scene and spatial layout & \axF{} & scene-type check + programmatic bearing match & \tl{scene\_consistency} & \tl{count(3)} \\
\dc{I-B4} & Shot framing and visual style & \axF{} & shot-attribute classes + composition, lighting + CSD style & \tl{style\_csd} \tl{framing\_check} & \textcolor{rulegray}{---} \\
\dc{I-C1} & Face identity & \axF{} & ArcFace for live action; anime-ID model or DINOv3 feature similarity + CIDS~\citep{dinov3,vistorybench}, selected by this audit & \tl{identity\_sim} & \tl{fine(4)} \\
\dc{I-C2} & Appearance and costume & \axF{} & zoom to compare against the character sheet; accessories need zoom & \textcolor{rulegray}{---} & \tl{fine(4)} \\
\dc{I-C3} & Key element fidelity\,\fail{(retired)} & \axF{} & SAM crop of the prop region + embedding or VLM vs.\ reference & \tl{sam\_count} & \tl{detail(2)} \\
\dc{I-D1} & Character consistency across panels & \axC{} & pairwise CIDS / DINOv3 feature similarity, aggregated & \tl{identity\_sim} & \tl{fine(4)} \\
\dc{I-D2} & Subject--scene integration & \axC{} & CPR~\citep{vistorybench} / NaturalScore~\citep{opens2v} + light and perspective blend; adversarial I-D1 & \tl{copypaste\_var} & \tl{detail(2)} \\
\dc{I-D3} & Scene and layout vs.\ reference & \axC{} & cross-image scene-bank embedding + spatial-structure recheck & \tl{scene\_consistency} & \tl{scene(3)} \\
\dc{I-D4} & Screen-direction axis & \axC{} & per-shot subject facing and motion emitted + rule-based axis check & \tl{axis\_check} & \tl{detail(2)} \\
\dc{I-D5} & Style consistency & \axC{} & cross-image CSD style embedding + grid and style-clash checks & \tl{style\_csd} & \tl{detail(2)} \\
\dc{I-E1} & Human anatomy plausibility & \axP{} & body and hand keypoints + zoom for deformity + pose naturalness & \tl{body\_structure} & \tl{count(3)} \\
\dc{I-E2} & Environmental physics & \axP{} & per-item detection: gravity / collision / perspective / lighting & \textcolor{rulegray}{---} & \tl{detail(2)} \\
\end{longtable}}\addtocounter{table}{-1}

\subsection*{\textcircled{3} Episode keyframes}

{\footnotesize\setlength{\tabcolsep}{4pt}\renewcommand{\arraystretch}{1.28}
\setlength{\LTpre}{2pt}\setlength{\LTpost}{10pt}
\begin{longtable}{@{}p{0.95cm} >{\raggedright\arraybackslash}p{2.75cm}
  >{\centering\arraybackslash}p{0.6cm}
  >{\raggedright\arraybackslash}p{\dimexpr\textwidth-8.9cm-10\tabcolsep\relax}
  >{\raggedright\arraybackslash}p{2.9cm} >{\raggedright\arraybackslash}p{1.7cm}@{}}
\toprule
\textbf{Code} & \textbf{Dimension} & \textbf{Axis} & \textbf{Implementation} & \textbf{Routed evidence} & \textbf{Model-callable} \\
\midrule
\endfirsthead
\toprule
\textbf{Code} & \textbf{Dimension} & \textbf{Axis} & \textbf{Implementation} & \textbf{Routed evidence} & \textbf{Model-callable} \\
\midrule
\endhead
\bottomrule
\endlastfoot
\dc{MI-D1} & Cross-shot character consistency & \axC{} & pairwise cross-shot identity similarity, aggregated + gap-decay & \tl{identity\_sim}\,\fail{(inert)} & \tl{fine(4)} \\
\dc{MI-D2} & Subject--scene integration & \axC{} & alert on excess cross-shot framing similarity; adversarial to MI-D1 & \tl{framing\_check} \tl{copypaste\_var} & \tl{detail(2)} \\
\dc{MI-D3} & Cross-shot scene consistency & \axC{} & cross-shot scene-bank embedding + zoom on set dressing per item & \tl{scene\_consistency} & \tl{lowlight(4)} \\
\dc{MI-D4} & Spatial-orientation coherence\,\fail{(retired)} & \axC{} & per-shot facing, position, camera estimate; rule-based axis check & \tl{axis\_check} & \tl{detail(2)} \\
\dc{MI-D5} & Cross-shot style consistency & \axC{} & cross-shot CSD style embedding + style-break zoom; tone by scene & \tl{style\_csd} & \tl{detail(2)} \\
\end{longtable}}\addtocounter{table}{-1}

\subsection*{\textcircled{4} Shot video}

{\footnotesize\setlength{\tabcolsep}{4pt}\renewcommand{\arraystretch}{1.28}
\setlength{\LTpre}{2pt}\setlength{\LTpost}{10pt}
\begin{longtable}{@{}p{0.95cm} >{\raggedright\arraybackslash}p{2.75cm}
  >{\centering\arraybackslash}p{0.6cm}
  >{\raggedright\arraybackslash}p{\dimexpr\textwidth-8.9cm-10\tabcolsep\relax}
  >{\raggedright\arraybackslash}p{2.9cm} >{\raggedright\arraybackslash}p{1.7cm}@{}}
\toprule
\textbf{Code} & \textbf{Dimension} & \textbf{Axis} & \textbf{Implementation} & \textbf{Routed evidence} & \textbf{Model-callable} \\
\midrule
\endfirsthead
\toprule
\textbf{Code} & \textbf{Dimension} & \textbf{Axis} & \textbf{Implementation} & \textbf{Routed evidence} & \textbf{Model-callable} \\
\midrule
\endhead
\bottomrule
\endlastfoot
\dc{V-A1} & Image quality and material & \axQ{} & DOVER / MUSIQ perceptual quality + Laplacian sharpness + material fit to style & \tl{quality\_metric} & \tl{v\_frame(4)} \\
\dc{V-A2} & Aesthetic style and motion texture & \axQ{} & aesthetic regression + style vs.\ genre; optical-flow motion texture & \tl{quality\_metric} \tl{aesthetic\_metric} \tl{style\_csd} & \textcolor{rulegray}{---} \\
\dc{V-B1} & Subject attribute accuracy & \axF{} & frame sampling + SAM + detection counts; satisfaction rate & \tl{sam\_count} & \tl{v\_count(5)} \\
\dc{V-B2} & Shot framing and visual style & \axF{} & shot-attribute classes + composition and lighting + cut count & \tl{framing\_check} & \tl{v\_frame(4)} \\
\dc{V-B3} & Keyframe adherence & \axF{} & SSIM~\citep{ssim} / DINO vs.\ given frame + subject-box match & \tl{frame\_follow} & \tl{v\_frame(4)} \\
\dc{V-C1} & Character appearance consistency & \axC{} & per-frame consistency, flicker, gap-decay, costume; fidelity-gated & \tl{identity\_sim} & \tl{v\_frame(4)} \\
\dc{V-C2} & Scene and prop consistency & \axC{} & scene-consistency anchor + prop embedding or cross-shot match & \tl{scene\_consistency} & \tl{v\_frame(4)} \\
\dc{V-C3} & Style consistency & \axC{} & CSD style embedding per frame and cross-shot + style-break check & \tl{style\_csd} & \tl{v\_frame(4)} \\
\dc{V-C4} & Action consistency & \axC{} & action recognition and cross-shot pose match + action continuity & \tl{body\_structure} & \tl{v\_frame(4)} \\
\dc{V-C5} & Subject--scene integration & \axC{} & ACP-Var~\citep{muss} / CPR + light and perspective blend; adversarial V-C1 & \tl{copypaste\_var} & \tl{v\_frame(4)} \\
\dc{V-C6} & On-screen text stability & \axC{} & per-frame OCR of text regions + adjacent-frame edit distance & \tl{ocr\_track} & \tl{v\_text(5)} \\
\dc{V-D1} & Motion smoothness & \axP{} & inter-frame optical flow, less jitter & \tl{quality\_metric} & \tl{v\_frame(4)} \\
\dc{V-D2} & Dynamic range\,\fail{(retired)} & \axP{} & flow magnitude + cut density & \tl{quality\_metric} & \textcolor{rulegray}{---} \\
\dc{V-D3} & Physical plausibility & \axP{} & VideoPhy-AutoEval physical commonsense, then plausibility on the footage & \tl{videophy} & \tl{v\_frame(4)} \\
\dc{V-D4} & Causal and temporal plausibility & \axP{} & sub-agent per time window → master agent: causal order, completion & \textcolor{rulegray}{---} & \tl{v\_frame(4)} \\
\dc{V-E1} & Speech quality & \axE{} & DNSMOS quality + ASR WER / CER on lines + voiceprint timbre & \tl{dialogue\_align} \tl{asr\_wer} & \tl{v\_speech(7)} \\
\dc{V-E2} & Audio-visual sync & \axE{} & SyncNet lip-sync + ImageBind audio-visual alignment & \tl{syncnet} \tl{imagebind\_av} & \tl{v\_sync(5)} \\
\dc{V-E3} & Sound effects and music & \axE{} & fit of music and sound effects to scene emotion and on-screen action & \textcolor{rulegray}{---} & \tl{v\_audio(4)} \\
\dc{V-F1} & Composition & \axE{} & shot size / depth / blocking + subject-box head-crop check & \tl{framing\_check} & \textcolor{rulegray}{---} \\
\dc{V-F2} & Camera movement plausibility & \axE{} & camera-move intent, position emotion, momentum + cut density & \textcolor{rulegray}{---} & \textcolor{rulegray}{---} \\
\dc{V-F3} & Character performance & \axE{} & acting and expression naturalness & \textcolor{rulegray}{---} & \textcolor{rulegray}{---} \\
\end{longtable}}\addtocounter{table}{-1}

\subsection*{\textcircled{5} Episode video}

{\footnotesize\setlength{\tabcolsep}{4pt}\renewcommand{\arraystretch}{1.28}
\setlength{\LTpre}{2pt}\setlength{\LTpost}{10pt}
\begin{longtable}{@{}p{0.95cm} >{\raggedright\arraybackslash}p{2.75cm}
  >{\centering\arraybackslash}p{0.6cm}
  >{\raggedright\arraybackslash}p{\dimexpr\textwidth-8.9cm-10\tabcolsep\relax}
  >{\raggedright\arraybackslash}p{2.9cm} >{\raggedright\arraybackslash}p{1.7cm}@{}}
\toprule
\textbf{Code} & \textbf{Dimension} & \textbf{Axis} & \textbf{Implementation} & \textbf{Routed evidence} & \textbf{Model-callable} \\
\midrule
\endfirsthead
\toprule
\textbf{Code} & \textbf{Dimension} & \textbf{Axis} & \textbf{Implementation} & \textbf{Routed evidence} & \textbf{Model-callable} \\
\midrule
\endhead
\bottomrule
\endlastfoot
\dc{MV-C1} & Cross-shot character consistency & \axC{} & per-shot frame vs.\ first-appearance identity + gap-decay + costume & \tl{identity\_sim}\,\fail{(inert)} & \tl{v\_frame(4)} \\
\dc{MV-C2} & Cross-shot scene and props & \axC{} & cross-shot scene-bank embedding + prop-state embedding or match & \tl{scene\_consistency} & \tl{v\_frame(4)} \\
\dc{MV-C3} & Cross-shot style consistency & \axC{} & cross-shot CSD style embedding + style-break check & \tl{style\_csd} & \tl{v\_frame(4)} \\
\dc{MV-C4} & Cross-shot action continuity & \axC{} & cross-shot action and pose match + facing/position, camera, axis check & \tl{axis\_check} \tl{body\_structure} & \tl{v\_frame(4)} \\
\dc{MV-C5} & Subject--scene integration & \axC{} & alert on excess cross-shot framing similarity; adversarial to MV-C1 & \tl{framing\_check} \tl{copypaste\_var} & \tl{v\_frame(4)} \\
\dc{MV-C6} & Cross-shot text stability & \axC{} & OCR per shot's text region; same element compared across shots & \tl{ocr\_track} & \tl{v\_text(5)} \\
\dc{MV-D4} & Cross-shot causal/temporal plausibility & \axP{} & sub-agent in shot order → master agent judges causality + closure & \textcolor{rulegray}{---} & \tl{v\_frame(4)} \\
\dc{MV-E1} & Cross-shot speech quality & \axE{} & cross-shot voiceprint + audio-visual alignment + ASR continuity & \tl{imagebind\_av} \tl{asr\_wer} & \tl{v\_speech(7)} \\
\dc{MV-E3} & Cross-shot sound and music & \axE{} & cross-shot coherence and style consistency of music and sound effects & \textcolor{rulegray}{---} & \tl{v\_audio(4)} \\
\dc{MV-F3} & Cross-shot performance continuity & \axE{} & per-segment expression, emotion strength + transitions & \textcolor{rulegray}{---} & \textcolor{rulegray}{---} \\
\end{longtable}}\addtocounter{table}{-1}

\subsection*{\textcircled{6} Short drama}

{\footnotesize\setlength{\tabcolsep}{4pt}\renewcommand{\arraystretch}{1.28}
\setlength{\LTpre}{2pt}\setlength{\LTpost}{10pt}
\begin{longtable}{@{}p{0.95cm} >{\raggedright\arraybackslash}p{2.75cm}
  >{\centering\arraybackslash}p{0.6cm}
  >{\raggedright\arraybackslash}p{\dimexpr\textwidth-8.9cm-10\tabcolsep\relax}
  >{\raggedright\arraybackslash}p{2.9cm} >{\raggedright\arraybackslash}p{1.7cm}@{}}
\toprule
\textbf{Code} & \textbf{Dimension} & \textbf{Axis} & \textbf{Implementation} & \textbf{Routed evidence} & \textbf{Model-callable} \\
\midrule
\endfirsthead
\toprule
\textbf{Code} & \textbf{Dimension} & \textbf{Axis} & \textbf{Implementation} & \textbf{Routed evidence} & \textbf{Model-callable} \\
\midrule
\endhead
\bottomrule
\endlastfoot
\dc{O-A1} & Event completion & \axF{} & key-event list from the script → per-event completion on the cut & \textcolor{rulegray}{---} & \tl{v\_frame(4)} \\
\dc{O-B1} & Style & \axC{} & CSD whole-episode style embedding; tone grouped by scene & \tl{style\_csd} & \tl{v\_frame(4)} \\
\dc{O-B2} & Cross-episode consistency & \axC{} & character and scene-bank anchors + gap-decay & \tl{identity\_sim}\,\fail{(inert)} \tl{scene\_consistency} & \tl{v\_frame(4)} \\
\dc{O-B3} & Cross-episode progression & \axC{} & against script beats: motivated progression vs.\ arbitrary change & \textcolor{rulegray}{---} & \textcolor{rulegray}{---} \\
\dc{O-C1} & Watchability of the cut & \axE{} & calibrated whole-episode score & \textcolor{rulegray}{---} & \textcolor{rulegray}{---} \\
\dc{O-C2} & Pacing and transitions & \axE{} & human-cut distribution gap + transition naturalness; cut rhythm & \textcolor{rulegray}{---} & \tl{v\_cut(5)} \\
\end{longtable}}\addtocounter{table}{-1}

\subsection{The attribution-tag vocabulary}
\label{sec:tagvocab}

Any score below $5$, from an annotator or from the automated scorer, must select at least one tag
from the list its dimension declares; the two sides draw from the same list, which is what makes
\emph{what was seen} comparable with \emph{how it was scored}. The
$66$ dimensions declare $298$ tag slots drawn from $253$ distinct tags, and the corpus holds
$255{,}925$ human attributions over them. Every dimension additionally offers \emph{other}, which is
omitted from the tables below; a free-text reason is required alongside the tag in all cases.

\subsection*{\textcircled{1} Storyboard design}

{\scriptsize
\setlength{\LTpre}{2pt}\setlength{\LTpost}{8pt}
\begin{longtable}{@{}p{1.05cm} >{\raggedright\arraybackslash}p{3.7cm} >{\raggedright\arraybackslash}p{\dimexpr\textwidth-4.75cm-4\tabcolsep\relax}@{}}
\toprule
\textbf{Code} & \textbf{Dimension} & \textbf{Attribution tags} \\
\midrule
\endfirsthead
\toprule
\textbf{Code} & \textbf{Dimension} & \textbf{Attribution tags} \\
\midrule
\endhead
\bottomrule
\endlastfoot
\dc{S-A1} & Event coverage & key event omitted $\cdot$ minor beat omitted $\cdot$ action or voiceover omitted $\cdot$ events do not join up $\cdot$ surplus plot invented $\cdot$ plot left vague \\
\dc{S-A2} & Dialogue fidelity & key line omitted $\cdot$ line rewritten $\cdot$ wrong speaker $\cdot$ wrong emotional register \\
\dc{S-B1} & Description executability & ambiguous description $\cdot$ viewpoint unclear $\cdot$ blocking unclear $\cdot$ several actions in one shot $\cdot$ shot carries too little $\cdot$ verbose padding $\cdot$ blocking not physically possible $\cdot$ blocking too complex \\
\dc{S-B2} & Restraint in additions & key detail not supplied $\cdot$ transition not supplied $\cdot$ surplus detail added $\cdot$ implausible content invented \\
\dc{S-C1} & Narrative flow & abrupt transition $\cdot$ jarring join $\cdot$ order reversed within a shot $\cdot$ misplaced voiceover $\cdot$ mixed shot sizes in one shot $\cdot$ setup gives away the payoff $\cdot$ redundant narration $\cdot$ self-contradiction \\
\dc{S-C2} & Shooting rhythm & too much action in one shot $\cdot$ over-fragmented cutting $\cdot$ key scene under-covered $\cdot$ transitional scene drags $\cdot$ no reaction shot $\cdot$ monotonous camera work \\
\dc{S-C3} & Emotional expression & flat, no tension $\cdot$ emotion not externalised $\cdot$ conflict poorly staged \\
\dc{S-C4} & Audiovisual style treatment & style mismatched to genre $\cdot$ inconsistent style $\cdot$ key scene not dramatised $\cdot$ wrong shot grammar \\
\end{longtable}}\addtocounter{table}{-1}

\subsection*{\textcircled{2} Single keyframe}

{\scriptsize
\setlength{\LTpre}{2pt}\setlength{\LTpost}{8pt}
\begin{longtable}{@{}p{1.05cm} >{\raggedright\arraybackslash}p{3.7cm} >{\raggedright\arraybackslash}p{\dimexpr\textwidth-4.75cm-4\tabcolsep\relax}@{}}
\toprule
\textbf{Code} & \textbf{Dimension} & \textbf{Attribution tags} \\
\midrule
\endfirsthead
\toprule
\textbf{Code} & \textbf{Dimension} & \textbf{Attribution tags} \\
\midrule
\endhead
\bottomrule
\endlastfoot
\dc{I-A1} & Technical quality & blurred $\cdot$ visible noise $\cdot$ artefacts and colour blocks $\cdot$ ghosting $\cdot$ watermark or garbled glyphs $\cdot$ structural distortion \\
\dc{I-A2} & Aesthetic style & aesthetic mismatched to genre $\cdot$ cluttered frame $\cdot$ unbalanced composition $\cdot$ colour and lighting clash $\cdot$ mixed art styles $\cdot$ cheap AI look \\
\dc{I-B1} & Subject attributes & subject missing $\cdot$ subject feature wrong or missing $\cdot$ wrong subject colour $\cdot$ wrong subject material $\cdot$ wrong subject count $\cdot$ extra subject $\cdot$ wrong subject position $\cdot$ left/right mirrored $\cdot$ wrong expression \\
\dc{I-B2} & Action and interaction & action not performed $\cdot$ wrong action $\cdot$ action mirrored $\cdot$ wrong interaction target $\cdot$ stiff pose \\
\dc{I-B3} & Scene and spatial layout & wrong scene type $\cdot$ wrong bearing $\cdot$ wrong occlusion $\cdot$ wrong relative position $\cdot$ scene element missing \\
\dc{I-B4} & Shot framing and visual style & wrong shot size or viewpoint $\cdot$ wrong composition $\cdot$ wrong lighting mood $\cdot$ wrong art style \\
\dc{I-C1} & Face identity & features do not match $\cdot$ wrong face shape $\cdot$ wrong expression $\cdot$ distinguishing feature lost $\cdot$ identity drift $\cdot$ face breakdown \\
\dc{I-C2} & Appearance and costume & wrong hairstyle $\cdot$ wrong hair colour $\cdot$ wrong garment style $\cdot$ wrong garment colour $\cdot$ accessory missing or wrong $\cdot$ wrong to station or period \\
\dc{I-C3} & Key element fidelity \fail{(retired)} & prop drawn as something else $\cdot$ wrong prop form $\cdot$ wrong prop colour $\cdot$ wrong prop material $\cdot$ wrong prop count $\cdot$ prop missing \\
\dc{I-D1} & Character consistency across panels & face differs across panels $\cdot$ costume differs across panels $\cdot$ accessories differ across panels $\cdot$ identity drift $\cdot$ face breakdown \\
\dc{I-D2} & Subject--scene integration & cut-out, pasted-on look $\cdot$ subject and scene lighting mismatch $\cdot$ floating, not grounded $\cdot$ pose or expression repeated $\cdot$ panels copy-pasted \\
\dc{I-D3} & Scene and layout vs.\ reference & spatial structure changed $\cdot$ object bearing changed $\cdot$ light direction inconsistent $\cdot$ prop placement changed \\
\dc{I-D4} & Screen-direction axis & characters swapped left/right $\cdot$ position drift $\cdot$ excessive displacement across panels $\cdot$ eyelines inconsistent $\cdot$ motion direction reversed \\
\dc{I-D5} & Style consistency & art style differs across panels $\cdot$ colour grade differs within a scene $\cdot$ animation and live action mixed \\
\dc{I-E1} & Human anatomy plausibility & extra or missing fingers $\cdot$ limbs twisted or fused $\cdot$ limb clipping $\cdot$ limb missing $\cdot$ facial proportions off $\cdot$ impossible pose $\cdot$ stiff, unnatural pose \\
\dc{I-E2} & Environmental physics & object floating $\cdot$ clipping through geometry $\cdot$ perspective or scale distorted $\cdot$ lighting or reflection contradictory \\
\end{longtable}}\addtocounter{table}{-1}

\subsection*{\textcircled{3} Episode keyframes}

{\scriptsize
\setlength{\LTpre}{2pt}\setlength{\LTpost}{8pt}
\begin{longtable}{@{}p{1.05cm} >{\raggedright\arraybackslash}p{3.7cm} >{\raggedright\arraybackslash}p{\dimexpr\textwidth-4.75cm-4\tabcolsep\relax}@{}}
\toprule
\textbf{Code} & \textbf{Dimension} & \textbf{Attribution tags} \\
\midrule
\endfirsthead
\toprule
\textbf{Code} & \textbf{Dimension} & \textbf{Attribution tags} \\
\midrule
\endhead
\bottomrule
\endlastfoot
\dc{MI-D1} & Cross-shot character consistency & face differs across shots $\cdot$ costume differs across shots $\cdot$ accessories differ across shots $\cdot$ identity drift $\cdot$ face swap across shots \\
\dc{MI-D2} & Subject--scene integration & cut-out, pasted-on look $\cdot$ subject and scene lighting mismatch $\cdot$ shots look alike $\cdot$ composition and pose unchanged \\
\dc{MI-D3} & Cross-shot scene consistency & layout change $\cdot$ furnishing change $\cdot$ lighting or colour shift $\cdot$ disordered scene cuts \\
\dc{MI-D4} & Spatial-orientation coherence \fail{(retired)} & axis crossed, left/right swapped $\cdot$ eyelines inconsistent $\cdot$ motion direction reversed $\cdot$ abrupt camera-position change \\
\dc{MI-D5} & Cross-shot style consistency & art style jumps $\cdot$ colour grade differs within a scene $\cdot$ lighting differs within a scene $\cdot$ inconsistent texture \\
\end{longtable}}\addtocounter{table}{-1}

\subsection*{\textcircled{4} Shot video}

{\scriptsize
\setlength{\LTpre}{2pt}\setlength{\LTpost}{8pt}
\begin{longtable}{@{}p{1.05cm} >{\raggedright\arraybackslash}p{3.7cm} >{\raggedright\arraybackslash}p{\dimexpr\textwidth-4.75cm-4\tabcolsep\relax}@{}}
\toprule
\textbf{Code} & \textbf{Dimension} & \textbf{Attribution tags} \\
\midrule
\endfirsthead
\toprule
\textbf{Code} & \textbf{Dimension} & \textbf{Attribution tags} \\
\midrule
\endhead
\bottomrule
\endlastfoot
\dc{V-A1} & Image quality and material & blurred, low resolution $\cdot$ noise and artefacts $\cdot$ material distorted $\cdot$ waxy, plastic skin \\
\dc{V-A2} & Aesthetic style and motion texture & poor composition and colour $\cdot$ art style mismatched to genre $\cdot$ cheap AI look $\cdot$ cheap, coarse motion \\
\dc{V-B1} & Subject attribute accuracy & subject missing $\cdot$ appearance differs from the reference $\cdot$ wrong subject colour $\cdot$ wrong subject material $\cdot$ wrong subject count $\cdot$ extra subject \\
\dc{V-B2} & Shot framing and visual style & wrong shot size or viewpoint $\cdot$ wrong composition $\cdot$ camera move does not follow the instruction $\cdot$ wrong lighting mood $\cdot$ wrong art style \\
\dc{V-B3} & Keyframe adherence & deviates from the given keyframe $\cdot$ drifts from the reference during motion \\
\dc{V-C1} & Character appearance consistency & subject flicker $\cdot$ slight appearance change $\cdot$ face breakdown or swap $\cdot$ garment colour change $\cdot$ garment style change \\
\dc{V-C2} & Scene and prop consistency & background layout change $\cdot$ furnishing change $\cdot$ colour grade change $\cdot$ prop state jumps \\
\dc{V-C3} & Style consistency & art style jumps $\cdot$ colour grade jumps \\
\dc{V-C4} & Action consistency & action jumps at the join $\cdot$ stiff action join $\cdot$ action contradicts itself \\
\dc{V-C5} & Subject--scene integration & cut-out, pasted-on look $\cdot$ subject and scene lighting mismatch $\cdot$ floating, not grounded $\cdot$ pose repeated and static $\cdot$ frames duplicated \\
\dc{V-C6} & On-screen text stability & garbled glyphs $\cdot$ glyph deformation $\cdot$ text flicker \\
\dc{V-D1} & Motion smoothness & stutter and dropped frames $\cdot$ teleporting $\cdot$ jitter \\
\dc{V-D2} & Dynamic range \fail{(retired)} & too static $\cdot$ motion amplitude distorted \\
\dc{V-D3} & Physical plausibility & weightless floating $\cdot$ clipping $\cdot$ unconvincing collision $\cdot$ unnatural inertia $\cdot$ appears or vanishes from nowhere \\
\dc{V-D4} & Causal and temporal plausibility & effect precedes cause $\cdot$ events out of order $\cdot$ cause and effect do not match $\cdot$ action not performed $\cdot$ action wrong or incomplete \\
\dc{V-E1} & Speech quality & robotic, vocoded $\cdot$ noise and clipping $\cdot$ muffled $\cdot$ unstable loudness $\cdot$ line misread or words dropped $\cdot$ lines out of order $\cdot$ wrong speaker $\cdot$ unintelligible $\cdot$ timbre drift \\
\dc{V-E2} & Audio-visual sync & lips out of sync $\cdot$ mouth moves, no sound $\cdot$ sound effect misaligned \\
\dc{V-E3} & Sound effects and music & music missing $\cdot$ music does not fit the emotion $\cdot$ sound effect missing $\cdot$ sound effect does not fit the picture \\
\dc{V-F1} & Composition & wrong shot size $\cdot$ subject not salient $\cdot$ flat, no depth $\cdot$ foreground and background confused \\
\dc{V-F2} & Camera movement plausibility & almost no camera movement $\cdot$ aimless drifting camera $\cdot$ gratuitous camera moves \\
\dc{V-F3} & Character performance & blank, monotonous expression $\cdot$ wrong eyeline $\cdot$ overacted, out of character \\
\end{longtable}}\addtocounter{table}{-1}

\subsection*{\textcircled{5} Episode video}

{\scriptsize
\setlength{\LTpre}{2pt}\setlength{\LTpost}{8pt}
\begin{longtable}{@{}p{1.05cm} >{\raggedright\arraybackslash}p{3.7cm} >{\raggedright\arraybackslash}p{\dimexpr\textwidth-4.75cm-4\tabcolsep\relax}@{}}
\toprule
\textbf{Code} & \textbf{Dimension} & \textbf{Attribution tags} \\
\midrule
\endfirsthead
\toprule
\textbf{Code} & \textbf{Dimension} & \textbf{Attribution tags} \\
\midrule
\endhead
\bottomrule
\endlastfoot
\dc{MV-C1} & Cross-shot character consistency & face differs across shots $\cdot$ face breakdown or swap $\cdot$ garment colour change $\cdot$ garment style change $\cdot$ identity drift \\
\dc{MV-C2} & Cross-shot scene and props & background layout change $\cdot$ furnishing change $\cdot$ colour grade change $\cdot$ prop state jumps \\
\dc{MV-C3} & Cross-shot style consistency & art style jumps to another medium $\cdot$ colour grade differs within a scene $\cdot$ animation and live action mixed \\
\dc{MV-C4} & Cross-shot action continuity & action discontinuity at the join $\cdot$ action contradicts itself $\cdot$ axis crossed, left/right swapped $\cdot$ eyelines inconsistent $\cdot$ motion direction reversed $\cdot$ abrupt camera-position change \\
\dc{MV-C5} & Subject--scene integration & cut-out, pasted-on look $\cdot$ subject and scene lighting mismatch $\cdot$ shots look alike $\cdot$ camera and composition unchanged \\
\dc{MV-C6} & Cross-shot text stability & text changes across shots $\cdot$ garbled glyphs $\cdot$ glyph deformation \\
\dc{MV-D4} & Cross-shot causal/temporal plausibility & effect precedes cause across shots $\cdot$ events out of order across shots $\cdot$ cause and effect do not match across shots $\cdot$ action never completed across shots \\
\dc{MV-E1} & Cross-shot speech quality & timbre drift across shots $\cdot$ unstable loudness across shots $\cdot$ audio quality differs across shots $\cdot$ dialogue broken across shots \\
\dc{MV-E3} & Cross-shot sound and music & music drops out across shots $\cdot$ abrupt music change $\cdot$ music style jumps $\cdot$ sound effects differ across shots \\
\dc{MV-F3} & Cross-shot performance continuity & emotion breaks between shots $\cdot$ emotional intensity jumps $\cdot$ performance style shifts $\cdot$ out of character across shots \\
\end{longtable}}\addtocounter{table}{-1}

\subsection*{\textcircled{6} Short drama}

{\scriptsize
\setlength{\LTpre}{2pt}\setlength{\LTpost}{8pt}
\begin{longtable}{@{}p{1.05cm} >{\raggedright\arraybackslash}p{3.7cm} >{\raggedright\arraybackslash}p{\dimexpr\textwidth-4.75cm-4\tabcolsep\relax}@{}}
\toprule
\textbf{Code} & \textbf{Dimension} & \textbf{Attribution tags} \\
\midrule
\endfirsthead
\toprule
\textbf{Code} & \textbf{Dimension} & \textbf{Attribution tags} \\
\midrule
\endhead
\bottomrule
\endlastfoot
\dc{O-A1} & Event completion & key event missing $\cdot$ event not played out $\cdot$ main line broken $\cdot$ characters misidentified $\cdot$ invented, departs from the script \\
\dc{O-B1} & Style & style break between episodes $\cdot$ colour grade differs within a scene $\cdot$ inconsistent art style \\
\dc{O-B2} & Cross-episode consistency & character changes across episodes $\cdot$ signature feature inconsistent $\cdot$ scene changes across episodes $\cdot$ face swap across episodes \\
\dc{O-B3} & Cross-episode progression & character does not develop $\cdot$ costume does not change $\cdot$ appearance or scene changed arbitrarily $\cdot$ change unmotivated \\
\dc{O-C1} & Watchability of the cut & cut incomplete $\cdot$ looks cheap $\cdot$ too many hard defects to watch \\
\dc{O-C2} & Pacing and transitions & key scene rushed $\cdot$ transitional scene drags $\cdot$ uneven pacing $\cdot$ stiff transition \\
\end{longtable}}\addtocounter{table}{-1}

\subsection{The scoring prompt}
\label{sec:prompt}

There is one prompt template for all $66$ dimensions. Everything dimension-specific is injected into
fixed slots, which is what makes scoring auditable: two dimensions differing in score cannot differ
in framing. The skeleton, with the slots marked:

\begin{quote}\small\ttfamily\raggedright
You are a short-drama generation quality sub-agent. Score the single dimension
[\{code\} \{name\}] (axis \{axis\}, tier \{tier\}) on the five-band scale, and nothing else.\\[3pt]
[Five band anchors]: \{rubric[5..1]\}\\
{}[Per-shot checkpoints and banding rules]: \{guide, optional --- carries the checklist\}\\
{}[Applicability (decide N/A before scoring)]: \{na\_when, optional\}\\
{}[Attribution tags] (at least one required when the score is below 5): \{tags\}\\
{}[Automated method hint]: \{method\}\\
{}[Item] stage=\{\dots\} scope=\{\dots\} paradigm=\{\dots\}\\
{}[Source script (ground-truth reference, not the artefact under evaluation)]: \{script, conditional\}\\
{}[Objective tool evidence]: \{evidence, one line per measurement\}\\
{}[Hard scoring requirements (apply to every dimension)]: \{constant\}\\
Output strictly JSON only: \{"score", "confidence", "tags", "reason"\}
\end{quote}

Four pieces of the constant framing do real work. Two are quoted as they run. An anti-pattern rule
forbids the hedging that produces artificial distributions:
\emph{Never score using unevidenced, formulaic hedges such as ``by the score-compression principle'', ``to be conservative'', ``full marks are not usually given'' or ``good overall but still docking a point'', and never pull a score towards the middle or the extremes artificially. The score is determined only by real defects in the artefact and the evidence; make no distributional adjustment.}
A severity calibration makes the deduction proportional rather than cumulative:
\emph{The size of the deduction must match the defect's actual effect on the overall impression\dots\ With several blemishes, band by their combined effect on that impression rather than stacking a deduction for each one and ending up too low.}
The other two tie the score to the evidence: any score below $5$ must name a specific, checkable
defect and select at least one attribution tag from that dimension's vocabulary, and on the image
stage, where a checklist exists, the reason must open with the checklist conclusions in order rather
than summarise them.

\paragraph{The multi-round loop.}
Where a dimension declares model-callable tools, scoring runs as a loop rather than a single call.
Rounds $0$ to $n-1$ expose the tool schemas with the choice of whether to call them left open; each
tool result is appended to the history, and cropped images are fed back as the next turn's media. The
final round withdraws the tools entirely, which is what forces a verdict. Repeating the same call
with the same arguments more than twice returns an error, and a run that exhausts its rounds without
valid JSON gets one tool-free closing round before falling back to single-shot scoring.

\subsection{The evidence tool chain}
\label{app:toolchain}

The two registries behind those two paths are set out in \cref{tab:tools}, by family.

\begin{table}[htbp]
\centering
\caption{The evidence tools, by family: those the automated scorer may call itself, plus the routed
measurements that always run. Names are registry identifiers, shortened. No tool produces the rubric
score, though several return numbers of their own.}
\label{tab:tools}
\small
\begin{tabular}{@{}l p{11.2cm}@{}}
\toprule
\textbf{Family} & \textbf{Tools} \\
\midrule
Vision   & \texttt{zoom\_in} (crop and magnify), \texttt{draw\_bbox} (annotate a region),
           \texttt{count\_marker} (numbered point counting), and three
           pixel-altering ones withheld from \texttt{pixel\_sensitive} dimensions
           (\texttt{sharpen}, \texttt{super\_resolution}, \texttt{enhance\_contrast}). Grid
           panels are sliced into sub-panels before consistency is assessed. \\
Temporal & \texttt{extract\_frame} (sample by timestamp; extracted frames can be zoomed
           again), \texttt{detect\_scenes} (shot segmentation), \texttt{video\_probe}
           (container metadata). \\
Auditory & \texttt{asr\_segment} (transcription), \texttt{dnsmos} (no-reference audio
           quality), \texttt{detect\_audio\_events}, \texttt{extract\_subtitles} (burned-in
           subtitle track), \texttt{extract\_clip\_audio}, \texttt{ocr\_frame}. \\
Metric   & Routed rather than model-invoked: \texttt{videophy} (physical plausibility), \texttt{syncnet} $+$
           \texttt{imagebind\_av} (lip sync), \texttt{identity\_sim} (character consistency),
           \texttt{copypaste\_var} (human--scene integration, against pasted-on subjects),
           \texttt{ocr\_track} (on-screen text stability), \texttt{cut\_detect} (shot changes, injected
           with the base context rather than routed), and no-reference quality scorers; the
           aesthetic scorer is computed and then withheld at every stage. Several return a number of their
           own; whether that number is trusted is settled by the correlation test of
           \cref{sec:step2}, tool by tool, and most were withdrawn from anchoring by it. \\
\bottomrule
\end{tabular}
\end{table}

\paragraph{Cost and reliability.}
A median $14.6$ minutes of wall clock per item. Tools execute concurrently, so cumulative
machine time is considerably larger than that: the median tool time is $198.7$ seconds
\emph{per dimension}, with a median of $20$ sub-agent calls and about $19$ measurement calls
per job. Metric routing reaches $17$ of the $21$ video leaf dimensions. Of the eight primary
tools, only \texttt{ocr\_track} drops below $98.9\%$ availability, to $71.1\%$. Nearly all of
that cost is tool execution rather than inference, so additional reasoning rounds are close to
free while mounting another tool is not.

\subsection{Validating candidate metrics}
\label{sec:audit}

Every candidate metric was checked against the human consensus rather than adopted on the strength of
what it was built for. The audit was run once per granularity, each time with the tool battery that
applies there and against every human dimension that battery can speak to, over the full annotated
corpus at that granularity, with SRCC and $p$ computed within each generation paradigm as well as
pooled. A tool contributes several scalar fields rather than one (\texttt{quality\_metric} alone
returns MUSIQ, TOPIQ-NR, Laplacian sharpness, brightness and contrast), and every field was tested
on its own. Agreement settles one question only: whether a value may be used as a scoring anchor.
Whether a small model is kept at all is a second question, and it turns on whether the model supplies
a low-level perceptual signal that the judging model cannot obtain by looking.

\paragraph{Candidates tested.}
\emph{Structure and subject}: YOLO26n~\citep{yolo26} $+$ SAM2.1~\citep{sam2,sam21} for counting and
segmentation, YOLO26n-pose~\citep{yolo26} for anatomy, composition and screen direction.
\emph{Identity and consistency}: ArcFace~\citep{arcface}, OSNet Re-ID~\citep{osnet},
DINOv3 visual feature similarity~\citep{dinov3} for characters, DreamSim~\citep{dreamsim} for scenes,
CSD ViT-L~\citep{csd} for style. \emph{Quality and aesthetics}: MUSIQ~\citep{musiq},
TOPIQ-NR~\citep{topiq}, NIMA~\citep{nima}, DOVER and DOVER++~\citep{dover}, Laplacian sharpness,
RAFT~\citep{raft} optical flow. \emph{Audio}: Whisper~\citep{whisper} ASR (faster-whisper),
DNSMOS~\citep{dnsmos}, GE2E~\citep{ge2e} voiceprints (Resemblyzer), SyncNet~\citep{syncnet} lip-sync,
ImageBind~\citep{imagebind} audio-visual semantic alignment. \emph{Large-model
corroboration}: Q-Judger~\citep{qwenimagebench}, ArtiMuse~\citep{artimuse},
UniPercept~\citep{unipercept}, VideoPhy-AutoEval~\citep{videophy2}.

\paragraph{Result.}
Against the standard we set for treating a metric as a scoring anchor ($|\mathrm{SRCC}| \ge 0.3$
over at least $100$ items), most candidates fall short of that line, and the shortfall is
not confined to one family: structure and subject, identity and
consistency, quality and aesthetics, and the AIGC-specific discriminators are all in the list. Eight
representative cases give the range. On the general-purpose side NIMA aesthetics reaches $-0.152$, Laplacian sharpness
$-0.144$, TOPIQ-NR $-0.067$ and MUSIQ $-0.057$; on the AIGC-specific side Q-Judger reaches $0.004$,
ArtiMuse $0.030$, and UniPercept $0.190$ on its image-quality head and $0.110$ on its aesthetic one.

\paragraph{Why both halves fall short.}
In AI short drama the criterion for picture quality appears to be tied to genre and art direction
rather than being a property of the frame alone. A xianxia power fantasy pushes saturation
deliberately, for commercial appeal, while in other items that same exaggerated colour is the
clearest tell of an AI artefact. The same visual feature therefore cannot serve as a uniform
criterion, which is the most plausible source of the systematic misjudgement we see in both groups:
the general-purpose metrics are calibrated on natural photographs, and the AIGC-specific models
answer whether an image looks good rather than whether it does what the script asked for.

\paragraph{What we do with them.}
Rather than attaching the metrics unvalidated, we keep the small models as a source of low-level
evidence and do not let them produce a score. Where a value is shown to the judging model it is shown
as a measurement with a direction rather than as a calibrated band, the exceptions being the physics,
reference-similarity and voiceprint fields, which keep theirs. The aesthetic scorer is the one
candidate computed and never shown: it is dropped before the prompt at every stage. The suppression is per stage
rather than global: DOVER, for one, is weakly \emph{positive} at the video stage, so blocking it
there as well would have discarded a correctly-signed signal.

\subsection{Training with human annotations}
\label{app:train}

Training follows the SFT--GRPO procedure in \cref{sec:judgetrain}. Each training example is a
prompt for an item, evaluation axis and output form, with score and attribution supervision from
human annotations. Outputs retain \judger{}'s axis-organised structure, including leaf-level
scores, brief rationales and a composite score. The test items in \cref{sec:ablation} are excluded
from both training phases.

The keyframe and storyboard scorer omits the audio pathway. Attribution rewards use the human
annotations available for each item; missing annotations are not treated as failures. Storyboard
training omits the attribution reward.

With greedy decoding, inference takes $6$--$9$\,s without reasoning and $13$--$27$\,s with reasoning
per item--axis--output-form prompt. These timings describe individual prompts, not a full evaluation
across all applicable axes. They are therefore reported separately from \judger{}'s per-item
wall-clock time in \cref{app:toolchain}.

\section{Drama Manifest}
\label{app:dramas}

The script corpus comprises $20$ dramas of $10$ episodes each, spanning audience channels,
periods/themes and genres. It includes $10$ live-action and $10$ animated dramas, with $16$ in Chinese
dialogue and $4$ in English, so that every slice supports a stratified check (\cref{sec:slice-style},
\cref{sec:slice-lang}). Every stage of the benchmark uses the first three episodes of each
drama, $60$ episodes in total. From these scripts, \pipe{} generates storyboards, keyframes, shot videos and finished
dramas. \Cref{tab:dramas} lists the manifest.

Each drama has a shared cast list and episode outlines of three to five hundred characters. The
scripts follow Hongguo's shooting-script format, specifying scene numbers, locations and times, cast
members, actions, camera notes and dialogue with emotional cues.

\begin{table}[htbp]
\centering
\caption{Drama manifest. English titles are given first; Chinese-language titles carry a
romanised original. The four English-dialogue titles use Chinese scene descriptions and
differ only in dialogue and on-screen text, which is what isolates the language effect
from production difficulty.}
\label{tab:dramas}
\small
\setlength{\tabcolsep}{4pt}
\renewcommand{\arraystretch}{1.18}
\begin{tabularx}{\linewidth}{@{}l l >{\raggedright\arraybackslash}X c c r@{}}
\toprule
\textbf{Channel} & \textbf{Period / theme} & \textbf{Genre} &
\textbf{Lang.} & \textbf{Style} & \textbf{Ep.} \\
\midrule
\multicolumn{6}{@{}l}{\emph{Male-oriented}} \\
& Contemporary urban & War god returns --- \emph{Blade Reclaimed} (Gui Ren) & zh & animated & 3 \\
& Contemporary urban & Live-in son-in-law rises --- \emph{Bladeless Son-in-Law} (Zhuixu Wufeng) & zh & live & 3 \\
& Contemporary urban & Sudden-wealth flex --- \emph{The Squander Decree} (Sancai Ling) & zh & live & 3 \\
& Historical costume & Court intrigue --- \emph{Chronicle of Buried Injustice} (Chenyuan Lu) & zh & live & 3 \\
& Historical costume & Time-slip revenge --- \emph{A Ming Dynasty Hotpot Tale} (Mingchao Huoguo Ji) & zh & animated & 3 \\
& Fantasy premise & Post-apocalyptic survival --- \emph{Permafrost Bastion} (Dongtu Baolei) & zh & animated & 3 \\
& Fantasy premise & Rebirth revenge --- \emph{No Second Crossing} (Chonglai Budu) & zh & live & 3 \\
\addlinespace[2pt]
\multicolumn{6}{@{}l}{\emph{All audiences}} \\
& Fantasy premise & Isekai settlement --- \emph{Homesteading the Cretaceous} (Bai'e Kaihuang) & zh & animated & 3 \\
& Modern period & Campus youth --- \emph{The Year of Cicadas} (Nanian Chanming) & zh & live & 3 \\
& Modern period & Rural farming --- \emph{Granaries Full of Wheat} (Maidun Mancang) & zh & animated & 3 \\
\addlinespace[2pt]
\multicolumn{6}{@{}l}{\emph{Female-oriented}} \\
& Urban romance & Marriage first, love later --- \emph{It Had to Be You} (Pianpian Nuan Ni) & zh & live & 3 \\
& Urban romance & Swapped heiresses --- \emph{I Don't Want It} (Bu Xihan) & zh & animated & 3 \\
& Urban romance & Dynastic tragic romance --- \emph{Moon Understudy} (Ti Yue) & zh & animated & 3 \\
& Historical romance & Costume time-slip --- \emph{The Lie-Flat Consort} (Tangying Niangniang) & zh & live & 3 \\
& Historical romance & Forced possession --- \emph{I Insist on Taking You} (Pianyao Qiang Qing) & zh & animated & 3 \\
& Female lead, no CP & Coming of age --- \emph{Over the Mountain} (Fan Shan) & zh & live & 3 \\
\addlinespace[2pt]
\multicolumn{6}{@{}l}{\emph{Overseas, male-oriented}} \\
& Contemporary urban & Revenge rise --- \emph{The Janitor's Billion-Dollar Revenge} & en & animated & 3 \\
& Contemporary urban & Hidden billionaire --- \emph{The Bellboy Owns This Hotel} & en & live & 3 \\
\addlinespace[2pt]
\multicolumn{6}{@{}l}{\emph{Overseas, female-oriented}} \\
& Contemporary urban & Vampire romance --- \emph{The Vampire Lord's Forbidden Blood} & en & animated & 3 \\
& Contemporary urban & Contract marriage --- \emph{The Billionaire's Two-Year Wife} & en & live & 3 \\
\midrule
\multicolumn{5}{@{}l}{\textbf{Total}} & \textbf{60} \\
\bottomrule
\end{tabularx}
\end{table}

\begin{figure}[p]
\centering
\setlength{\tabcolsep}{1.1pt}
\renewcommand{\arraystretch}{0.6}
\newcommand{\exhfrm}[1]{\includegraphics[width=0.175\linewidth]{figures/img/#1}}

{\footnotesize\textbf{EP1} \ the contract expires, the stand-in exits \quad
 {\scriptsize\textcolor{inkgray}{$89.8$\,s}}}\\[2pt]
\begin{tabular}{@{}ccccc@{}}
\exhfrm{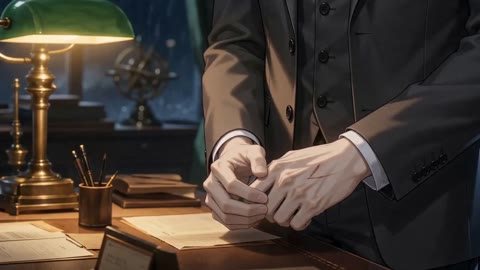} & \exhfrm{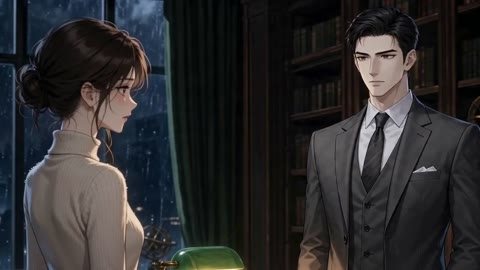} & \exhfrm{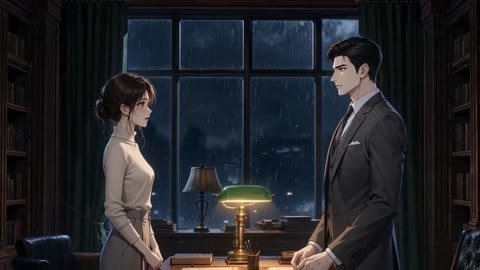} & \exhfrm{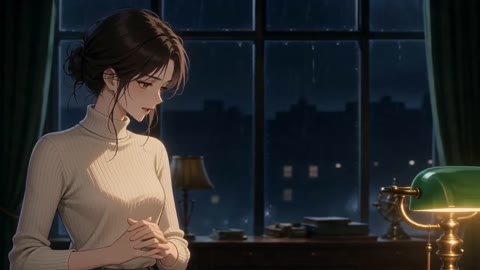} & \exhfrm{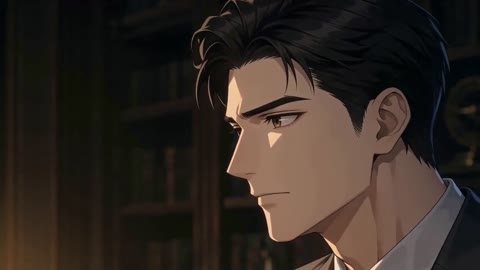} \\
\exhfrm{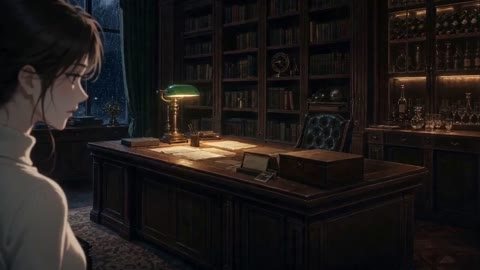} & \exhfrm{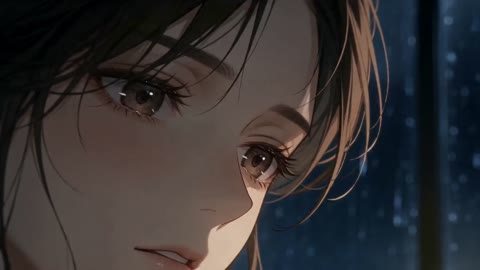} & \exhfrm{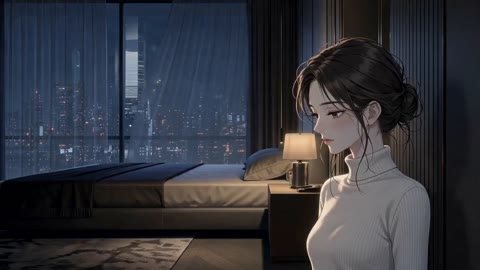} & \exhfrm{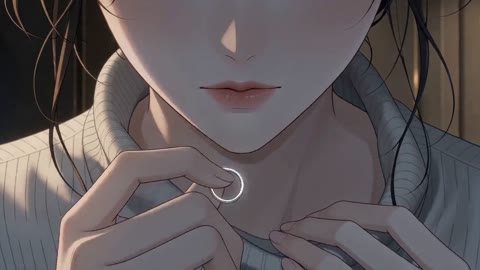} & \exhfrm{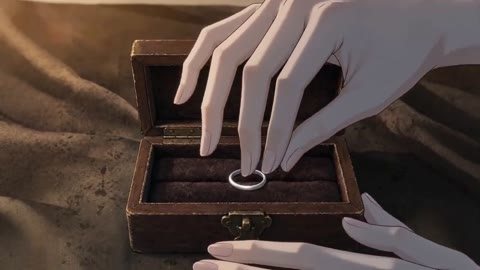} \\
\exhfrm{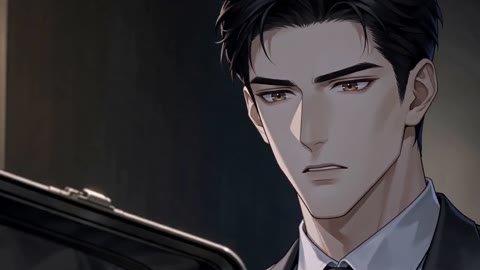} & \exhfrm{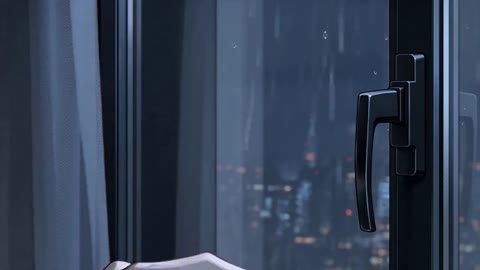} & \exhfrm{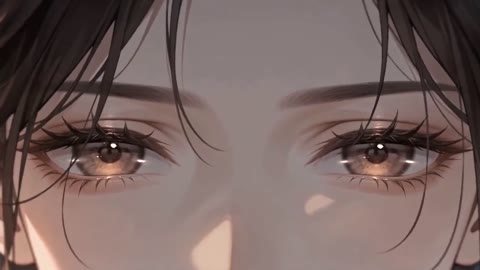} & \exhfrm{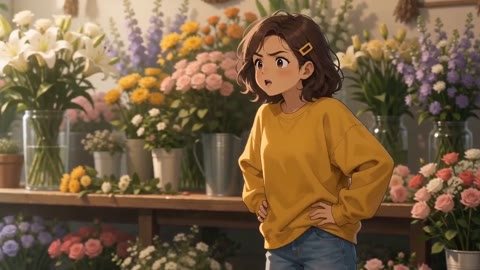} & \exhfrm{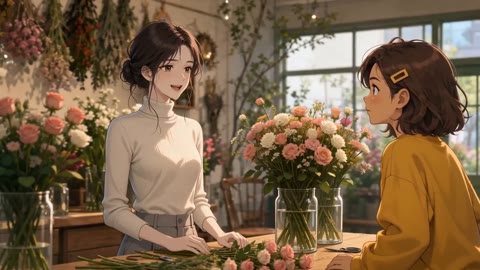} \\
\exhfrm{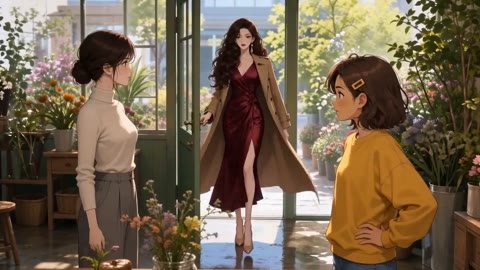} & \exhfrm{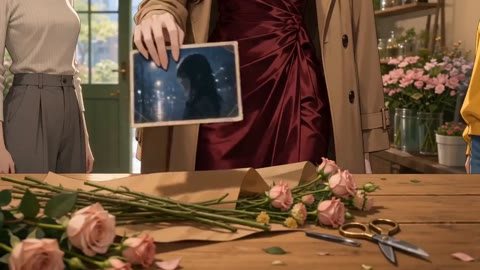} & \exhfrm{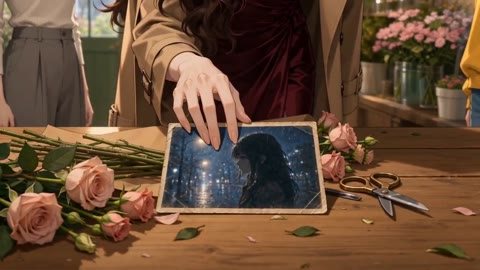} & \exhfrm{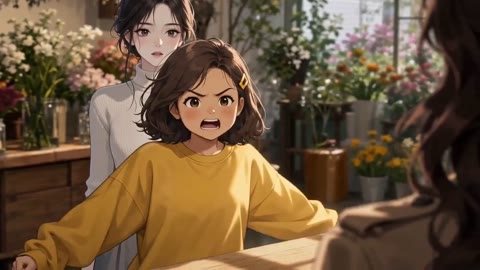} & \exhfrm{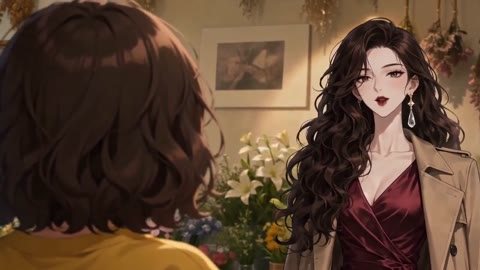} \\
\end{tabular}

\vspace{4pt}
{\footnotesize\textbf{EP2} \ ``what you stand in for is me'' \quad
 {\scriptsize\textcolor{inkgray}{$92.8$\,s}}}\\[2pt]
\begin{tabular}{@{}ccccc@{}}
\exhfrm{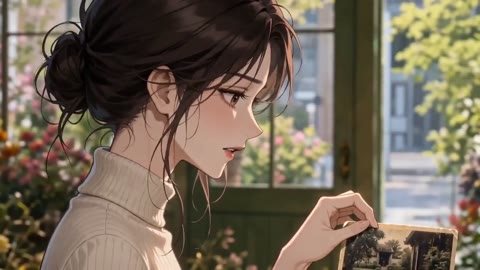} & \exhfrm{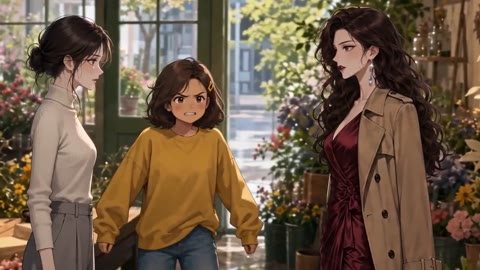} & \exhfrm{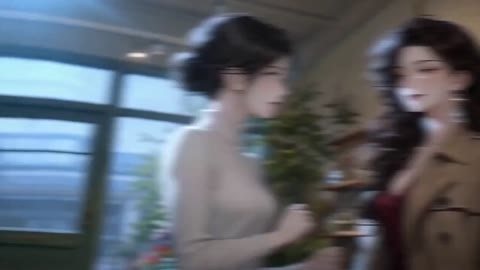} & \exhfrm{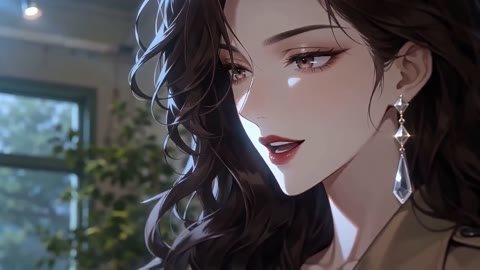} & \exhfrm{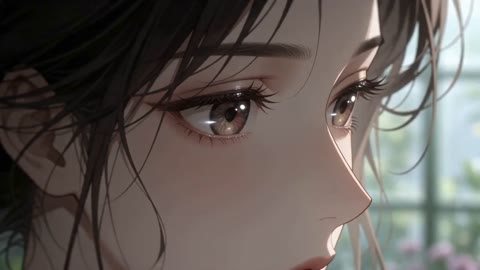} \\
\exhfrm{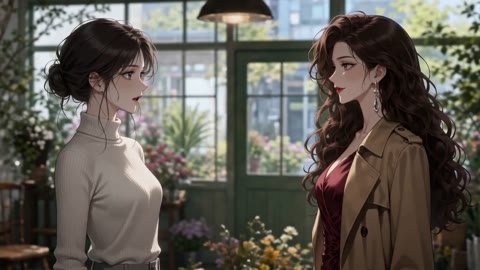} & \exhfrm{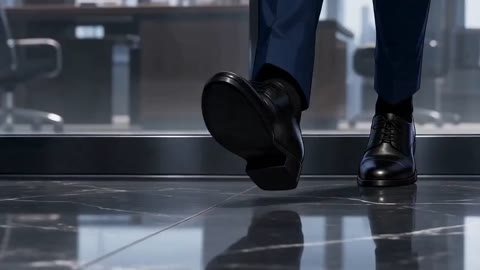} & \exhfrm{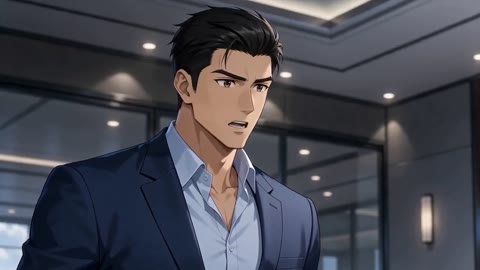} & \exhfrm{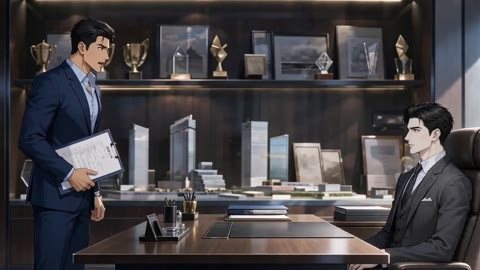} & \exhfrm{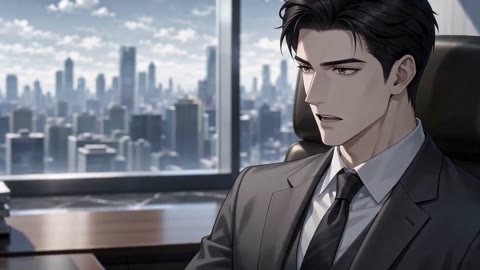} \\
\exhfrm{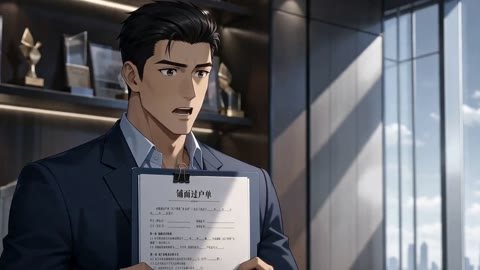} & \exhfrm{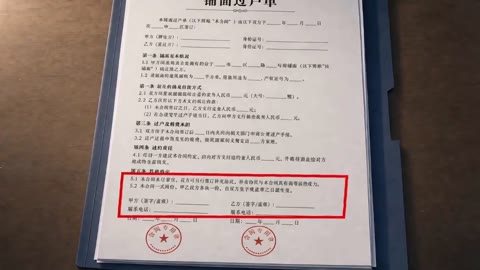} & \exhfrm{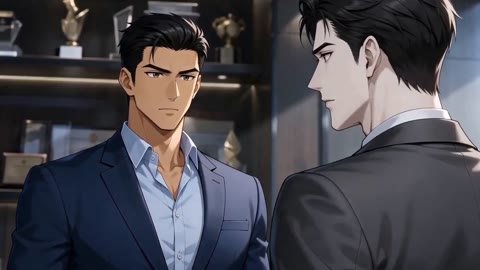} & \exhfrm{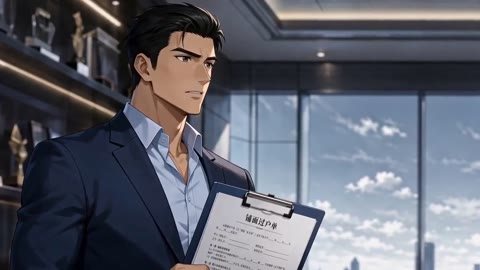} & \exhfrm{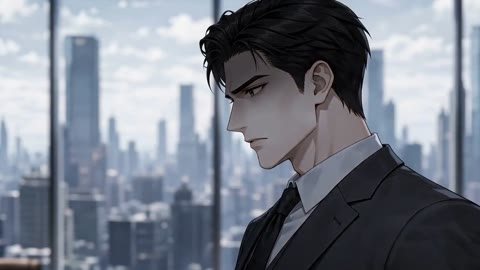} \\
\exhfrm{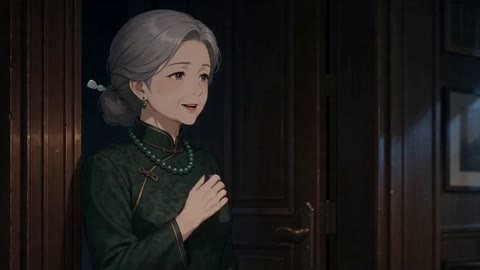} & \exhfrm{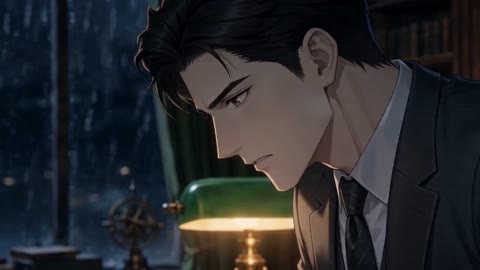} & \exhfrm{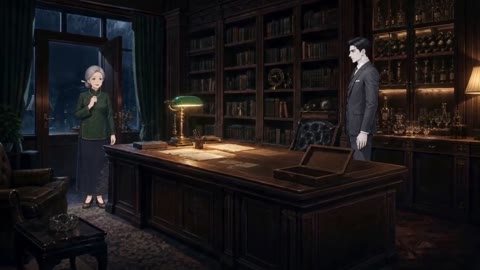} & \exhfrm{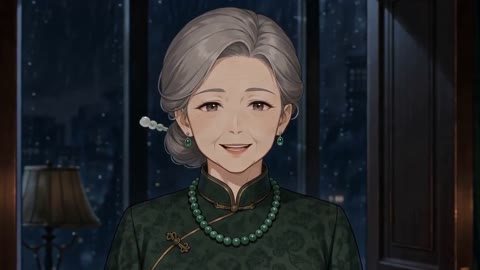} & \exhfrm{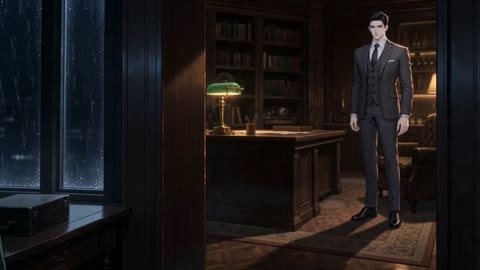} \\
\end{tabular}

\vspace{4pt}
{\footnotesize\textbf{EP3} \ the truth about to surface, and she is gone \quad
 {\scriptsize\textcolor{inkgray}{$116.0$\,s}}}\\[2pt]
\begin{tabular}{@{}ccccc@{}}
\exhfrm{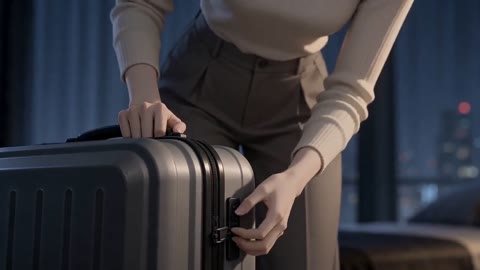} & \exhfrm{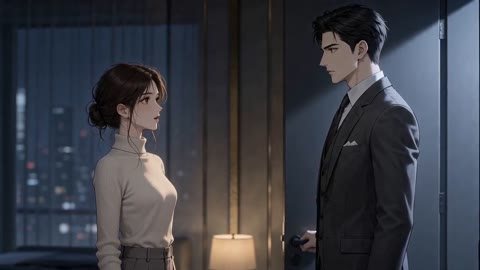} & \exhfrm{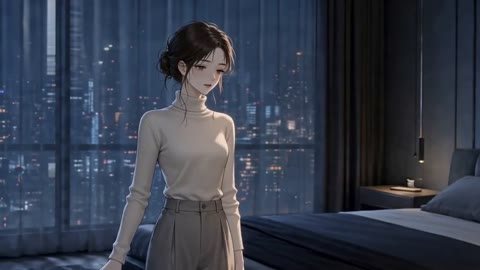} & \exhfrm{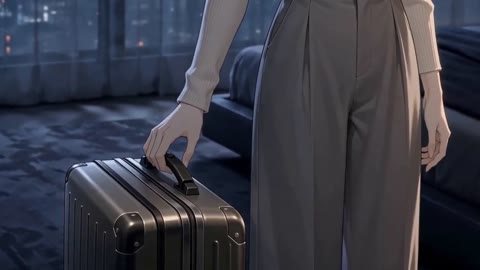} & \exhfrm{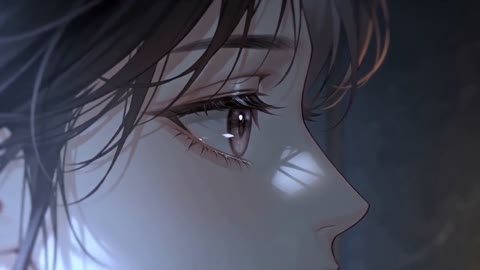} \\
\exhfrm{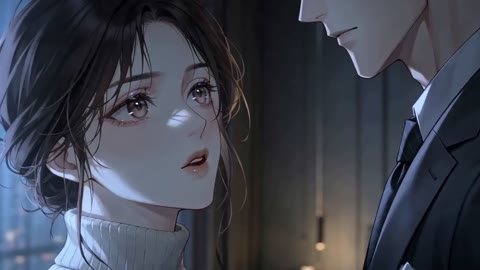} & \exhfrm{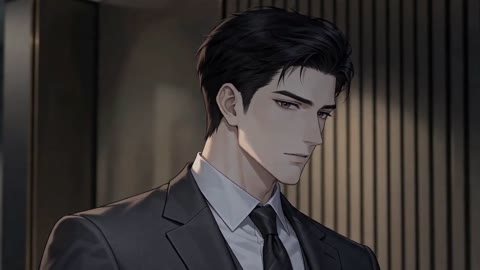} & \exhfrm{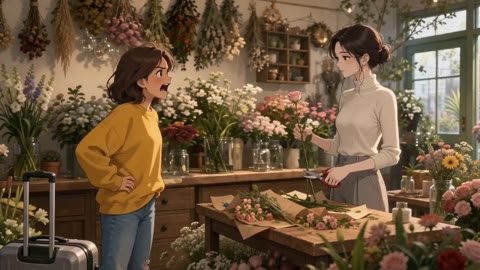} & \exhfrm{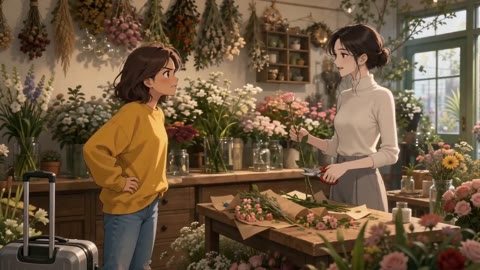} & \exhfrm{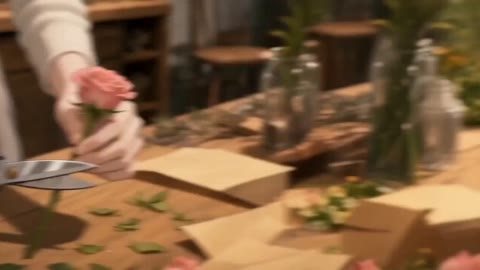} \\
\exhfrm{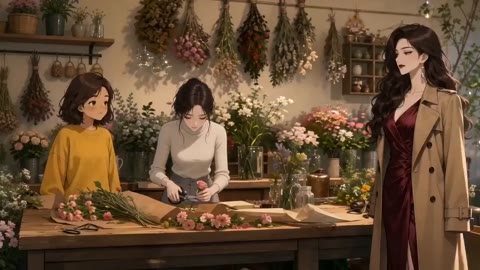} & \exhfrm{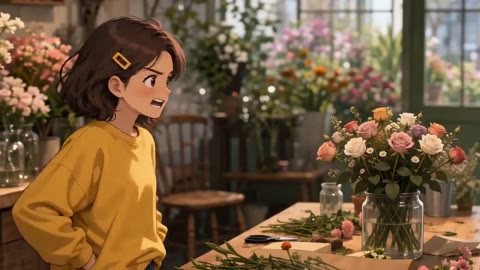} & \exhfrm{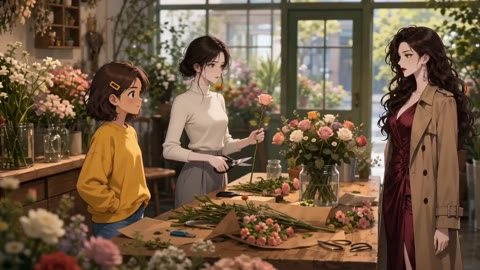} & \exhfrm{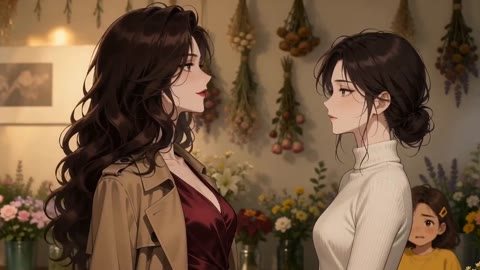} & \exhfrm{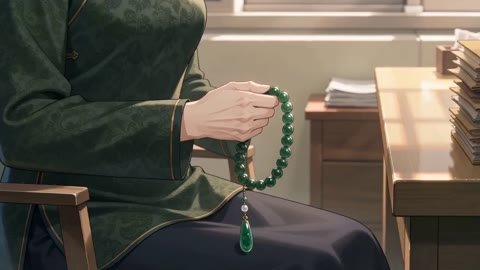} \\
\exhfrm{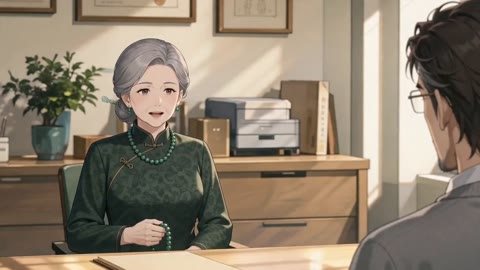} & \exhfrm{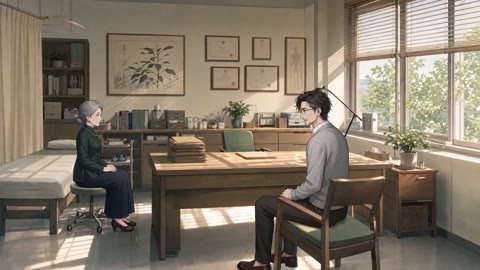} & \exhfrm{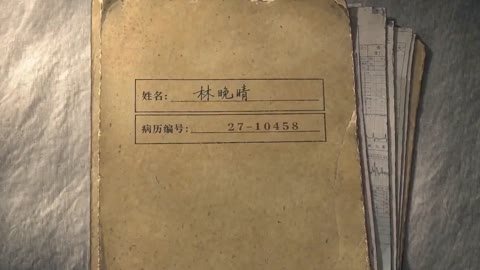} & \exhfrm{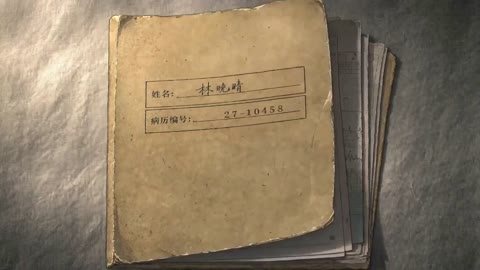} & \exhfrm{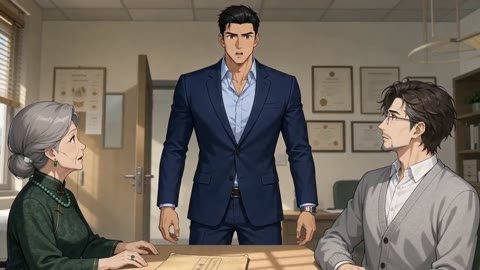} \\
\end{tabular}

\caption{\zhen{%
\pipe{}'s output for \emph{Moon Understudy} (Ti Yue), an animated, Chinese-dialogue urban romance
for a female audience (\cref{tab:dramas}). In a three-year contract marriage, the heroine believes
she is a substitute for a tycoon's first love, who moved abroad: her room and routines are modelled
after that woman. When the contract expires, she discovers that the photograph he has kept for three
years depicts her. She returns the keys before learning why his first love left. Every stage
uses each drama's first three episodes; each block shows $20$ evenly sampled frames in reading order.}{%
\pipe{} 生成的《替月》：动画、中文对白、女频都市情感（\cref{tab:dramas}）。三年契约婚姻中，女主认为自己是
富商出国初恋的替身，房间和生活习惯都按对方塑造。合约到期时，她发现他保存三年的照片中竟是自己；
尚未得知初恋离开的原因，她便退还钥匙离开。各环节均使用每部剧的前三集；各块展示一集均匀采样的 $20$ 帧，按阅读顺序排列。}}
\label{fig:exhibit}
\end{figure}

\section{Content Slices of the Automated Score}
\label{app:slices}
\label{sec:slices}

The three slices in this appendix vary the generated artefact: the drama's visual style, its dialogue
language, and the input paradigm the pipeline supplies to the model (\cref{tab:slices}). They
measure how far the automated absolute score moves as a result (\cref{tab:slicefx}). The three effects
order as follows: input paradigm is the largest, visual style second, dialogue language smallest.

\begin{table}[htbp]
\centering
\caption{The three content slices: what each one varies, and what it does to the automated score.}
\label{tab:slices}
\small
\begin{adjustbox}{max width=\linewidth}
\begin{tabular}{@{}l l l@{}}
\toprule
\textbf{Slice} & \textbf{What it varies} & \textbf{Effect on the automated score} \\
\midrule
Visual style      & live action against animation &
  $+0.08$ to $+0.22$ toward live action at five of six stages \\
Dialogue language & English against Chinese dialogue &
  at most $0.12$, and almost all of it on \axF{} \\
Input paradigm    & first/last frame, grid panel, multi-reference &
  up to $0.42$, reversing between the two keyframe stages \\
\bottomrule
\end{tabular}
\end{adjustbox}
\end{table}

\begin{table}[htbp]
\centering
\caption{\zhen{How far each content slice moves the automated composite, and which paradigm leads,
taken within the models that ran every level of that slice.}{三种内容切分各把自动化综合分移动了多少、
以及哪种范式领先，差值只在跑遍该切面所有档位的模型内部计算。}}
\label{tab:slicefx}
\small
\begin{tabular}{@{}l r r r l@{}}
\toprule
\textbf{\zhen{Granularity}{粒度}} &
\makecell[r]{\zhen{live action}{真人}\\[-1pt]\footnotesize $-$ \zhen{animation}{动画}} &
\makecell[r]{\zhen{English}{英文}\\[-1pt]\footnotesize $-$ \zhen{Chinese}{中文}} &
\makecell[r]{\zhen{paradigm}{范式}\\[-1pt]\footnotesize \zhen{spread}{极差}} &
\makecell[l]{\zhen{leading}{领先}\\[-1pt]\footnotesize \zhen{paradigm}{范式}} \\
\midrule
\textcircled{1} \zhen{Storyboard design}{分镜设计}   & $+0.14$ & $+0.07$ & $0.09$ & \emph{n/a} \\
\textcircled{2} \zhen{Single keyframe}{单分镜图}     & $-0.08$ & $+0.02$ & $0.37$ & \zhen{first/last frame}{首尾帧} \\
\textcircled{3} \zhen{Episode keyframes}{集分镜图}   & $+0.08$ & $+0.12$ & $0.42$ & \zhen{grid panel}{宫格} \\
\textcircled{4} \zhen{Shot video}{分镜视频} & $+0.15$ & $+0.04$ & $0.14$ & \zhen{grid panel}{宫格} \\
\textcircled{5} \zhen{Episode video}{集分镜视频}     & $+0.22$ & $+0.12$ & $0.19$ & \zhen{grid panel}{宫格} \\
\textcircled{6} \zhen{Short drama}{短剧成片}   & $+0.14$ & $+0.08$ & $0.29$ & \zhen{grid panel}{宫格} \\
\bottomrule
\end{tabular}
\end{table}

\subsection{By visual style: live action scores higher}\label{sec:slice-style}

Within the models that ran both styles, the automated composite is higher on live action at five of
the six granularities, by $0.08$ to $0.22$; single keyframes are the exception, by $0.08$ the other
way (\cref{tab:slicefx}). The composite hides the mechanism, because two axes pull against each
other: at shot video \axC{} is $0.32$ higher on live action and \axQ{} $0.25$ higher on
animation, while \axF{} is identical. The dimensional extremes are larger still, \dc{V-C3} style
consistency running over a point in favour of live action and \dc{V-A1} image quality and material
two thirds of a point the other way. Style changes which kind of defect gets found rather
than making items uniformly easier.

\subsection{By dialogue language: small, and confined to fidelity}\label{sec:slice-lang}

Scene descriptions are Chinese throughout, so the four English-dialogue dramas differ from the other
sixteen only in dialogue and on-screen text, and the two levels are balanced on visual style. The
automated composite is higher on English at every granularity but never by more than $0.12$
(\cref{tab:slicefx}), and almost all of that movement sits on one axis: \axF{} is $0.64$ higher on
English on the short drama, while \axQ{} is identical to two decimals at both stages where it
exists. The dimensions that move are the ones that carry language (event completion, speech
quality, dialogue fidelity), against cross-shot character and cross-episode consistency in the
other direction. Language moves the dimensions that carry language, not the ones that carry
pixels.

\subsection{By input paradigm: largest, and it reverses along the chain}\label{sec:slice-para}

Input paradigm is the largest of the three effects on average and the only one whose direction is not stable
along the chain: first/last frame leads grid panel by $0.37$ on single keyframes and trails it by
$0.42$ on episode keyframes, while grid panel is top at all three video granularities, though only
narrowly at shot video and at the short drama (\cref{tab:slicefx}). The single-keyframe figure
needs one correction first, since the whole \axC{}
axis exists only under grid panel there, and restricted to the ten shared dimensions the
first/last-frame lead \emph{widens} to $0.61$. Each paradigm tends to be strongest where
its inputs fix something. First/last frame hands over both endpoint images, so it leads \axF{}
input fidelity at the keyframe stage ($3.57$ against $2.73$) and the frame-anchored dimensions at
shot video (scene and prop consistency, shot framing and visual style), while trailing on
camera plausibility, which those endpoints over-constrain. Grid panel carries a whole episode in one
image, so it leads cross-shot consistency (episode keyframes $3.53$ against $3.11$) and composition,
at the cost of audio-visual sync. Multi-reference supplies no frame to adhere to, so it is last on
\axF{} ($2.92$ against $3.39$) and best on what nothing else constrains: camera plausibility,
audio-visual sync and \axE{} cinematic expressiveness.

\section{Agreement Slices of Automated Evaluation}
\label{app:agreeslices}

The four slices below hold the corpus fixed and change what is being scored: the stage, the
evaluation axis, the modality, and how reproducible the human judgement is
(\cref{tab:agreeslices}). Each asks the same question: where does \judger{} still match
the three-annotator consensus? Figures are per-item agreement against that consensus, given either
directly or as a multiple of the leave-one-out human baseline. Both boards are continuous means rather
than categorical labels, so agreement is reported as correlation, and the effects in \cref{app:slices}
are given in score units rather than standardised~\citep{cohend}, which keeps them comparable to the
$3.0$ usable line.

\begin{table}[htbp]
\centering
\caption{The four agreement slices: what each one varies, and where agreement stops holding.}
\label{tab:agreeslices}
\small
\begin{adjustbox}{max width=\linewidth}
\begin{tabular}{@{}l l l@{}}
\toprule
\textbf{Slice} & \textbf{What it varies} & \textbf{Where agreement stops} \\
\midrule
Pipeline stage    & the six granularities &
  crosses the human baseline between images and video \\
Evaluation axis   & the five evaluation axes &
  decays checkable $\rightarrow$ perceptual $\rightarrow$ aesthetic \\
Modality & text, pixels, time $+$ audio &
  one criterion degrades step by step across the three \\
Human consensus   & how reproducible the human judgement is &
  accurate on exactly the dimensions people agree on \\
\bottomrule
\end{tabular}
\end{adjustbox}
\end{table}

\subsection{By stage: the baseline is crossed between images and video}
\label{sec:slice-stage}

Read as a multiple of the leave-one-out human baseline, per-item agreement runs at $2.35\times$ on
storyboard design, $1.19\times$ on single keyframes and $1.48\times$ on episode keyframes, then
$0.79\times$, $0.85\times$ and $0.85\times$ on the three video granularities (\cref{tab:agree}). The
image side therefore sits at $1.19$--$2.35\times$ and the video side at $0.79$--$0.85\times$, and the
crossing of $1\times$ falls between the image and the video stages rather than between single-asset
and multi-asset granularities: episode keyframes already require cross-panel comparison and still
clear the baseline, while a single clip judges one artefact and does not. What decides the outcome is
the modality under judgement, not the depth along the chain (\cref{fig:ratio}).

\subsection{By axis: agreement falls as the criterion loses its external reference}
\label{sec:slice-axis}

Collapsing the $63$ leaf dimensions onto the five axes orders per-item agreement as \axF{} $0.423$,
\axC{} $0.235$, \axP{} $0.146$, \axE{} $0.132$ and \axQ{} $0.121$ (\cref{tab:axes}). Only \axF{}
is above the human baseline on net, by $+0.107$ and winning $9$ of its $12$ dimensions, while \axQ{}
wins none of its four and sits $0.086$ below. The ordering follows how much of the criterion can be
checked against something outside the artefact (an upstream output for \axF{}, a peer artefact for
\axC{}, nothing for \axQ{}), which is the same divide that made every one of the small-model
quality metrics fail as a scoring anchor (\cref{sec:audit}). This layer pools the three modalities,
so a single axis splits further once the modality changes.

\subsection{By modality: one criterion degrades from text to pixels to time and audio}
\label{sec:slice-axismod}

\Cref{tab:axes} is the axis decomposition with the modality collapsed out; \cref{tab:axismod} keeps
both factors and is read in \cref{sec:modality}. The clearest case is \axE{}: in the text modality of
the storyboard script it stands $+0.175$ above the human baseline and wins $4$ of $4$ dimensions,
while the same criterion applied to video with time and audio stands $0.190$ below it and wins $0$ of
$11$. \axF{} decays monotonically over the same three modalities ($+0.317$ on text, $+0.101$ on
pixels, $+0.012$ on time and audio), and \axQ{} is below the baseline in both modalities where it is
defined. An axis validated in one modality therefore cannot be assumed to transfer to the next, which
is why the report never quotes a single agreement figure for a criterion without saying on what it was
measured.

\begin{table}[htbp]
\centering
\caption{\zhen{Per-item agreement by evaluation axis, pooled across six stages. \hbase{} denotes
the leave-one-out human baseline; \textbf{judge} $\ge$ \hbase{} counts dimensions meeting it.
The final column describes the available references.}{按评价轴汇总六个阶段的逐条目一致性。
\hbase{} 为留一法人工基线；\textbf{judge} $\ge$ \hbase{} 统计达到该基线的维度数。末列说明可用参照。}}
\label{tab:axes}
\small
\begin{adjustbox}{max width=\linewidth}
\begin{tabular}{@{}l l c c c c c l@{}}
\toprule
\textbf{Axis} & & \textbf{dims} & \hbase{} & \textbf{judge} & $\Delta$ &
\makecell{\textbf{judge} $\ge$ \hbase} & \textbf{What the automated scorer has to work with} \\
\midrule
\axF{} Input fidelity      & & 12 & 0.316 & \best{0.423} & \up{0.107} & \pass{9/12}  & the upstream artefact, checkable line by line \\
\axC{} Internal consistency & & 24 & 0.239 & 0.235 & \dn{0.004} & 12/24 & another shot, as a referent \\
\axP{} Generation plausibility & &  8 & 0.184 & 0.146 & \dn{0.038} & 3/8   & no referent; common sense only \\
\axQ{} Visual quality       & &  4 & 0.207 & 0.121 & \dn{0.086} & \fail{0/4} & pure perception, no checkable answer \\
\axE{} Cinematic expressiveness & & 15 & 0.225 & \fail{0.132} & \dn{0.093} & 4/15  & pure impression, no checkable answer \\
\bottomrule
\end{tabular}
\end{adjustbox}
\tabnote{\zhen{Values are means of leaf-level PLCCs: judge against the three-annotator consensus,
and the leave-one-out human baseline. $\Delta$ is judge $-$ \hbase. Low agreement on \axQ{}
is consistent with the difficulty of using low-level metrics as quality-scoring anchors
(\cref{sec:step2}).}{数值为叶级PLCC均值，分别对应自动评分与三人共识的一致性及留一法人工基线。
$\Delta$ 为自动评分器 $-$ \hbase。\axQ{} 的低一致性与低层指标难以用作质量评分锚点的结果一致（\cref{sec:step2}）。}}
\end{table}

\subsection{By human consensus: accurate where people agree}
\label{sec:slice-consensus}

The last slice changes the independent variable from \emph{what} is scored to how reproducible the
human judgement on it is. Across the $63$ leaf dimensions the human baseline and the automated scorer
correlate at PLCC $0.418$ (SRCC $0.416$): dimensions on which annotators agree more closely tend to
be those on which the automated scorer also agrees more closely, rather than the two having
complementary strengths (\cref{fig:scatter63}). The association does not establish that either score
is accurate. Two things follow.
Where both sides are low the obstacle is the operational clarity of the criterion rather than
annotation budget or prompt engineering, so tuning the automated scorer there returns little. And the human
baseline rather than the automated score is the right filter for retiring a dimension, since a
dimension the annotators cannot reproduce gives the automated scorer nothing to be aligned against.

\begin{figure}[htbp]
\centering
\begin{adjustbox}{max width=\linewidth}
\begin{tikzpicture}
\begin{axis}[
  dcbase, width=11.2cm, height=7.6cm,
  xmin=-0.05, xmax=0.52, ymin=-0.17, ymax=0.80,
  xlabel={human baseline \hbase{} \ (per-item PLCC)},
  ylabel={automated evaluation \ (per-item PLCC)},
  legend style={at={(0.015,0.985)}, anchor=north west, legend columns=1, font=\scriptsize},
  legend cell align=left,
]
\addplot[black, dashed, thick, forget plot] coordinates {(-0.05,-0.05) (0.52,0.52)};
\node[anchor=south east, rotate=36, font=\scriptsize\itshape, text=black] at (axis cs:0.495,0.495) {equal agreement};
\addplot[only marks, mark=*, mark size=1.9pt, color=axF, opacity=0.9] coordinates {
  (0.1500,0.4970) (0.4420,0.7280) (0.3620,0.5250) (0.4010,0.5270) (0.3060,0.4430) (0.2240,0.2960) (0.4400,0.4850) (0.4090,0.4710) (0.3140,0.2640) (0.1630,0.1550) (0.3800,0.3240) (0.2000,0.3630)
};
\addlegendentry{\axF{} \ 12 dims}
\addplot[only marks, mark=square*, mark size=1.9pt, color=axC, opacity=0.9] coordinates {
  (0.3910,0.4590) (0.2180,0.2740) (0.1890,0.1820) (0.1660,0.3170) (0.1310,0.1300) (0.3390,0.4560) (0.1100,0.2290) (0.0980,0.1600) (0.2400,0.3040) (0.2820,0.2310) (0.3040,0.1940) (0.2450,0.1490) (0.2110,0.2470) (0.1440,0.1270) (0.3450,0.3930) (0.2880,0.2830) (0.1940,0.0990) (0.2990,0.3060) (0.1660,-0.0220) (0.1230,0.0270) (0.2270,0.3330) (0.4610,0.3020) (0.4330,0.3060) (0.1300,0.1490)
};
\addlegendentry{\axC{} \ 24 dims}
\addplot[only marks, mark=triangle*, mark size=1.9pt, color=axP, opacity=0.9] coordinates {
  (0.0460,0.0760) (-0.0230,0.1100) (0.4390,0.3110) (0.2340,0.1520) (0.2080,0.0690) (0.3310,0.2310) (0.1580,0.1360) (0.0830,0.0870)
};
\addlegendentry{\axP{} \ 8 dims}
\addplot[only marks, mark=diamond*, mark size=1.9pt, color=axQ, opacity=0.9] coordinates {
  (0.2750,0.1670) (0.1940,0.1460) (0.1910,0.0760) (0.1690,0.0960)
};
\addlegendentry{\axQ{} \ 4 dims}
\addplot[only marks, mark=pentagon*, mark size=1.9pt, color=axE, opacity=0.9] coordinates {
  (0.1990,0.2560) (0.0920,0.3070) (0.1280,0.3320) (0.0410,0.2650) (0.4740,0.3540) (0.4480,0.0750) (0.2520,0.0610) (0.2160,0.1180) (0.1470,0.0700) (0.1830,0.1220) (0.4070,0.0190) (0.1720,0.0250) (0.1220,0.0170) (0.3430,0.0740) (0.1520,-0.1120)
};
\addlegendentry{\axE{} \ 15 dims}
\node[above, font=\tiny, text=inkgray, inner sep=1.2pt] at (axis cs:0.1500,0.4970) {\dc{S-A1}};
\node[above left, font=\tiny, text=inkgray, inner sep=1.2pt] at (axis cs:0.4420,0.7280) {\dc{S-A2}};
\node[above, font=\tiny, text=inkgray, inner sep=1.2pt] at (axis cs:0.0410,0.2650) {\dc{S-C4}};
\node[below right, font=\tiny, text=inkgray, inner sep=1.2pt] at (axis cs:0.4390,0.3110) {\dc{I-E1}};
\node[below right, font=\tiny, text=inkgray, inner sep=1.2pt] at (axis cs:0.4480,0.0750) {\dc{V-E2}};
\node[below, font=\tiny, text=inkgray, inner sep=1.2pt] at (axis cs:0.4070,0.0190) {\dc{MV-E1}};
\node[below right, font=\tiny, text=inkgray, inner sep=1.2pt] at (axis cs:0.1520,-0.1120) {\dc{O-C2}};
\end{axis}
\end{tikzpicture}
\end{adjustbox}
\caption{\zhen{Automated agreement and the leave-one-out human baseline across all $63$ leaf
dimensions, measured by per-item PLCC. The dashed diagonal marks equal agreement; points above it
exceed the human baseline. The two measures are positively correlated. Most \axF{} dimensions lie
above the diagonal, while most \axE{} and all \axQ{} dimensions lie below it.}{全部 $63$ 个叶级维度的自动评分一致性与留一法人工基线，以逐条目PLCC衡量。
虚对角线表示一致性相等，线上方表示超过人工基线。两者呈正相关。
多数 \axF{} 维度位于线上方，多数 \axE{} 及全部 \axQ{} 维度位于线下方。}}
\label{fig:scatter63}
\end{figure}

\section{Case Study}
\label{app:cases}

The cases show what a single score hides (\cref{app:pairs}, \cref{app:crossshot}), where hidden defects
come from (\cref{app:shotcase}--\labelcref{sec:degrade}), how models differ once scores are broken up
(\cref{app:profiles}, \cref{app:seedance}), and where automated scoring fails (\cref{app:judgefail}).

\FloatBarrier
\subsection{Similar scores, different failure modes}
\label{app:pairs}

\begin{figure}[htbp]
\centering
\begin{subfigure}[t]{0.388\linewidth}\centering
  \includegraphics[width=\linewidth]{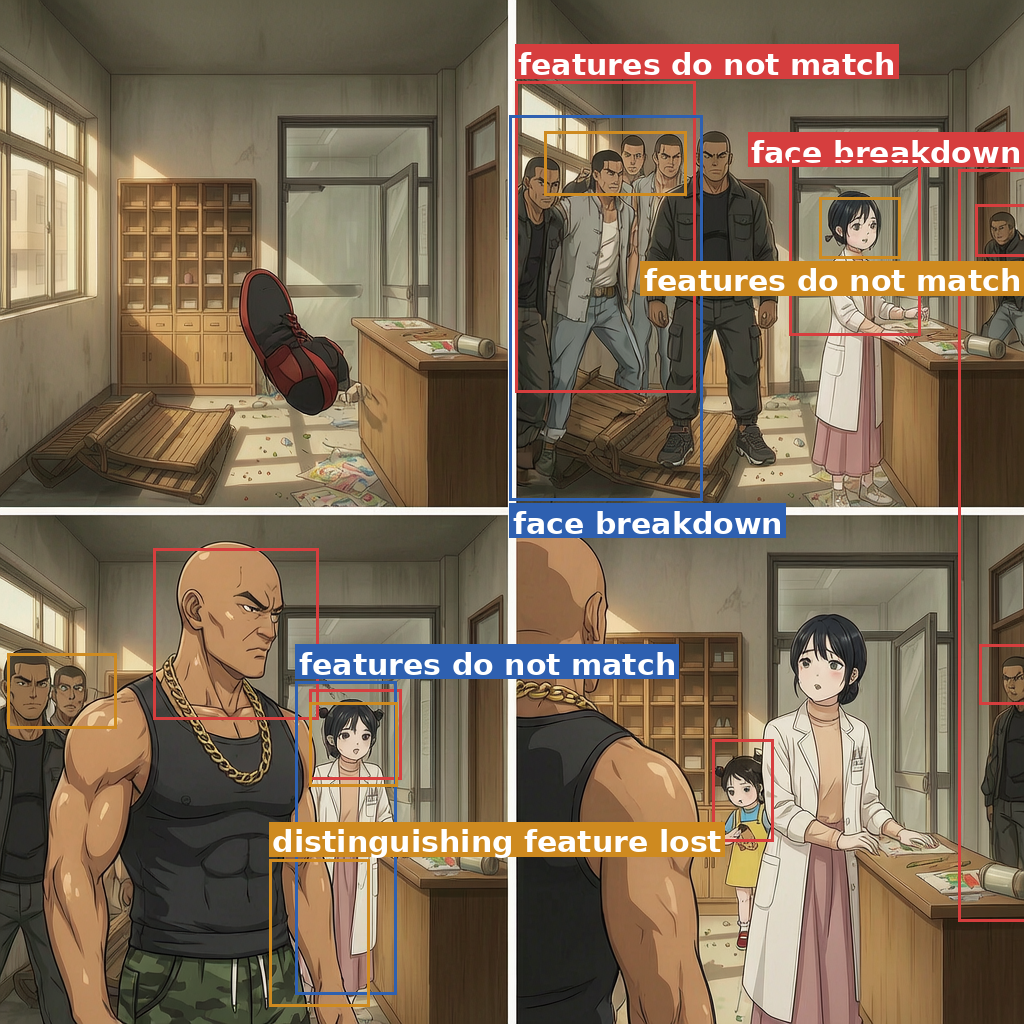}
  \subcaption{\zhen{\md{wan2.7-image-pro}: \dc{I-C1} face identity \fail{1.67}, $15$ boxes, every
  one on a face; action interaction scored $3.0$.}{\md{wan2.7-image-pro}：\dc{I-C1} 人脸身份 \fail{1.67}，$15$ 个框，
  每个都在脸上；这道题的动作交互得 $3.0$ 分。}}
  \label{fig:boxed:face}
\end{subfigure}\hfill
\begin{subfigure}[t]{0.582\linewidth}\centering
  \includegraphics[width=\linewidth]{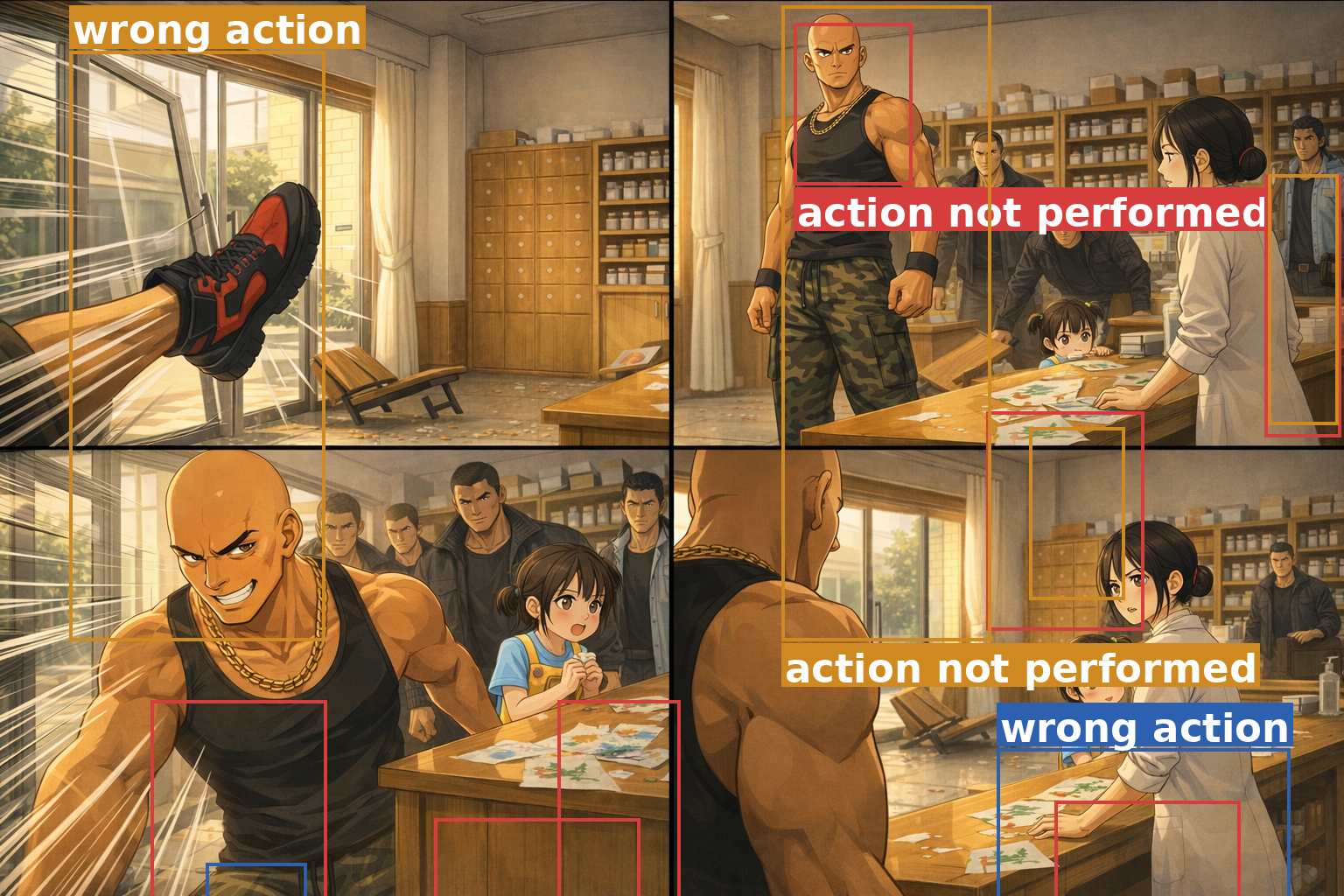}
  \subcaption{\zhen{\md{gpt-image-1.5}: \dc{I-B2} action interaction \fail{1.33}, $13$ boxes, every
  one on a limb; face identity scored $3.33$.}{\md{gpt-image-1.5}：\dc{I-B2} 动作交互 \fail{1.33}，$13$ 个框，
  每个都在肢体上；这道题的人脸身份得 $3.33$ 分。}}
  \label{fig:boxed:action}
\end{subfigure}

\vspace{7pt}

\begin{subfigure}[t]{0.485\linewidth}\centering
  \includegraphics[width=\linewidth]{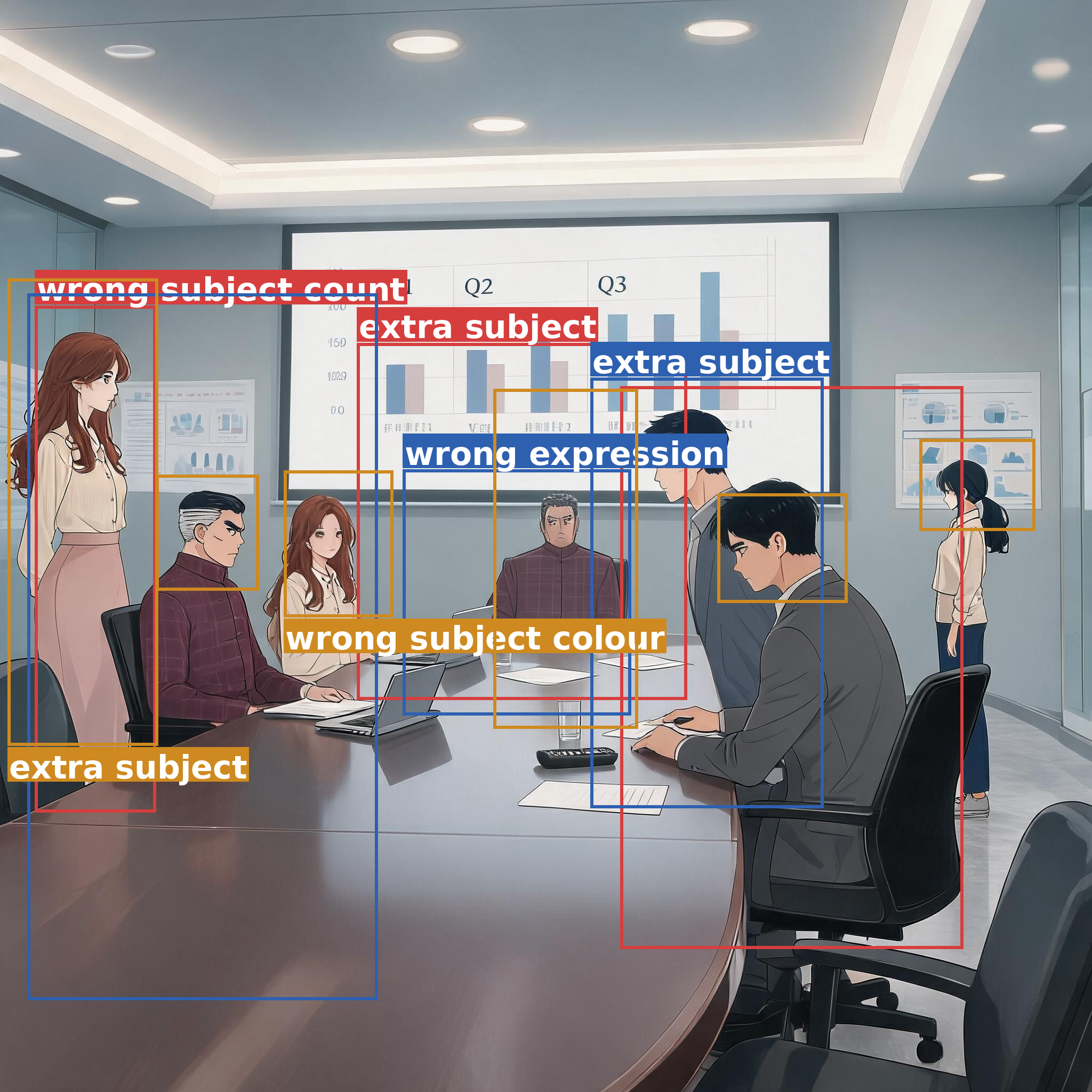}
  \subcaption{\zhen{\md{seedream-5.0-lite}: \dc{I-B1} subject attributes \fail{1.33}, $12$ boxes, all
  ``extra subject''; human anatomy scored $3.0$.}{\md{seedream-5.0-lite}：\dc{I-B1} 主体属性 \fail{1.33}，$12$ 个框，
  全部是「多余主体」；这道题的人体结构得 $3.0$ 分。}}
  \label{fig:boxed:extra}
\end{subfigure}\hfill
\begin{subfigure}[t]{0.485\linewidth}\centering
  \includegraphics[width=\linewidth]{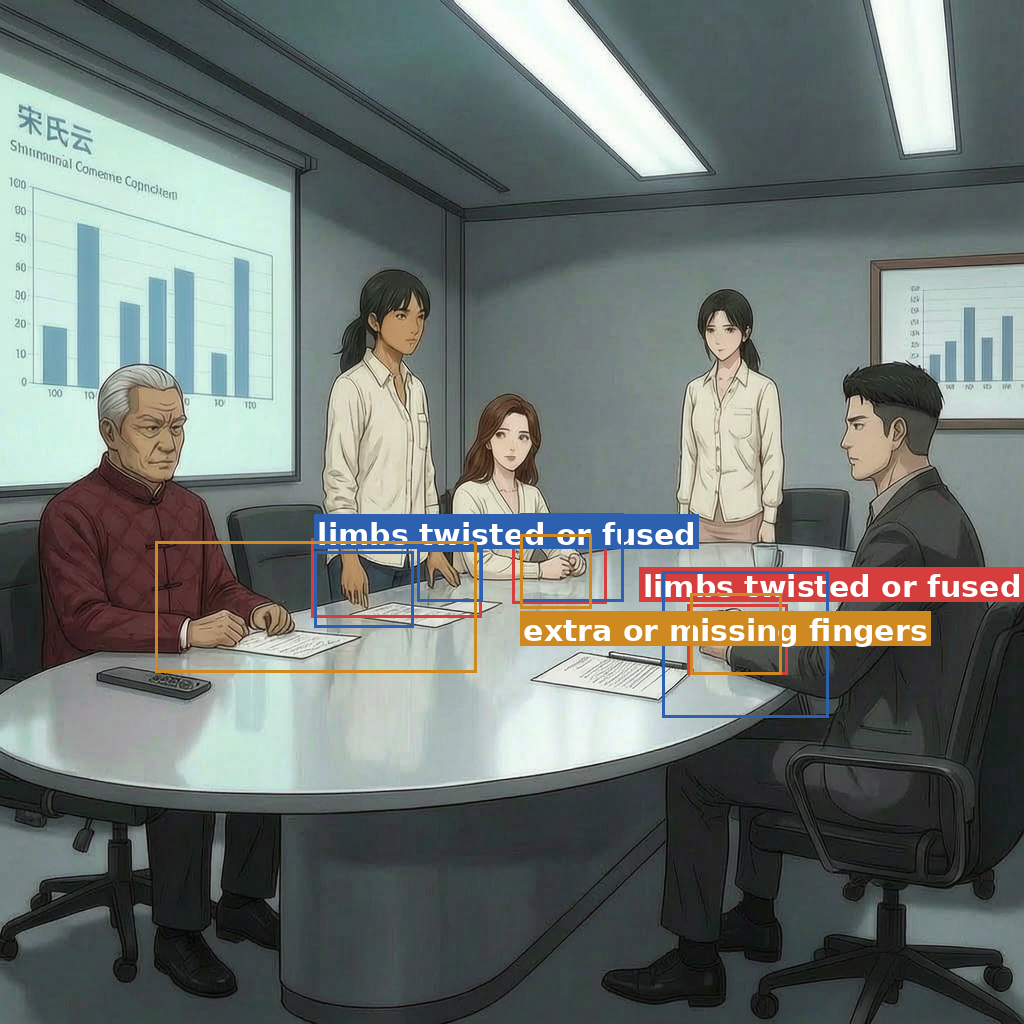}
  \subcaption{\zhen{\md{wan2.7-image-pro}: \dc{I-E1} human anatomy \fail{1.33}, $10$ boxes, all
  broken hands; subject attributes scored $3.33$.}{\md{wan2.7-image-pro}：\dc{I-E1} 人体结构 \fail{1.33}，$10$ 个框，
  全部是坏手；这道题的主体属性得 $3.33$ 分。}}
  \label{fig:boxed:hands}
\end{subfigure}

\caption{\zhen{Four items with three-annotator means of $1.3$--$1.7$ on \emph{some} dimension, which a composite would
treat as equally bad; the boxes fall in four disjoint places, and each pair swaps which dimension it
fails. Boxes are drawn independently by the three annotators, colour distinguishing them.}{四道题都在\emph{某个}维度上的三人均分为 $1.3$--$1.7$ 分，综合分会把它们视为同样差；
而框落在四个互不相交的位置，且每一对里两个模型互换了失败的维度。
框由三名标注员各自独立画出，框色区分标注员。}}
\label{fig:qual:boxed}
\end{figure}

\Cref{fig:qual:boxed} is the argument for forced attribution. All four items receive three-annotator means of $1.3$--$1.7$ on
some dimension and a composite would rank them together, but the boxes fall in four disjoint places:
on faces, on limbs mid-action, on people who should not be present, and on hands with the wrong
number of fingers. In each pair the two models trade which dimension they fail: the model that
loses identity holds action at $3.0$, and the model that loses action holds identity at $3.33$. This
is the item-level version of \cref{sec:closepairs}.

\FloatBarrier
\subsection{Cross-shot consistency is a separate capability}
\label{app:crossshot}

\begin{figure}[htbp]
\centering
\setlength{\tabcolsep}{1.4pt}
\renewcommand{\arraystretch}{0.6}

{\footnotesize\textbf{T1 \ \md{gpt-image-2}} \quad
 \dc{MI-D1} cross-shot character consistency \ \pass{\textbf{4.33}} \quad
 {\scriptsize\textcolor{inkgray}{alternate shots, $1,3,\dots,11$ of $13$}}}\\[2pt]
\begin{tabular}{@{}cccccc@{}}
\includegraphics[width=0.158\linewidth]{figures/img/x_t1_s0} &
\includegraphics[width=0.158\linewidth]{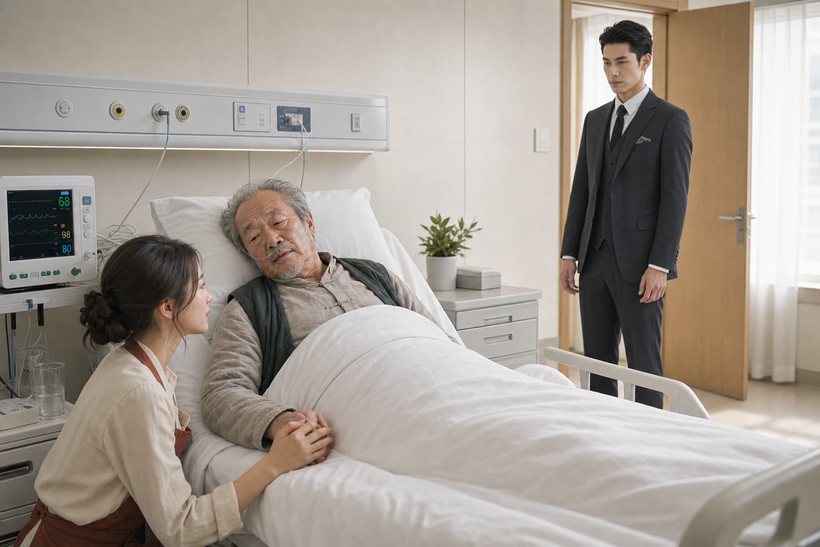} &
\includegraphics[width=0.158\linewidth]{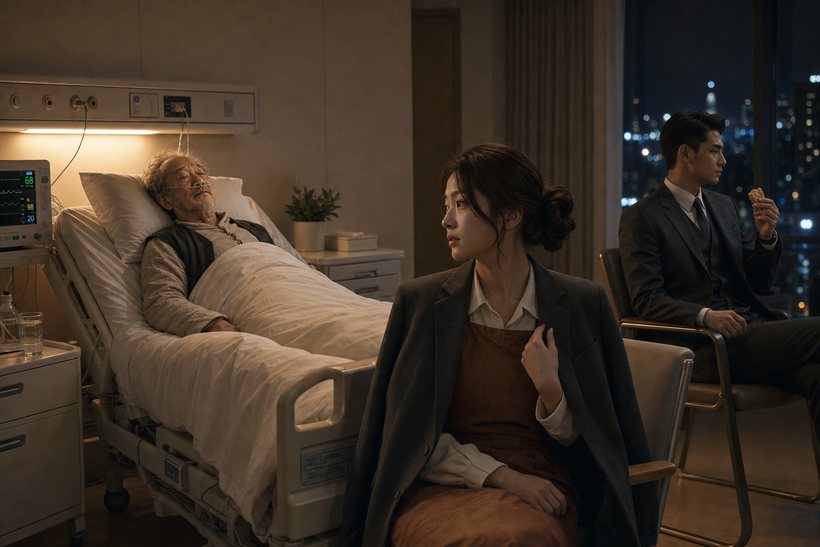} &
\includegraphics[width=0.158\linewidth]{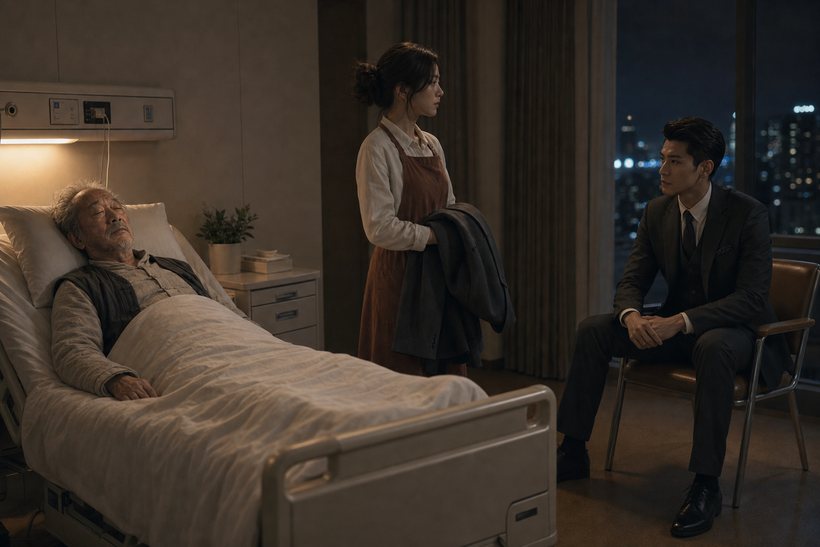} &
\includegraphics[width=0.158\linewidth]{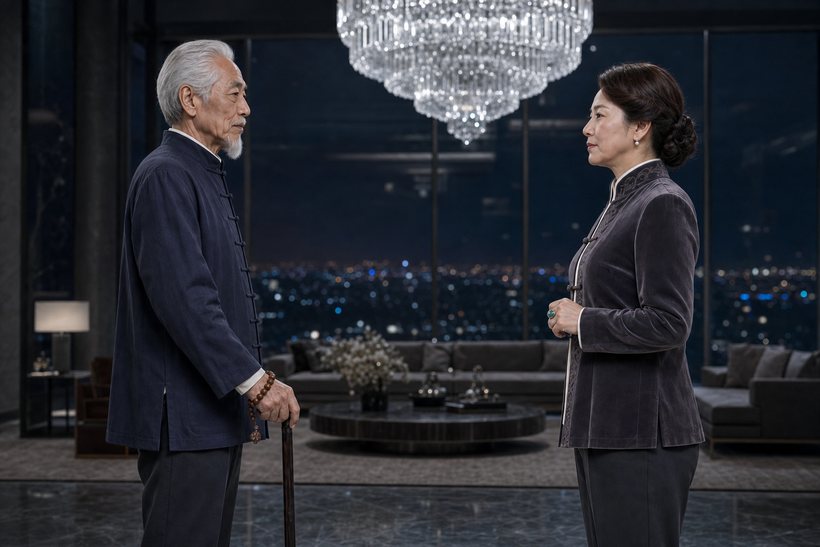} &
\includegraphics[width=0.158\linewidth]{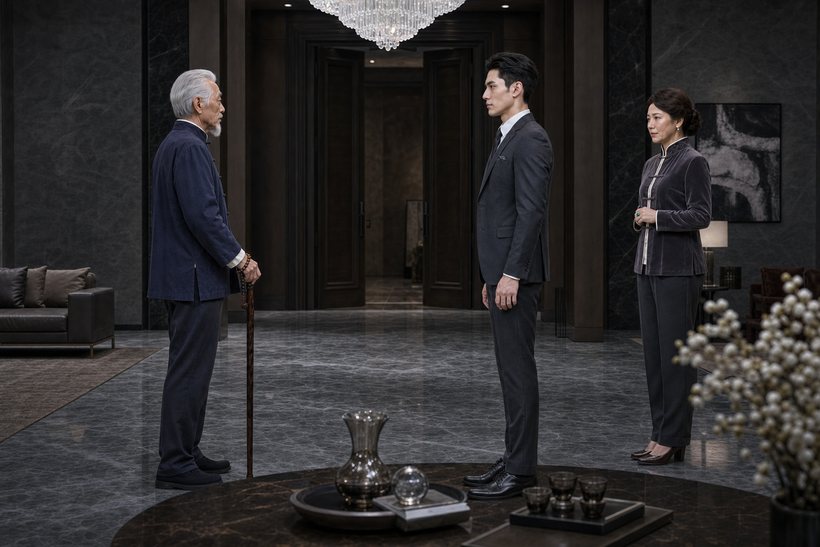} \\
{\scriptsize shot 1} & {\scriptsize shot 3} & {\scriptsize shot 5} &
{\scriptsize shot 7} & {\scriptsize shot 9} & {\scriptsize shot 11} \\
\end{tabular}

\vspace{5pt}
{\footnotesize\textbf{T4 \ \md{wan2.7-image-pro}} \quad
 \dc{MI-D1} cross-shot character consistency \ \fail{\textbf{1.67}}}\\[2pt]
\begin{tabular}{@{}cccccc@{}}
\includegraphics[width=0.158\linewidth]{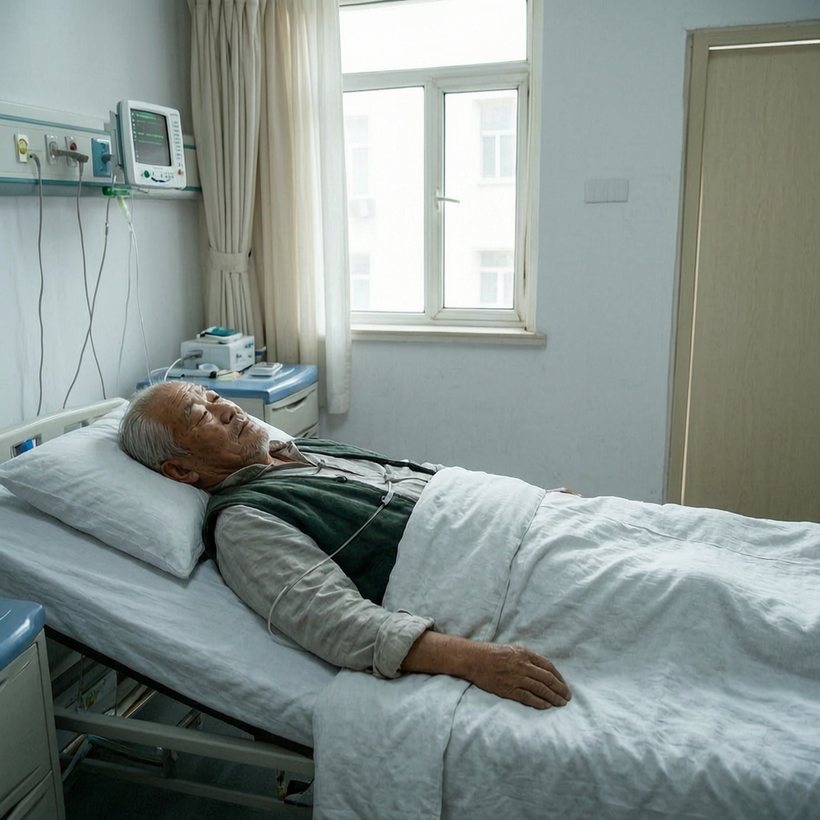} &
\includegraphics[width=0.158\linewidth]{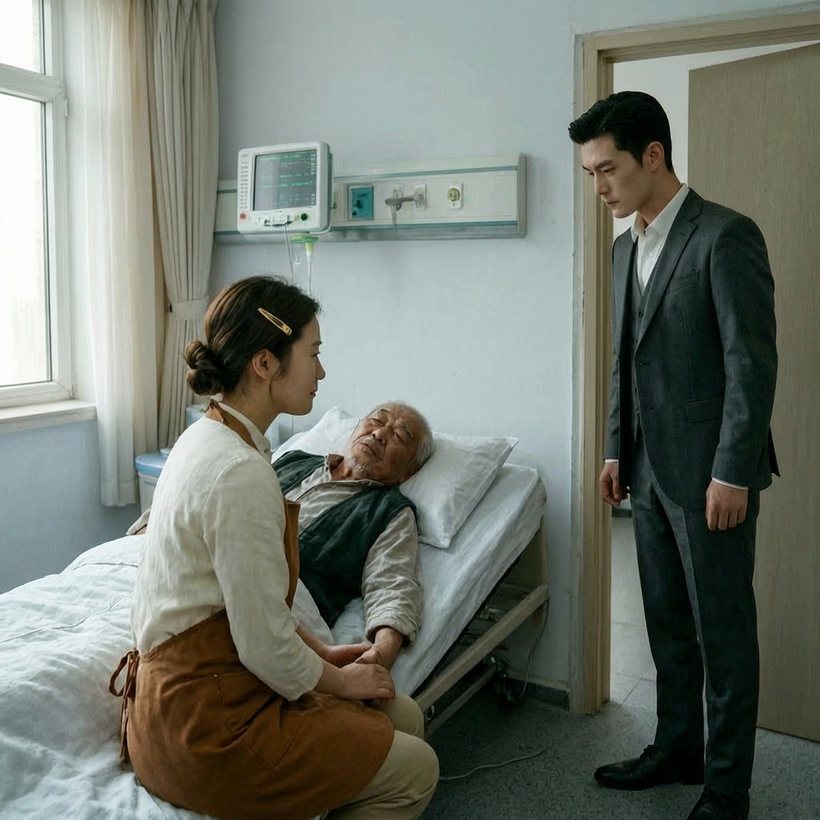} &
\includegraphics[width=0.158\linewidth]{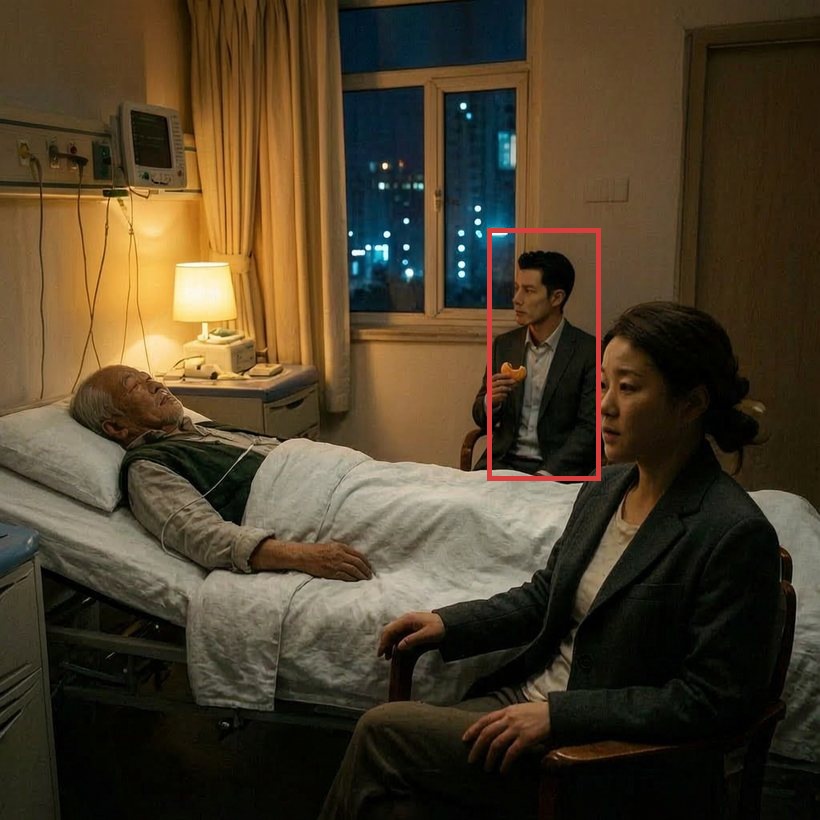} &
\includegraphics[width=0.158\linewidth]{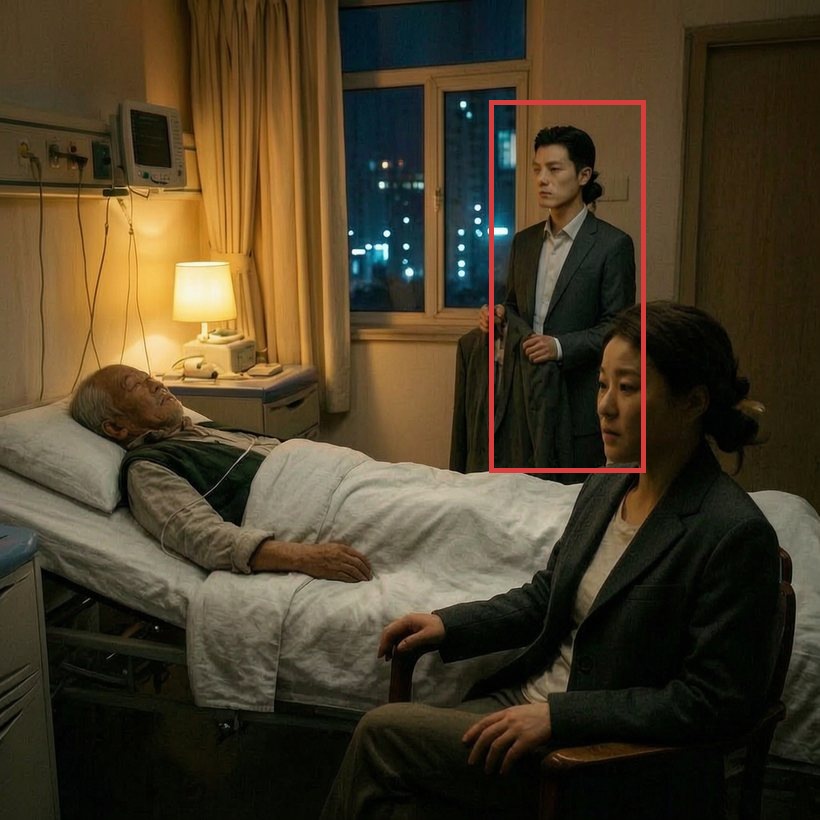} &
\includegraphics[width=0.158\linewidth]{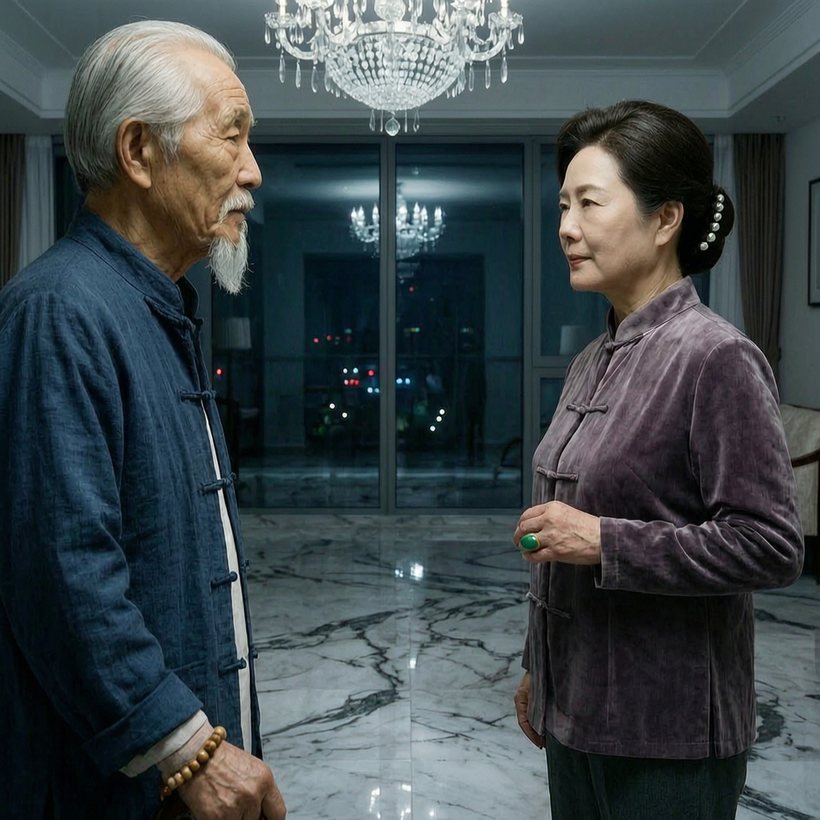} &
\includegraphics[width=0.158\linewidth]{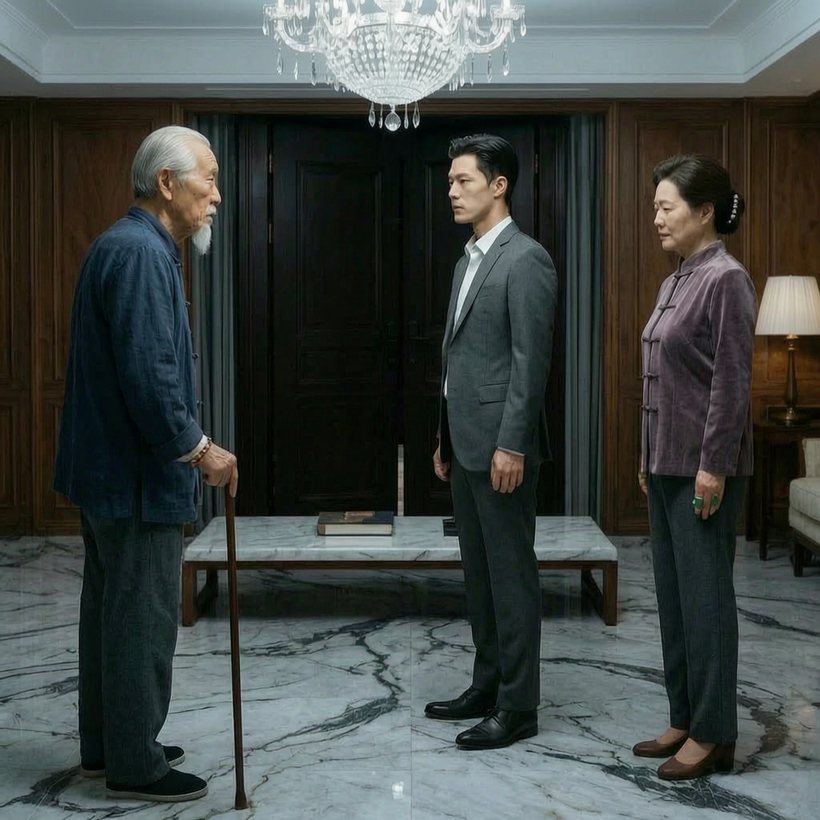} \\
{\scriptsize shot 1} & {\scriptsize shot 3} & {\scriptsize shot 5} &
{\scriptsize shot 7} & {\scriptsize shot 9} & {\scriptsize shot 11} \\
\end{tabular}

\caption{\zhen{Top- and bottom-tier models use the same episode and character sheets but differ by
$2.66$ points in three-annotator means on \dc{MI-D1}. The identity changes become apparent across shots rather than within
individual panels.}{高低档模型使用相同的单集剧本和角色设定卡，但 \dc{MI-D1} 的三人均分相差 $2.66$ 分。
身份变化需要跨镜头比较才能发现，单看各画格并不明显。}}
\label{fig:qual:crossshot}
\end{figure}

\Cref{fig:qual:crossshot} illustrates this directly. Two models draw the same
episode from the same character sheets, and the top-tier row holds its two leads across shots while
the bottom-tier row does not: a $2.66$-point gap in three-annotator means on \dc{MI-D1}. The important property is that
\emph{every individual panel in the failing row is defensible}: nothing in shot $7$ alone is wrong, it
is wrong relative to shot $5$. A single-asset board is not under-reporting cross-shot quality; it is
measuring a different quantity.

\FloatBarrier
\subsection{Shot count and storyboard executability}
\label{app:shotcase}

\Cref{tab:shots3612} puts three models on one identical episode script that forks only at the
storyboard stage. The three-shot output compresses the episode into $30$\,s of self-declared duration
and $10$ dialogue turns; all three annotators scored its executability (\dc{S-B1}) and shooting
rhythm (\dc{S-C2}) at $1.00$ with zero disagreement, marking the same few defects every time:
several actions in one shot, blocking too complex to execute, no reaction shot. Going to six shots
recovers the two \axF{} dimensions outright; going to twelve buys only executability
(\cref{fig:shotcount}).

\begin{table}[htbp]
\centering
\caption{The same episode script, three models, three shot counts. Scores are
means of the three annotators' scores. Only the storyboard stage forks; every model receives
the same script.}
\label{tab:shots3612}
\small
\begin{adjustbox}{max width=\linewidth}
\begin{tabular}{@{}l l r r r l@{}}
\toprule
\textbf{Dimension} & \textbf{Axis} &
\makecell{\md{mimo-v2.5-pro}\\\footnotesize 3 shots} &
\makecell{\md{gpt-5.5-xhigh}\\\footnotesize 6 shots} &
\makecell{\md{claude-opus-4.8-max}\\\footnotesize 12 shots} &
\textbf{Behaviour} \\
\midrule
\dc{S-A2} dialogue fidelity & \axF & \fail{1.33} & \best{5.00} & \best{5.00} & stops losing lines at 6 shots \\
\dc{S-A1} event coverage    & \axF & \fail{2.67} & \best{4.67} & \best{4.67} & same \\
\dc{S-B1} executability     & \axP & \fail{1.00} & 3.00 & \best{4.00} & rises monotonically with shot count \\
\dc{S-C2} shooting rhythm   & \axE & \fail{1.00} & \best{3.67} & \best{3.67} & 6 shots is already enough \\
\addlinespace[2pt]
Overall impression          &      & \fail{1.67} & \best{4.00} & \best{4.00} & --- \\
\addlinespace[2pt]
Problems marked             &      & 23 & 18 & 15 & fewer, and they change in kind \\
Annotator disagreement      &      & 1.75 & 0.75 & 0.63 & worse artefacts are harder to score \\
\bottomrule
\end{tabular}
\end{adjustbox}
\end{table}

\begin{figure}[htbp]
\centering
\begin{subfigure}[b]{0.545\linewidth}\centering
\begin{adjustbox}{max width=\linewidth}
\begin{tikzpicture}
\begin{axis}[dcbase, width=8.1cm, height=4.7cm,
  xmin=0.5, xmax=13.5, xtick={3,6,12},
  xticklabels={3 shots, 6 shots, 12 shots},
  xticklabel style={font=\scriptsize},
  ymin=0.6, ymax=5.4, ytick={1,2,3,4,5},
  ylabel={three-annotator consensus}, ylabel style={font=\scriptsize},
  legend style={at={(0.5,-0.24)}, anchor=north, font=\scriptsize, legend columns=2},
  legend cell align=left]
\addplot[color=axF, mark=*, mark size=2.2pt, thick]
  coordinates {(3,1.33) (6,5.00) (12,5.00)};
\addlegendentry{\dc{S-A2} dialogue fidelity}
\addplot[color=axF!60, mark=square*, mark size=2pt, thick, dashed]
  coordinates {(3,2.67) (6,4.67) (12,4.67)};
\addlegendentry{\dc{S-A1} event coverage}
\addplot[color=axP, mark=triangle*, mark size=2.6pt, thick]
  coordinates {(3,1.00) (6,3.00) (12,4.00)};
\addlegendentry{\dc{S-B1} executability}
\addplot[color=axE, mark=diamond*, mark size=2.4pt, thick, dashdotted]
  coordinates {(3,1.00) (6,3.67) (12,3.67)};
\addlegendentry{\dc{S-C2} shooting rhythm}
\addplot[black, dashed, forget plot] coordinates {(0.5,3.0) (13.5,3.0)};
\node[anchor=north west, font=\scriptsize] at (axis cs:0.7,2.94) {usable line};
\end{axis}
\end{tikzpicture}
\end{adjustbox}
\subcaption{\zhen{Scores by shot count; the $6$-shot output reaches the usable line on all four
dimensions shown.}{不同镜头数下的得分；$6$ 镜输出在图中四个维度上均达到可用线。}}
\label{fig:shot:scores}
\end{subfigure}\hfill
\begin{subfigure}[b]{0.435\linewidth}\centering
\begin{adjustbox}{max width=\linewidth}
\begin{tikzpicture}
\begin{axis}[dcbase, width=6.6cm, height=4.7cm,
  xmin=0.5, xmax=13.5, xtick={3,6,12},
  xticklabels={3, 6, 12}, xticklabel style={font=\scriptsize},
  xlabel={shots}, xlabel style={font=\scriptsize},
  ymin=0, ymax=25, ytick={0,5,10,15,20,25},
  ylabel={problems marked by the three annotators}, ylabel style={font=\scriptsize},
  axis y line*=left,
  legend style={at={(0.5,-0.24)}, anchor=north, font=\scriptsize, legend columns=1}]
\addplot[color=tierfour, mark=*, mark size=2.2pt, very thick]
  coordinates {(3,23) (6,18) (12,15)};
\addlegendentry{problems marked}
\addplot[color=inkgray, mark=square*, mark size=2pt, thick, dashed]
  coordinates {(3,17.5) (6,7.5) (12,6.3)};
\addlegendentry{disagreement ($\times 10$)}
\node[anchor=south east, align=right, font=\scriptsize, text=inkgray]
  at (axis cs:13.2,16.5)
  {at 12 shots \textbf{none} of the\\15 problems is an omission};
\end{axis}
\end{tikzpicture}
\end{adjustbox}
\subcaption{\zhen{Fewer problems, and they change in kind.}{问题变少，且性质在变。}}
\label{fig:shot:problems}
\end{subfigure}

\caption{\zhen{Three models produce different shot counts from the same $1{,}861$-character episode
script, forking only at storyboard design. (\subref{fig:shot:scores}) Going from $3$ to $6$ shots
recovers dialogue fidelity completely (at $3$ shots the model deletes the scene's opening question
outright), while $6$ to $12$ improves only executability. (\subref{fig:shot:problems}) In this case,
disagreement falls with the problem count: \emph{the more defects an output has, the harder it is
to agree how many points to deduct} (\cref{app:shotcase}).}{三个模型基于同一份 $1{,}861$ 字的分集剧本产出不同镜头数，只在分镜环节分叉。
（\subref{fig:shot:scores}）从 $3$ 镜到 $6$ 镜，对白保真完全恢复
（$3$ 镜时模型直接删掉了这场戏的开场提问），而从 $6$ 镜到 $12$ 镜只改善描述可执行性。
（\subref{fig:shot:problems}）在这一案例中，分歧度随问题数一同下降：
\emph{产物问题越多，人越难就该扣几分达成一致}（\cref{app:shotcase}）。}}
\label{fig:shotcount}
\end{figure}

\FloatBarrier
\subsection{Input paradigms expose different failure modes}
\label{app:paradigmcase}

\begin{figure}[htbp]
\centering
\begin{subfigure}[t]{0.344\linewidth}\centering
  \includegraphics[width=\linewidth]{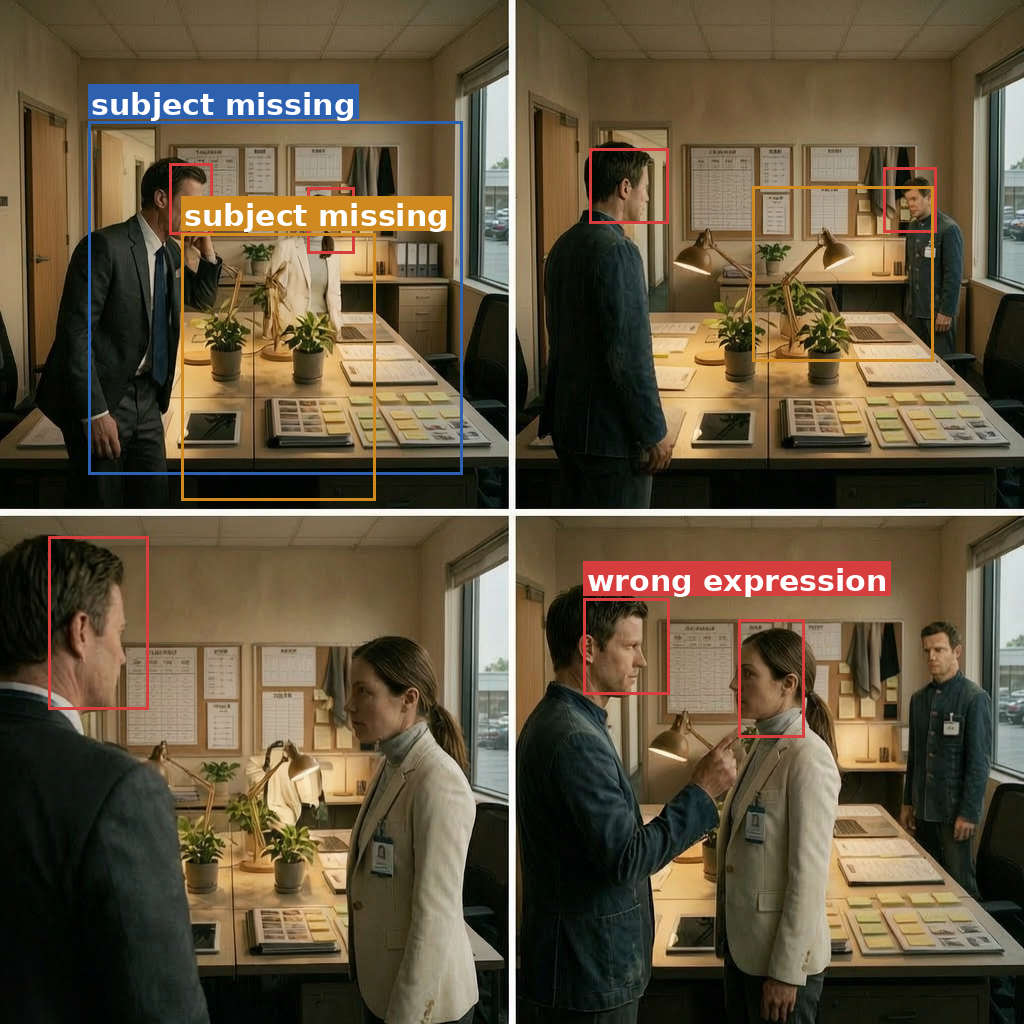}
  \subcaption{\zhen{\textbf{Grid panel omits people.} \md{wan2.7-image-pro}, \dc{I-B1} subject
  attributes \fail{1.33}: cells that should hold three people hold two.}{\textbf{宫格漏画人物。}\md{wan2.7-image-pro}，
  \dc{I-B1} 主体属性 \fail{1.33}：本该有三个人的格子里只有两个。}}
  \label{fig:disease:grid}
\end{subfigure}\hfill
\begin{subfigure}[t]{0.616\linewidth}\centering
  \includegraphics[width=\linewidth]{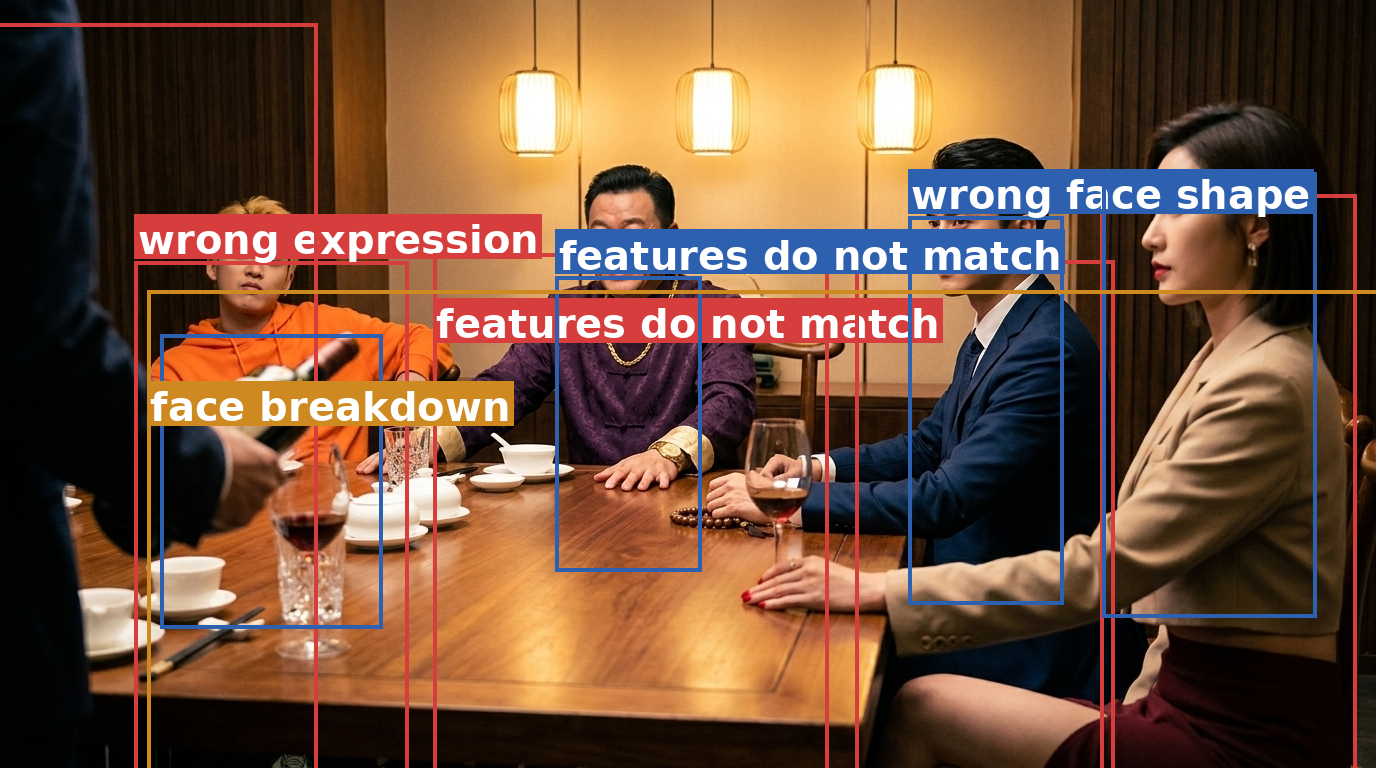}
  \subcaption{\zhen{\textbf{Identity drift between endpoint frames.} \md{nano-banana-pro},
  \dc{I-C1} face identity \fail{1.33}: all three annotators independently marked the identity mismatch.}{\textbf{首尾帧之间的身份漂移。}\md{nano-banana-pro}，
  \dc{I-C1} 人脸身份 \fail{1.33}：三名标注员均独立标记了身份不一致。}}
  \label{fig:disease:ff}
\end{subfigure}

\vspace{6pt}
\begin{subfigure}[b]{\linewidth}\centering
  \setlength{\tabcolsep}{0pt}%
  \begin{tabular}{@{}*{4}{>{\centering\arraybackslash}m{0.25\linewidth}}@{}}
  \includegraphics[height=1.55cm]{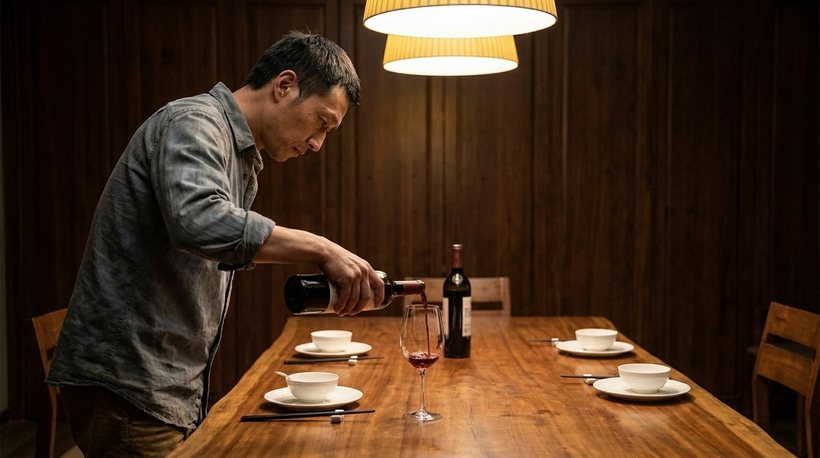} &
  \includegraphics[height=1.55cm]{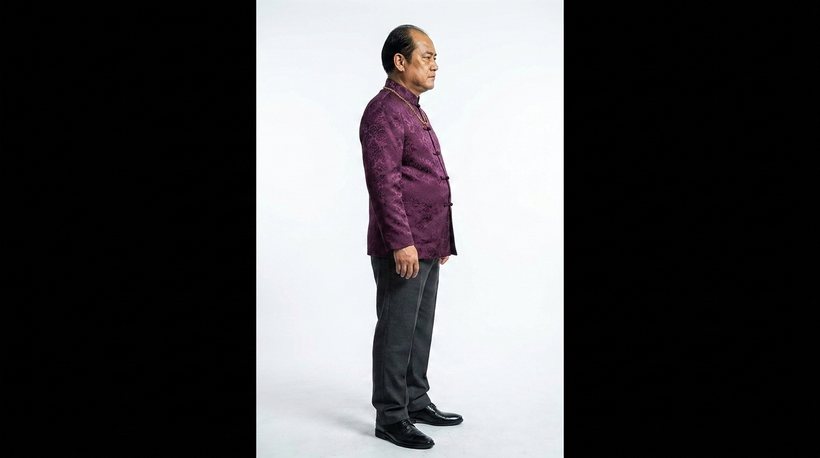} &
  \includegraphics[height=1.55cm]{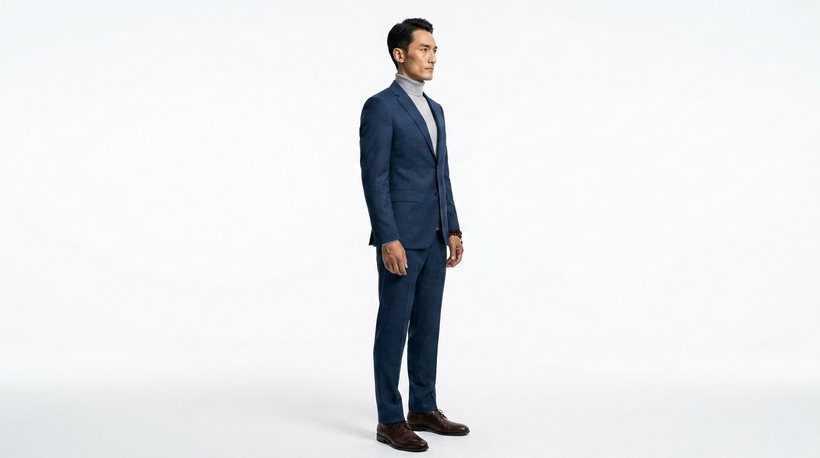} &
  \includegraphics[height=1.55cm]{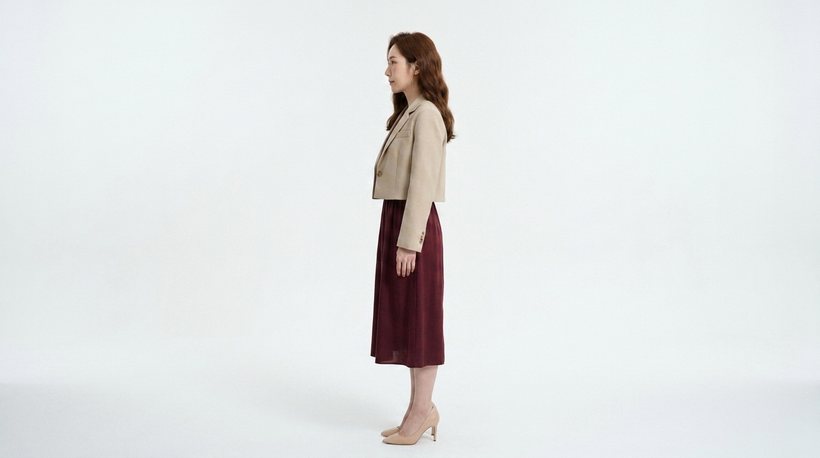} \\
  \end{tabular}
  \subcaption{\zhen{The input for (\subref{fig:disease:ff}): the shot's own first frame (left) and
  three character sheets.}{（\subref{fig:disease:ff}）的输入：该镜首帧（左）与三张角色设定卡。}}
  \label{fig:disease:refs}
\end{subfigure}

\caption{\zhen{Examples of distinct failure modes under the two input paradigms. Scores are three-annotator means; the numbers of
below-threshold dimensions are $1/10$ for first/last frame and $9/15$ for grid panel.}{两种输入范式的不同缺陷示例。
分数为三名标注员的均分；低于阈值的维度数分别为首尾帧 $1/10$、宫格 $9/15$。}}
\label{fig:qual:disease}
\end{figure}

\Cref{fig:qual:disease} pairs the two characteristic failure modes, and neither is available to the
other paradigm: grid panels omit people, because one image must fill several cells, and
first/last-frame shots change person between the two frames, because the frames are not produced in
one pass and identity is not held across them. That is why difficulty is set by paradigm rather than by visual style, and
why pooling paradigms dilutes the grid collapse.

On video the same split appears item by item (\cref{tab:paradigm}). Under grid one \md{pixverse-c1}
shot received a three-annotator mean of $1.67$ on keyframe adherence with all three annotators marking the same defect class,
``deviates from the given keyframe'', localised to $0.29$--$11.04$\,s, $2.98$--$4.87$\,s and
$6.56$--$8.15$\,s. The prompt in this paradigm describes only camera moves and dialogue, so
appearance and blocking are anchored solely by the grid image. Once that image is split into
cells, adherence to it is markedly weaker than to a first/last frame pair.

\FloatBarrier
\subsection{Upstream defects persist downstream}
\label{sec:degrade}

\begin{figure}[!htbp]
\centering
\setlength{\tabcolsep}{2pt}
\renewcommand{\arraystretch}{0.6}
\begin{tabular}{@{}cccc@{}}
\includegraphics[width=0.235\linewidth]{figures/img/drift_s0} &
\includegraphics[width=0.235\linewidth]{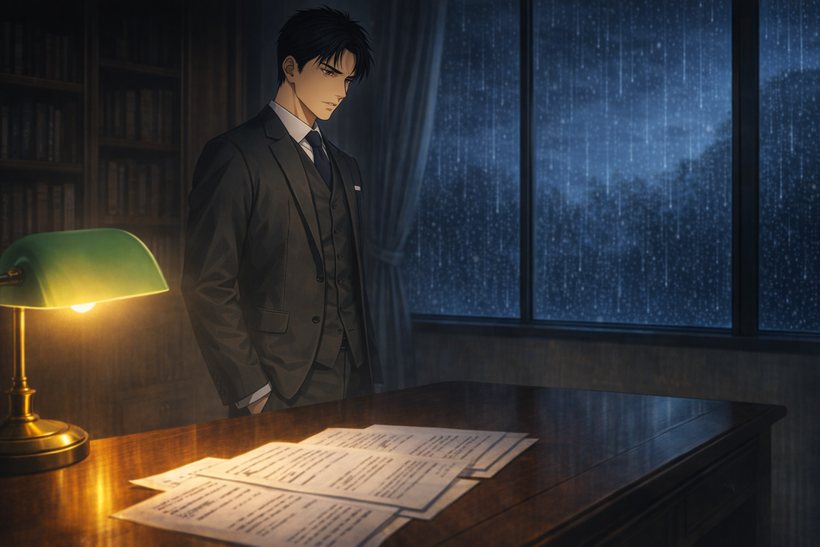} &
\includegraphics[width=0.235\linewidth]{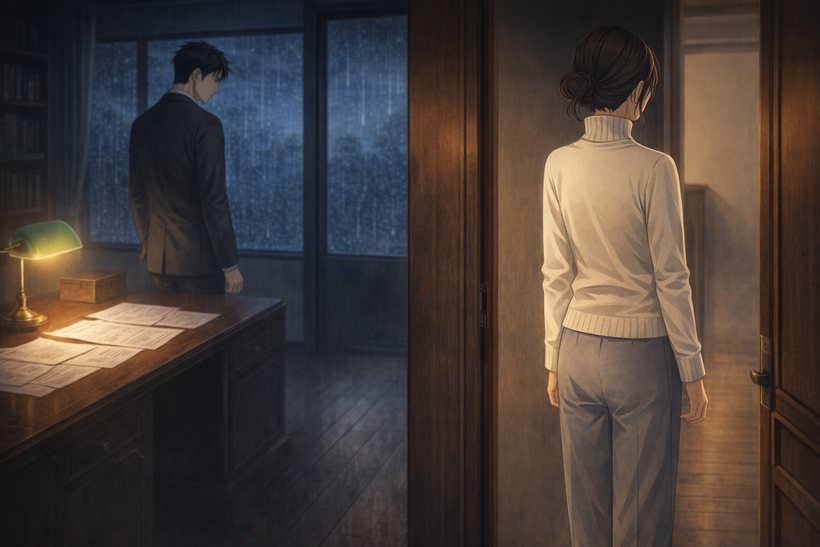} &
\includegraphics[width=0.235\linewidth]{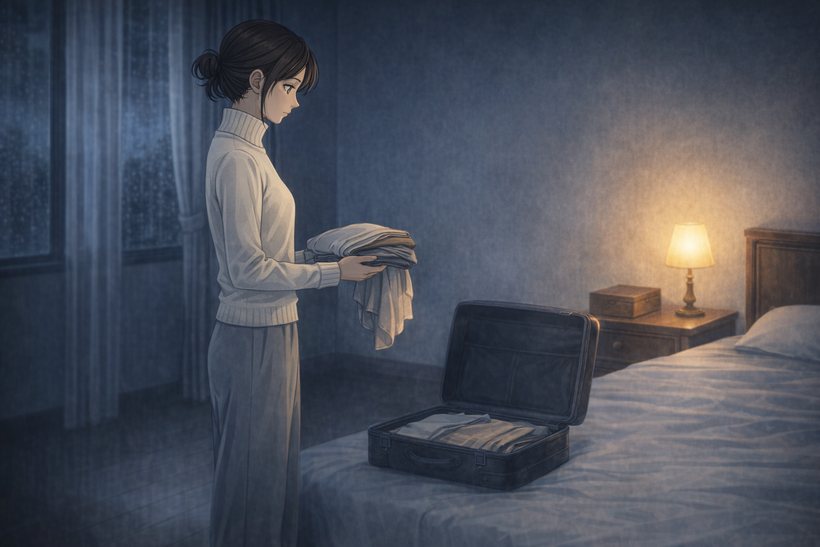} \\
{\scriptsize shot 1} & {\scriptsize shot 2} & {\scriptsize shot 3} & {\scriptsize shot 4} \\
\end{tabular}

\vspace{3pt}
{\footnotesize\md{gpt-image-1.5}, four consecutive shots \quad
 \dc{MI-D3} scene \fail{\textbf{2.00}} \quad vs \quad \dc{MI-D1} character \pass{\textbf{3.67}}}

\caption{\zhen{Scene consistency fails before character consistency: all three annotators marked
set-dressing changes across these consecutive shots, while the characters remained consistent.
Across the corpus, scene consistency has the lowest human mean ($2.42$) among the four
episode-keyframe dimensions.}{场景一致性先于角色一致性出现问题：三名标注员均标记了这些连续镜头中的陈设变化，而角色保持一致。
在全量语料中，场景一致性的人工均分（$2.42$）是四个单集关键帧维度中最低的。}}
\label{fig:qual:drift}
\end{figure}

{\sloppy \Cref{fig:prop} uses four dramas (two live action, two animated) under the first/\allowbreak last-frame
paradigm with the video model fixed. The baseline chain scores highest at every granularity ($3.03$--$3.29$ in
automated scores), and each of the four substitutions replaces one upstream stage and is read at
three granularities (\cref{sec:prop}); these chains were scored automatically and carry no human
annotation. \Cref{fig:qual:drift} shows the upstream side in one item: set dressing changes across
consecutive keyframes while character identities remain stable.\par}

\subsection{\texorpdfstring{\md{hy4-preview}}{hy4-preview} and \texorpdfstring{\md{hy3}}{hy3}: fidelity and production executability}
\label{app:profiles}

\Cref{tab:hy3} expands the automatic storyboard comparison in \cref{tab:main}.
\md{hy4-preview} scores higher than \md{hy3} overall ($4.02$ versus $3.89$), with higher \axF{}
($4.69$ versus $4.36$) and \axE{} ($3.91$ versus $3.78$) but lower \axP{} ($3.56$ versus $3.65$).
Better script reproduction need not mean better production executability.
\md{hy3} and \md{doubao-seed-2.1-pro} both score $3.89$ overall, but the latter leads in emotional
expression and narrative flow.

\begin{table}[htbp]
\centering
\caption{Automatic storyboard means ($n=60/57/60$ in column order). Samples are not fully paired;
\md{hy4-preview} has automatic scores only.}
\label{tab:hy3}
\small
\begin{adjustbox}{max width=\linewidth}
\begin{tabular}{@{}l l r r r@{}}
\toprule
\textbf{Dimension} & \textbf{Axis} & \md{hy4-preview} & \md{hy3} & \md{doubao-seed-2.1-pro} \\
\midrule
\dc{S-A1} event coverage         & \axF & \best{4.53} & 4.31 & 4.27 \\
\dc{S-A2} dialogue fidelity      & \axF & \best{4.84} & 4.42 & 4.25 \\
\dc{S-B1} executability          & \axP & 3.73 & 3.82 & \best{3.90} \\
\dc{S-B2} restraint in additions & \axP & 3.38 & \best{3.48} & 3.40 \\
\dc{S-C1} narrative flow         & \axE & 4.33 & 4.26 & \best{4.40} \\
\dc{S-C2} shooting rhythm        & \axE & \best{3.05} & 2.91 & 2.58 \\
\dc{S-C3} emotional expression   & \axE & 4.23 & 4.05 & \best{4.33} \\
\dc{S-C4} audiovisual style      & \axE & \best{4.03} & 3.89 & 3.99 \\
\addlinespace[2pt]
Overall                        &      & \best{4.02} & 3.89 & 3.89 \\
\bottomrule
\end{tabular}
\end{adjustbox}
\tabnote{Bold marks the highest mean in each row; Overall averages the eight leaf dimensions.}
\end{table}

On episode~$2$ of the animated \emph{I Insist on Taking You} ($20$ dialogue turns), the three outputs
keep $20/20/3$ turns verbatim in $10/10/7$ shots: \md{hy4-preview} adds camera movement and blocking
but chains several actions within one shot, \md{hy3} adds little beyond ``rapid cut'' and
``close-up'', and \md{doubao-seed-2.1-pro} adds visual detail but rewrites dialogue.

On the human side, three-annotator means over the annotated storyboard items give \md{hy3} a
composite of $3.43$ ($57$ items) against $3.60$ for \md{doubao-seed-2.1-pro} ($60$ items). Across the
eight leaf dimensions, \md{hy3} ranks highest on dialogue fidelity ($4.44$, fourth of nine and $0.20$
behind the leader) and lowest on emotional expression ($3.64$) and audiovisual style treatment
($3.91$), both eighth of nine. In the episode compared above, all three annotators flagged
\md{doubao-seed-2.1-pro}'s dialogue for tampering or omission.

\subsection{\texorpdfstring{\md{seedance-2.5}}{seedance-2.5}: stronger episode coherence, weaker clip quality}
\label{app:seedance}

In automated scores on paired items (\md{seedance-2.5} has no human annotation), the newer version loses on shot video and gains at both episode and
short-drama granularity (\cref{fig:seedance}). Everything that has to hold continuously along a
timeline got worse (audio-visual sync $-0.658$, motion smoothness, camera plausibility), and
everything about joining shots together got better (cross-shot style consistency $+1.143$). One likely contributor is that \md{seedance-2.5}
follows the prompt more literally, so a deficiency already in the prompt is executed faithfully
(\cref{fig:seed:case}); longer shots ($15$\,s to $30$\,s) also pack four internal cuts into one scored
shot.

\begin{figure}[htbp]
\centering
\begin{subfigure}[t]{0.40\linewidth}\centering
\begin{adjustbox}{max width=\linewidth}
\adjustbox{max width=\linewidth}{%
\begin{tikzpicture}[x=5.0cm, y=1.32cm]
\def\X#1{{(#1-3.0)/0.8}}
\draw[rulegray] (\X{3.0},0.3) -- (\X{3.8},0.3);
\foreach \v in {3.0,3.2,3.4,3.6,3.8}
  {\draw[rulegray] (\X{\v},0.3) -- (\X{\v},0.22);
   \node[font=\scriptsize, text=inkgray, anchor=north] at (\X{\v},0.2) {\v};}
\foreach \y/\lab/\a/\b/\d in {%
  3/{shot video}/3.261/3.134/{$-0.127$},
  2/{episode video}/3.127/3.306/{$+0.179$},
  1/{short drama}/3.484/3.699/{$+0.215$}}
{
  \node[font=\scriptsize, anchor=east] at (\X{2.99},\y) {\lab};
  \draw[rulegray!70, line width=1.4pt] (\X{\a},\y) -- (\X{\b},\y);
  \draw[fill=white, draw=inkgray, line width=0.7pt] (\X{\a},\y) circle (2.2pt);
  \fill[axC!85] (\X{\b},\y) circle (2.4pt);
  \node[font=\scriptsize, anchor=west, text=inkgray] at (\X{3.81},\y) {\d};
}
\node[font=\scriptsize, anchor=west, text=inkgray] at (\X{2.99},3.8) {$\circ$\,2.0 \quad $\bullet$\,2.5};
\end{tikzpicture}}
\end{adjustbox}
\subcaption{\zhen{Paired means at the three video granularities.}{三个视频粒度上的配对均分。}}
\label{fig:seed:paired}
\end{subfigure}\hfill
\begin{subfigure}[t]{0.575\linewidth}
\begin{adjustbox}{max width=\linewidth}
\begin{tikzpicture}
\node[draw=rulegray, fill=softrow, rounded corners=2pt, inner sep=5pt, text width=8.0cm,
      align=left, font=\scriptsize] {%
  \textbf{\textsf{Multi-reference $\cdot$ animated $\cdot$ EP1 shot 2 $\cdot$ 21.1\,s}}\\[2pt]
  \dc{V-E2} audio-visual sync: \fail{$1$} \ (\md{seedance-2.0} on the same item: \pass{$5$})\\[3pt]
  \textcolor{inkgray}{\emph{Prompt, excerpt.}} shot 3 --- she lowers her eyes, still facing right,
  \{\,Come clean, leave clean. I owe nobody.\,\} \ shot 4 --- \textbf{Fu Jingxing in profile}
  [Image 2] stands in the doorway, facing the open suitcase, his gaze passing over
  \textbf{Qiao Wan from behind} [Image 4]; then, voice lowered, \{\,You are packing now?\,\}\\[3pt]
  \textcolor{inkgray}{\emph{Reference sheets} ($10$ for this shot, $4$ named).} Qiao Wan $\cdot$
  front \ $|$ \ Qiao Wan $\cdot$ \textbf{back} \ $|$ \ Fu Jingxing $\cdot$ \textbf{profile} \ $|$ \
  master bedroom\\[3pt]
  Both lines have a named speaker, and the prompt pins both speakers to a back view or a profile
  silhouette; one of the ten sheets is a dedicated rear view. There is no mouth for the dialogue
  to come out of.};
\end{tikzpicture}
\end{adjustbox}
\subcaption{\zhen{A case in which following the prompt preserves its conflicting requirements.}{遵循提示词却保留其中冲突要求的案例。}}
\label{fig:seed:case}
\end{subfigure}

\caption{\zhen{Automated scores of \md{seedance-2.5} and \md{seedance-2.0} on paired animated, multi-reference items,
matched by drama, episode and shot ($39$/$14$/$10$ pairs). The newer version scores higher at
episode and short-drama granularity but slightly lower on individual clips.}{\md{seedance-2.5} 与 \md{seedance-2.0}
在动画、多参考条目上的配对比较（自动化评分），按剧、集和镜位匹配（$39$/$14$/$10$ 对）。
新版本在单集和整剧粒度上得分更高，但单条镜头视频得分略低。}}
\label{fig:seedance}
\end{figure}

\FloatBarrier
\subsection{Automated scoring without an external reference}
\label{app:judgefail}

\begin{figure}[!htbp]
\centering
\begin{minipage}[c]{0.335\linewidth}\centering
  \includegraphics[width=\linewidth]{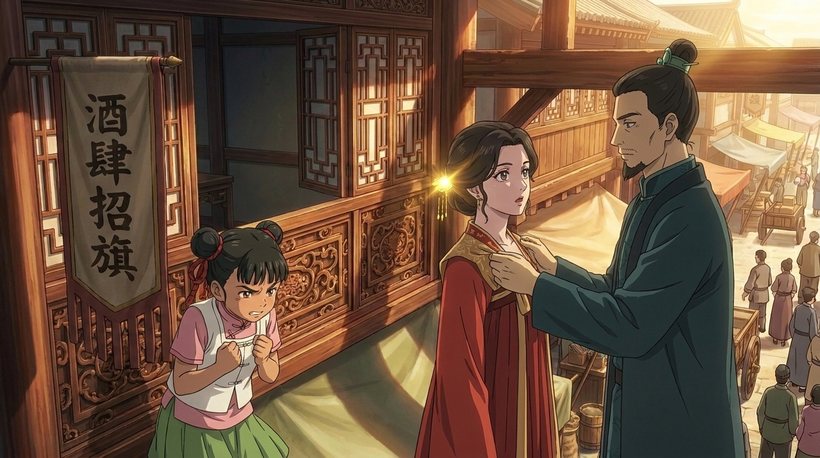}
\end{minipage}\hfill
\begin{minipage}[c]{0.635\linewidth}
\footnotesize
\setlength{\tabcolsep}{4pt}
\begin{tabular}{@{}l r r r l@{}}
\toprule
\textbf{Dimension} & \textbf{human} & \textbf{judge} & \textbf{err} & \\
\midrule
\dc{I-B1} subject attributes  & 2.00 & 2 & \pass{0.00} & correct \\
\dc{I-C2} appearance, costume & 2.33 & 2 & \pass{0.33} & correct \\
\dc{I-B3} scene and layout    & 2.67 & 3 & \pass{0.33} & correct \\
\addlinespace[2pt]
\dc{I-A1} technical quality   & \fail{1.33} & \fail{5} & \fail{3.67} & \textbf{overestimated} \\
\bottomrule
\end{tabular}

\vspace{4pt}
The three annotators scored technical quality $1$\,/\,$1$\,/\,$2$ and tagged it consistently:
structural distortion, visible noise, blur, ghosting.
\end{minipage}

\caption{\zhen{One \md{nano-banana-pro} panel, all four scored dimensions shown: every dimension
with a reference to check against is right, and the one purely perceptual dimension is off by almost
four points in the lenient direction.}{一张 \md{nano-banana-pro} 的分镜图，四个打分维度全部列出：
凡有参照物可核对的维度都判对了，而唯一一个纯感知维度偏高近四分。}}
\label{fig:qual:judgefail}
\end{figure}

\Cref{fig:qual:judgefail} shows the failure mode in one item: dimensions with a reference are scored
correctly, while the purely perceptual one is overestimated by $3.67$ points, the item-level form of
what metric validation found (\cref{sec:step2}, \cref{tab:axes}).

\end{document}